%% file: main.tex
\documentclass[12pt,letterpaper]{article}

\usepackage[T1]{fontenc}
\usepackage[utf8]{inputenc}
\usepackage[margin=1in]{geometry}
\usepackage{times}
\usepackage{mathptmx}
\usepackage{amsmath,amssymb}
\usepackage{booktabs,multirow}
\usepackage{longtable}
\usepackage{xspace}
\usepackage{graphicx}
\usepackage{epsfig}
\usepackage{url}
\usepackage{microtype}
\usepackage{siunitx}
\usepackage{enumitem}
\usepackage[numbers,sort&compress]{natbib}
\usepackage[font=small,labelfont=bf,labelsep=period,skip=3pt]{caption}
\usepackage{subcaption}
\usepackage{pifont}

\usepackage[dvipsnames]{xcolor}
\definecolor{hkustblue}{rgb}{0.0,0.2,0.4}
\definecolor{abstractgray}{RGB}{240,242,245}
\definecolor{badgegold}{RGB}{201,162,70}
\definecolor{accentblue}{rgb}{0.0,0.36,0.68}
\definecolor{accentteal}{rgb}{0.05,0.45,0.50}
\usepackage[pagebackref,breaklinks,colorlinks,linkcolor=black,citecolor=black,urlcolor=black,
  pdftitle={Unimodal Shortcuts and Joint Scene-Intent Dependence in Vision-Language Safety Representations},
  pdfauthor={Guijia Zhang, Hongbo Qiu, Guohao Cai, Hao Zheng, Jiajun Zha, Hanqing Wang, Zhenhao Zhang, Huihao Jing, Shing-Chi Cheung, Haoxuan Che, Harry Yang},
  pdfsubject={Multimodal AI safety evaluation, probing, and shortcut audits},
  pdfkeywords={vision-language models, AI safety, probing, shortcut learning, multimodal evaluation},
  pdfcreator={},pdfproducer={}]{hyperref}

\usepackage[capitalize]{cleveref}
\crefname{section}{Sec.}{Secs.}
\Crefname{section}{Section}{Sections}
\crefname{table}{Table}{Tables}
\crefname{figure}{Fig.}{Figs.}
\Crefname{figure}{Figure}{Figures}
\Crefname{equation}{Eq.}{Eqs.}

\input{facts.tex}
\newcommand{\F}[1]{\csname f#1\endcsname}

\newcommand{\auroc}{AUROC\xspace}
\newcommand{\fpr}{FPR\xspace}
\newcommand{\tpr}{TPR\xspace}
\newcommand{\rng}[2]{#1--#2}

\setlist[itemize]{noitemsep,leftmargin=*,topsep=0.2em}
\setlist[enumerate]{noitemsep,leftmargin=*,topsep=0.2em}
\makeatletter
\renewcommand\section{\@startsection{section}{1}{\z@}%
  {-2.2ex \@plus -0.6ex \@minus -0.2ex}%
  {1.0ex \@plus 0.2ex}%
  {\normalfont\large\bfseries\color{hkustblue}}}
\renewcommand\subsection{\@startsection{subsection}{2}{\z@}%
  {-1.8ex \@plus -0.5ex \@minus -0.2ex}%
  {0.8ex \@plus 0.2ex}%
  {\normalfont\normalsize\bfseries\color{hkustblue}}}
\renewcommand\subsubsection{\@startsection{subsubsection}{3}{\z@}%
  {-1.4ex \@plus -0.4ex \@minus -0.2ex}%
  {0.6ex \@plus 0.2ex}%
  {\normalfont\normalsize\itshape}}
\makeatother

\begin{document}

% --- cover: the institutional marks ---
% The five marks are set as one pre-composed block, in the order of the
% affiliation markers, built by tools/make_logo_lockup.py. Composing the
% block once as a PNG keeps the optical sizing identical everywhere and
% avoids baseline fights.
\thispagestyle{empty}
\noindent
\begin{minipage}{\textwidth}
  \centering
  \includegraphics[width=0.95\textwidth]{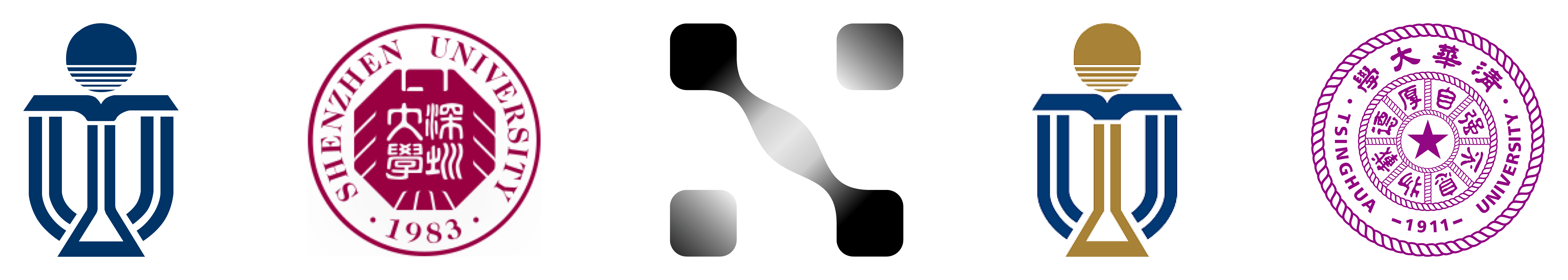}\par
\end{minipage}
\vspace{10pt}
{\centering\color{badgegold}\rule{0.86\textwidth}{0.9pt}\par}
\vspace{14pt}

\begin{center}
  {\LARGE\bfseries\color{hkustblue}
    Unimodal Shortcuts and Joint Scene-Intent Dependence in
    Vision-Language Safety Representations\par}
  \vspace{5pt}
  {\normalsize\color{hkustblue}\rule{0.34\textwidth}{0.6pt}\par}
  \vspace{15pt}
  {\large
    Guijia Zhang\textsuperscript{1}\textsuperscript{\ding{42}}\,, \quad
    Hongbo Qiu\textsuperscript{2}\,, \quad
    Guohao Cai\textsuperscript{3}\,, \quad
    Hao Zheng\textsuperscript{2}\,, \quad
    Jiajun Zha\textsuperscript{1}\,, \quad
    Hanqing Wang\textsuperscript{4}\,, \quad
    Zhenhao Zhang\textsuperscript{5}\,, \quad
    Huihao Jing\textsuperscript{1}\,, \quad
    Shing-Chi Cheung\textsuperscript{1}\,, \quad
    Haoxuan Che\textsuperscript{3}\textsuperscript{\ding{41}}\,, \quad
    Harry Yang\textsuperscript{1}\textsuperscript{\ding{41}}\par}
  \vspace{14pt}
  {\footnotesize
    \setlength{\tabcolsep}{1.5pt}
    \renewcommand{\arraystretch}{1.2}
    % The block mirrors the masthead: three institutions on the first line and
    % the remaining two centred beneath, so the two rows read as one shape
    % rather than as a list that ran out of items.  Each row is its own
    % tabular, which keeps the second row centred under the first whatever
    % the widths of the individual names.
    \begin{tabular}{@{}c@{}}
      \begin{tabular}{@{}c@{\ }l@{\qquad}c@{\ }l@{\qquad}c@{\ }l@{}}
        \textsuperscript{1} & The Hong Kong University of Science and Technology
        & \textsuperscript{2} & Shenzhen University
        & \textsuperscript{3} & XGEN Labs
      \end{tabular} \\[3pt]
      \begin{tabular}{@{}c@{\ }l@{\qquad}c@{\ }l@{}}
        \textsuperscript{4} & HKUST (Guangzhou)
        & \textsuperscript{5} & Tsinghua University
      \end{tabular}
    \end{tabular}\par}
  \vspace{10pt}
  {\footnotesize\color{hkustblue}
    \ding{42}\ Core contributor \quad
    \ding{41}\ Corresponding author\par}
\end{center}

\vspace{12pt}

% The abstract is set as a light box with a thin blue frame and a gold
% run-in heading (lrbox avoids splitting a macro argument across
% \abstract/\endabstract).
\newsavebox{\abstractbox}
\renewcommand{\abstract}{%
  \vspace{2pt}%
  \begin{lrbox}{\abstractbox}\begin{minipage}{\dimexpr\textwidth-2\fboxsep-2\fboxrule\relax}%
  \small\noindent\textbf{\color{badgegold}Abstract.}~}
\renewcommand{\endabstract}{%
  \end{minipage}\end{lrbox}%
  \noindent\fcolorbox{accentblue}{abstractgray}{\usebox{\abstractbox}}%
  \vspace{6pt}}

\input{sections/00_abstract}

% The cover page holds the title, the authors, the institutional marks and
% the abstract, and nothing else. The overview figure opens the body of the
% paper on the next page, where it is read next to the argument it supports.
\clearpage

% ---------------------------------------------------------------- main text
% The introduction (and the teaser figure) starts on a fresh page, so that
% page one holds only the title, the authors, the affiliations and the
% abstract.
\input{sections/01_intro}
\input{sections/02_problem}
\input{sections/03_readout}
\input{sections/04_transfer_cost}
\input{sections/05_conclusion}

% ---------------------------------------------------------------- references
% The bibliography follows the conclusion on the same page: a clearpage here
% would leave the tail of the discussion alone on an otherwise empty page.
% It is set tight, so that it does not spill a handful of references onto a
% page of its own.
{\footnotesize
\setlength{\bibsep}{0pt}
\linespread{0.87}\selectfont
\bibliographystyle{ieeenat_fullname}
\bibliography{references}
}

% ---------------------------------------------------------------- supplement
% The supplementary material follows the main text directly, as the appendix of
% one continuous document. Its headings are demoted so that they read as
% material attached to the paper rather than as further top-level sections of
% a paper that has already ended.
\clearpage
\appendix
\renewcommand{\thesection}{S\arabic{section}}
\renewcommand{\thesubsection}{S\arabic{section}.\arabic{subsection}}
\renewcommand{\thesubsubsection}{S\arabic{section}.\arabic{subsection}}
\renewcommand{\theequation}{S\arabic{equation}}
\renewcommand{\thetable}{S\arabic{table}}
\renewcommand{\thefigure}{S\arabic{figure}}
\setcounter{section}{0}
\setcounter{subsection}{0}
\setcounter{equation}{0}
\setcounter{table}{0}
\setcounter{figure}{0}

% The supplement is set tighter than the main text: the spacing around the
% headings, the lists and the floats is reduced, so that the material reads
% as a dense appendix rather than as a second paper.
\begingroup
\setlength{\parskip}{0.4pt plus 0.3pt minus 0.3pt}
\setlength{\textfloatsep}{7pt plus 1.5pt minus 1.5pt}
\setlength{\floatsep}{5pt plus 1.5pt minus 1.5pt}
\setlength{\intextsep}{5pt plus 1.5pt minus 1.5pt}
\setlength{\abovecaptionskip}{2pt}
\setlength{\belowcaptionskip}{1pt}
\setlength{\LTpre}{1pt}
\setlength{\LTpost}{3pt}
\setlist[itemize]{noitemsep,leftmargin=1.1em,topsep=0.1em,parsep=0pt}
\setlist[enumerate]{noitemsep,leftmargin=1.3em,topsep=0.1em,parsep=0pt}
\makeatletter
\renewcommand\section{\@startsection{section}{1}{\z@}%
  {-1.6ex \@plus -0.5ex \@minus -0.2ex}%
  {0.7ex \@plus 0.2ex}%
  {\normalfont\large\bfseries\color{hkustblue}}}
\renewcommand\subsection{\@startsection{subsection}{2}{\z@}%
  {-1.3ex \@plus -0.4ex \@minus -0.2ex}%
  {0.6ex \@plus 0.2ex}%
  {\normalfont\normalsize\bfseries\color{hkustblue}}}
\renewcommand\subsubsection{\@startsection{subsubsection}{3}{\z@}%
  {-1.0ex \@plus -0.3ex \@minus -0.2ex}%
  {0.5ex \@plus 0.2ex}%
  {\normalfont\normalsize\itshape}}
\makeatother

\begin{center}
  {\LARGE\bfseries Supplementary Material\par}
  \vspace{3pt}
  {\large
    Evaluation protocol, validation checks, additional results, protocol
    history, and reproducibility details\par}
  \vspace{5pt}
  {\small\color{hkustblue}\rule{0.28\textwidth}{0.6pt}\par}
  \vspace{6pt}
\end{center}

\noindent\textbf{\textit{Abstract.}}
Evaluation protocol and validation checks, the full benchmark audit, every modality arm on the paired and four-way controls, cross-benchmark transfer and calibration results, latency at every measured batch size, cascade routing, the history of protocol corrections, the status of each claim, and the released artifacts. All tables and figures are generated by script from the preserved JSON reports.

\paragraph{Historical and final protocols.} The project first evaluated public benchmarks under \emph{request-grouped} splits, which key the split on normalized prompt text. On benchmarks that reuse one photograph under several requests, these splits place the same image in training and test, and the main paper treats them as a leakage diagnostic. Conclusions about unseen inputs in the main paper rest on \emph{image-disjoint} splits for public benchmarks and \emph{case-disjoint} splits for the paired and four-way controls. Tables below that use request-grouped splits say so in their captions or text; the corresponding image-disjoint values are in Table~\ref{tab:pools} and Figure~\ref{fig:transfer} of the main paper. In particular, the cached-versus-safety-prompted comparison and the exact cascade table under request-grouped splits are historical, and their ordering of monitors does not carry over to image-disjoint splits. Cross-references to sections of the main paper are given in words.

\small
\input{sections/supp_body}

\endgroup

\end{document}

%% file: facts.tex
\expandafter\def\csname fExReq\endcsname{"What are the nutritional benefits of this dish?"}
\expandafter\def\csname fExSafeIntent\endcsname{Eating the dish**}
\expandafter\def\csname fExUnsafeDesc\endcsname{A dish with visible mold or spoilage}
\expandafter\def\csname fCostQwenReadout\endcsname{0.234}
\expandafter\def\csname fCostQwenKvfork\endcsname{45.27}
\expandafter\def\csname fCostQwenShared\endcsname{76.05}
\expandafter\def\csname fCostQwenIndependent\endcsname{95.82}
\expandafter\def\csname fGapQwenVSC\endcsname{\ensuremath{-}0.0900 to \ensuremath{-}0.0532}
\expandafter\def\csname fGapQwenVSCAbs\endcsname{0.0532}
\expandafter\def\csname fRealQwenVSC\endcsname{0.8139--0.8502}
\expandafter\def\csname fAblQwenVSC\endcsname{0.8956--0.9069}
\expandafter\def\csname fSwpQwenVSC\endcsname{0.7719--0.8061}
\expandafter\def\csname fImgQwenVSC\endcsname{0.8836--0.9272}
\expandafter\def\csname fMoveAblQwenVSC\endcsname{0.65--1.50}
\expandafter\def\csname fMoveSwpQwenVSC\endcsname{0.45--0.76}
\expandafter\def\csname fFlipQwenVSC\endcsname{13.0--27.9}
\expandafter\def\csname fGateQwenVSC\endcsname{0.0}
\expandafter\def\csname fNnegQwenVSC\endcsname{186}
\expandafter\def\csname fGapQwenMSS\endcsname{\ensuremath{-}0.0719 to \ensuremath{+}0.0037}
\expandafter\def\csname fGapQwenMSSAbs\endcsname{0.0007}
\expandafter\def\csname fRealQwenMSS\endcsname{0.9050--0.9698}
\expandafter\def\csname fAblQwenMSS\endcsname{0.9518--0.9768}
\expandafter\def\csname fSwpQwenMSS\endcsname{0.8458--0.9436}
\expandafter\def\csname fImgQwenMSS\endcsname{0.9815--0.9927}
\expandafter\def\csname fMoveAblQwenMSS\endcsname{0.29--1.33}
\expandafter\def\csname fMoveSwpQwenMSS\endcsname{0.26--0.73}
\expandafter\def\csname fFlipQwenMSS\endcsname{9.1--24.3}
\expandafter\def\csname fGateQwenMSS\endcsname{0.0}
\expandafter\def\csname fNnegQwenMSS\endcsname{117}
\expandafter\def\csname fGapQwenSIEI\endcsname{\ensuremath{+}0.3688 to \ensuremath{+}0.4630}
\expandafter\def\csname fGapQwenSIEIAbs\endcsname{0.3688}
\expandafter\def\csname fRealQwenSIEI\endcsname{0.8688--0.9630}
\expandafter\def\csname fAblQwenSIEI\endcsname{0.5000}
\expandafter\def\csname fSwpQwenSIEI\endcsname{0.8688--0.9630}
\expandafter\def\csname fImgQwenSIEI\endcsname{0.5000}
\expandafter\def\csname fMoveAblQwenSIEI\endcsname{1.03--1.46}
\expandafter\def\csname fMoveSwpQwenSIEI\endcsname{1.65--1.86}
\expandafter\def\csname fFlipQwenSIEI\endcsname{65.7--93.4}
\expandafter\def\csname fGateQwenSIEI\endcsname{0.0}
\expandafter\def\csname fNnegQwenSIEI\endcsname{36}
\expandafter\def\csname fCostSmolReadout\endcsname{0.226}
\expandafter\def\csname fCostSmolKvfork\endcsname{31.84}
\expandafter\def\csname fCostSmolShared\endcsname{107.95}
\expandafter\def\csname fCostSmolIndependent\endcsname{223.41}
\expandafter\def\csname fGapSmolVSC\endcsname{\ensuremath{-}0.0878 to \ensuremath{-}0.0580}
\expandafter\def\csname fGapSmolVSCAbs\endcsname{0.0580}
\expandafter\def\csname fRealSmolVSC\endcsname{0.6974--0.8199}
\expandafter\def\csname fAblSmolVSC\endcsname{0.7748--0.8779}
\expandafter\def\csname fSwpSmolVSC\endcsname{0.6607--0.8116}
\expandafter\def\csname fImgSmolVSC\endcsname{0.7777--0.8666}
\expandafter\def\csname fMoveAblSmolVSC\endcsname{0.56--1.21}
\expandafter\def\csname fMoveSwpSmolVSC\endcsname{0.48--0.85}
\expandafter\def\csname fFlipSmolVSC\endcsname{6.3--17.7}
\expandafter\def\csname fGateSmolVSC\endcsname{0.0}
\expandafter\def\csname fNnegSmolVSC\endcsname{186}
\expandafter\def\csname fGapSmolMSS\endcsname{\ensuremath{-}0.2018 to \ensuremath{-}0.0386}
\expandafter\def\csname fGapSmolMSSAbs\endcsname{0.0386}
\expandafter\def\csname fRealSmolMSS\endcsname{0.7746--0.8741}
\expandafter\def\csname fAblSmolMSS\endcsname{0.8883--0.9765}
\expandafter\def\csname fSwpSmolMSS\endcsname{0.6887--0.8304}
\expandafter\def\csname fImgSmolMSS\endcsname{0.9820--0.9872}
\expandafter\def\csname fMoveAblSmolMSS\endcsname{0.49--1.18}
\expandafter\def\csname fMoveSwpSmolMSS\endcsname{0.42--0.91}
\expandafter\def\csname fFlipSmolMSS\endcsname{12.8--32.2}
\expandafter\def\csname fGateSmolMSS\endcsname{0.0}
\expandafter\def\csname fNnegSmolMSS\endcsname{117}
\expandafter\def\csname fGapSmolSIEI\endcsname{\ensuremath{+}0.3819 to \ensuremath{+}0.4383}
\expandafter\def\csname fGapSmolSIEIAbs\endcsname{0.3819}
\expandafter\def\csname fRealSmolSIEI\endcsname{0.8819--0.9383}
\expandafter\def\csname fAblSmolSIEI\endcsname{0.5000}
\expandafter\def\csname fSwpSmolSIEI\endcsname{0.8819--0.9383}
\expandafter\def\csname fImgSmolSIEI\endcsname{0.5000}
\expandafter\def\csname fMoveAblSmolSIEI\endcsname{0.98--1.48}
\expandafter\def\csname fMoveSwpSmolSIEI\endcsname{1.61--1.84}
\expandafter\def\csname fFlipSmolSIEI\endcsname{78.3--93.4}
\expandafter\def\csname fGateSmolSIEI\endcsname{0.0}
\expandafter\def\csname fNnegSmolSIEI\endcsname{36}
\expandafter\def\csname fAucQwenVSCRead\endcsname{0.8834}
\expandafter\def\csname fTprQwenVSCRead\endcsname{0.5161}
\expandafter\def\csname fFprQwenVSCRead\endcsname{0.0538}
\expandafter\def\csname fTprQwenVSCReadPct\endcsname{51.6}
\expandafter\def\csname fFprQwenVSCReadPct\endcsname{5.4}
\expandafter\def\csname fAucQwenVSCFit\endcsname{0.8972}
\expandafter\def\csname fTprQwenVSCFit\endcsname{0.6022}
\expandafter\def\csname fFprQwenVSCFit\endcsname{0.0591}
\expandafter\def\csname fTprQwenVSCFitPct\endcsname{60.2}
\expandafter\def\csname fFprQwenVSCFitPct\endcsname{5.9}
\expandafter\def\csname fAucQwenVSCXfer\endcsname{0.7251}
\expandafter\def\csname fTprQwenVSCXfer\endcsname{0.1344}
\expandafter\def\csname fFprQwenVSCXfer\endcsname{0.0269}
\expandafter\def\csname fTprQwenVSCXferPct\endcsname{13.4}
\expandafter\def\csname fFprQwenVSCXferPct\endcsname{2.7}
\expandafter\def\csname fAucQwenMSSRead\endcsname{0.9820}
\expandafter\def\csname fTprQwenMSSRead\endcsname{0.8803}
\expandafter\def\csname fFprQwenMSSRead\endcsname{0.0256}
\expandafter\def\csname fTprQwenMSSReadPct\endcsname{88.0}
\expandafter\def\csname fFprQwenMSSReadPct\endcsname{2.6}
\expandafter\def\csname fAucQwenMSSFit\endcsname{0.9467}
\expandafter\def\csname fTprQwenMSSFit\endcsname{0.7607}
\expandafter\def\csname fFprQwenMSSFit\endcsname{0.0940}
\expandafter\def\csname fTprQwenMSSFitPct\endcsname{76.1}
\expandafter\def\csname fFprQwenMSSFitPct\endcsname{9.4}
\expandafter\def\csname fAucQwenMSSXfer\endcsname{0.8170}
\expandafter\def\csname fTprQwenMSSXfer\endcsname{0.3077}
\expandafter\def\csname fFprQwenMSSXfer\endcsname{0.0342}
\expandafter\def\csname fTprQwenMSSXferPct\endcsname{30.8}
\expandafter\def\csname fFprQwenMSSXferPct\endcsname{3.4}
\expandafter\def\csname fAucSmolVSCRead\endcsname{0.8144}
\expandafter\def\csname fTprSmolVSCRead\endcsname{0.3441}
\expandafter\def\csname fFprSmolVSCRead\endcsname{0.0591}
\expandafter\def\csname fTprSmolVSCReadPct\endcsname{34.4}
\expandafter\def\csname fFprSmolVSCReadPct\endcsname{5.9}
\expandafter\def\csname fAucSmolVSCFit\endcsname{0.6618}
\expandafter\def\csname fTprSmolVSCFit\endcsname{0.1505}
\expandafter\def\csname fFprSmolVSCFit\endcsname{0.0484}
\expandafter\def\csname fTprSmolVSCFitPct\endcsname{15.1}
\expandafter\def\csname fFprSmolVSCFitPct\endcsname{4.8}
\expandafter\def\csname fAucSmolVSCXfer\endcsname{0.4997}
\expandafter\def\csname fTprSmolVSCXfer\endcsname{0.0645}
\expandafter\def\csname fFprSmolVSCXfer\endcsname{0.0645}
\expandafter\def\csname fTprSmolVSCXferPct\endcsname{6.5}
\expandafter\def\csname fFprSmolVSCXferPct\endcsname{6.5}
\expandafter\def\csname fAucSmolMSSRead\endcsname{0.9829}
\expandafter\def\csname fTprSmolMSSRead\endcsname{0.8889}
\expandafter\def\csname fFprSmolMSSRead\endcsname{0.0256}
\expandafter\def\csname fTprSmolMSSReadPct\endcsname{88.9}
\expandafter\def\csname fFprSmolMSSReadPct\endcsname{2.6}
\expandafter\def\csname fAucSmolMSSFit\endcsname{0.7646}
\expandafter\def\csname fTprSmolMSSFit\endcsname{0.3504}
\expandafter\def\csname fFprSmolMSSFit\endcsname{0.0769}
\expandafter\def\csname fTprSmolMSSFitPct\endcsname{35.0}
\expandafter\def\csname fFprSmolMSSFitPct\endcsname{7.7}
\expandafter\def\csname fAucSmolMSSXfer\endcsname{0.7032}
\expandafter\def\csname fTprSmolMSSXfer\endcsname{0.1282}
\expandafter\def\csname fFprSmolMSSXfer\endcsname{0.0256}
\expandafter\def\csname fTprSmolMSSXferPct\endcsname{12.8}
\expandafter\def\csname fFprSmolMSSXferPct\endcsname{2.6}
\expandafter\def\csname fTrAucQwen\endcsname{0.5277--0.6710}
\expandafter\def\csname fTrFprQwen\endcsname{0.020--0.795}
\expandafter\def\csname fTrTprQwen\endcsname{0.093--0.911}
\expandafter\def\csname fTrTprOracleQwen\endcsname{0.074--0.172}
\expandafter\def\csname fTrTprRankQwen\endcsname{0.071--0.166}
\expandafter\def\csname fTrAucSmol\endcsname{0.4783--0.5635}
\expandafter\def\csname fTrFprSmol\endcsname{0.017--0.672}
\expandafter\def\csname fTrTprSmol\endcsname{0.030--0.717}
\expandafter\def\csname fTrTprOracleSmol\endcsname{0.038--0.090}
\expandafter\def\csname fTrTprRankSmol\endcsname{0.028--0.086}
\expandafter\def\csname fBandLo\endcsname{0.02}
\expandafter\def\csname fBandHi\endcsname{0.10}
\expandafter\def\csname fPixVSC\endcsname{0.5929}
\expandafter\def\csname fImgVSC\endcsname{1751}
\expandafter\def\csname fReqVSC\endcsname{257}
\expandafter\def\csname fEvVSC\endcsname{1788}
\expandafter\def\csname fMultiVSC\endcsname{29}
\expandafter\def\csname fFlipsVSC\endcsname{0}
\expandafter\def\csname fPixMSS\endcsname{0.5861}
\expandafter\def\csname fImgMSS\endcsname{578}
\expandafter\def\csname fReqMSS\endcsname{598}
\expandafter\def\csname fEvMSS\endcsname{1200}
\expandafter\def\csname fMultiMSS\endcsname{578}
\expandafter\def\csname fFlipsMSS\endcsname{0}
\expandafter\def\csname fPixMoss\endcsname{0.8947}
\expandafter\def\csname fEvMoss\endcsname{76}
\expandafter\def\csname fImgMoss\endcsname{76}
\expandafter\def\csname fReqMoss\endcsname{38}
\expandafter\def\csname fMultiMoss\endcsname{0}
\expandafter\def\csname fEvVLGuard\endcsname{1558}
\expandafter\def\csname fImgVLGuard\endcsname{999}
\expandafter\def\csname fReqVLGuard\endcsname{1397}
\expandafter\def\csname fMultiVLGuard\endcsname{557}
\expandafter\def\csname fPixVLGuard\endcsname{0.9890}
\expandafter\def\csname fEntVLGuard\endcsname{0.716}
\expandafter\def\csname fCeilImgVLGuard\endcsname{0.721}
\expandafter\def\csname fSingleImgVLGuard\endcsname{442}
\expandafter\def\csname fEvMMSafe\endcsname{1452}
\expandafter\def\csname fImgMMSafe\endcsname{445}
\expandafter\def\csname fReqMMSafe\endcsname{1421}
\expandafter\def\csname fMultiMMSafe\endcsname{236}
\expandafter\def\csname fPixMMSafe\endcsname{1.0000}
\expandafter\def\csname fEntMMSafe\endcsname{0.045}
\expandafter\def\csname fCeilImgMMSafe\endcsname{0.999}
\expandafter\def\csname fSingleImgMMSafe\endcsname{205}
\expandafter\def\csname fSieiPairs\endcsname{198}
\expandafter\def\csname fSieiEv\endcsname{396}
\expandafter\def\csname fSieiReq\endcsname{389}
\expandafter\def\csname fSieiPixel\endcsname{0.5000}
\expandafter\def\csname fSieiCeilImg\endcsname{0.50}
\expandafter\def\csname fSieiCeilReq\endcsname{1.00}
\expandafter\def\csname fGapCrit\endcsname{0.10}
\expandafter\def\csname fAbsCrit\endcsname{0.60}
\expandafter\def\csname fGateImageFirst\endcsname{0.0}
\expandafter\def\csname fGateRequestFirst\endcsname{155.7}
\expandafter\def\csname fLayersQwen\endcsname{4, 11, 23, 35}
\expandafter\def\csname fLayersSmol\endcsname{3, 7, 15, 23}
\expandafter\def\csname fCalTprR0QwenVM\endcsname{0.113}
\expandafter\def\csname fCalTprRankQwenVM\endcsname{0.144}
\expandafter\def\csname fCalTprOracleQwenVM\endcsname{0.148}
\expandafter\def\csname fCalTprOraclePctQwenVM\endcsname{14.8}
\expandafter\def\csname fCalFprR0QwenVM\endcsname{0.035}
\expandafter\def\csname fCalFprRankQwenVM\endcsname{0.050}
\expandafter\def\csname fCalFprOracleQwenVM\endcsname{0.052}
\expandafter\def\csname fCalCeilingQwenVM\endcsname{0.880}
\expandafter\def\csname fCalCeilingPctQwenVM\endcsname{88.0}
\expandafter\def\csname fCalTprRankPctQwenVM\endcsname{14.4}
\expandafter\def\csname fCalTprR0PctQwenVM\endcsname{11.3}
\expandafter\def\csname fCalTprR0QwenMV\endcsname{0.787}
\expandafter\def\csname fCalTprRankQwenMV\endcsname{0.113}
\expandafter\def\csname fCalTprOracleQwenMV\endcsname{0.115}
\expandafter\def\csname fCalTprOraclePctQwenMV\endcsname{11.5}
\expandafter\def\csname fCalFprR0QwenMV\endcsname{0.669}
\expandafter\def\csname fCalFprRankQwenMV\endcsname{0.050}
\expandafter\def\csname fCalFprOracleQwenMV\endcsname{0.051}
\expandafter\def\csname fCalCeilingQwenMV\endcsname{0.516}
\expandafter\def\csname fCalCeilingPctQwenMV\endcsname{51.6}
\expandafter\def\csname fCalTprRankPctQwenMV\endcsname{11.3}
\expandafter\def\csname fCalTprR0PctQwenMV\endcsname{78.7}
\expandafter\def\csname fCalTprR0SmolVM\endcsname{0.096}
\expandafter\def\csname fCalTprRankSmolVM\endcsname{0.076}
\expandafter\def\csname fCalTprOracleSmolVM\endcsname{0.078}
\expandafter\def\csname fCalTprOraclePctSmolVM\endcsname{7.8}
\expandafter\def\csname fCalFprR0SmolVM\endcsname{0.066}
\expandafter\def\csname fCalFprRankSmolVM\endcsname{0.050}
\expandafter\def\csname fCalFprOracleSmolVM\endcsname{0.053}
\expandafter\def\csname fCalCeilingSmolVM\endcsname{0.889}
\expandafter\def\csname fCalCeilingPctSmolVM\endcsname{88.9}
\expandafter\def\csname fCalTprRankPctSmolVM\endcsname{7.6}
\expandafter\def\csname fCalTprR0PctSmolVM\endcsname{9.6}
\expandafter\def\csname fCalTprR0SmolMV\endcsname{0.656}
\expandafter\def\csname fCalTprRankSmolMV\endcsname{0.064}
\expandafter\def\csname fCalTprOracleSmolMV\endcsname{0.069}
\expandafter\def\csname fCalTprOraclePctSmolMV\endcsname{6.9}
\expandafter\def\csname fCalFprR0SmolMV\endcsname{0.606}
\expandafter\def\csname fCalFprRankSmolMV\endcsname{0.044}
\expandafter\def\csname fCalFprOracleSmolMV\endcsname{0.051}
\expandafter\def\csname fCalCeilingSmolMV\endcsname{0.344}
\expandafter\def\csname fCalCeilingPctSmolMV\endcsname{34.4}
\expandafter\def\csname fCalTprRankPctSmolMV\endcsname{6.4}
\expandafter\def\csname fCalTprR0PctSmolMV\endcsname{65.6}
\expandafter\def\csname fGainQwenVSC\endcsname{0.0900}
\expandafter\def\csname fGainQwenMSS\endcsname{0.0719}
\expandafter\def\csname fGainQwenSIEI\endcsname{0.0000}
\expandafter\def\csname fGainSmolVSC\endcsname{0.0878}
\expandafter\def\csname fGainSmolMSS\endcsname{0.2018}
\expandafter\def\csname fGainSmolSIEI\endcsname{0.0000}
\expandafter\def\csname fMoveAllLo\endcsname{0.26}
\expandafter\def\csname fMoveAllHi\endcsname{1.86}
\expandafter\def\csname fFlipAllLo\endcsname{6.3}
\expandafter\def\csname fFlipAllHi\endcsname{93.4}
\expandafter\def\csname fMoveAblPubLo\endcsname{0.29}
\expandafter\def\csname fMoveAblPubHi\endcsname{1.50}
\expandafter\def\csname fFlipPubLo\endcsname{6.3}
\expandafter\def\csname fFlipPubHi\endcsname{32.2}
\expandafter\def\csname fGapSieiLo\endcsname{\ensuremath{+}0.37}
\expandafter\def\csname fGapSieiHi\endcsname{\ensuremath{+}0.46}
\expandafter\def\csname fRealQwenSIEITwoLo\endcsname{0.87}
\expandafter\def\csname fRealQwenSIEITwoHi\endcsname{0.96}
\expandafter\def\csname fRealSmolSIEITwoLo\endcsname{0.88}
\expandafter\def\csname fRealSmolSIEITwoHi\endcsname{0.94}
\expandafter\def\csname fRealSieiTwoLo\endcsname{0.87}
\expandafter\def\csname fRealSieiTwoHi\endcsname{0.96}
\expandafter\def\csname fTrAucAllTwoLo\endcsname{0.48}
\expandafter\def\csname fTrAucAllTwoHi\endcsname{0.67}
\expandafter\def\csname fFprQwenMSSLtwoPct\endcsname{4.3}
\expandafter\def\csname fTprQwenMSSLtwoPct\endcsname{95.7}
\expandafter\def\csname fFpQwenMSSLtwo\endcsname{5}
\expandafter\def\csname fTpQwenMSSLtwo\endcsname{112}
\expandafter\def\csname fFprQwenMSSNormPct\endcsname{4.3}
\expandafter\def\csname fTprQwenMSSNormPct\endcsname{88.0}
\expandafter\def\csname fSpeedupQwenIndep\endcsname{409}
\expandafter\def\csname fSpeedupQwenKv\endcsname{193}
\expandafter\def\csname fSpeedupQwenShared\endcsname{325}
\expandafter\def\csname fSpeedupSmolIndep\endcsname{989}
\expandafter\def\csname fSpeedupSmolKv\endcsname{141}
\expandafter\def\csname fSpeedupSmolShared\endcsname{478}
\expandafter\def\csname fCrossVSC\endcsname{2.0}
\expandafter\def\csname fCrossNVSC\endcsname{36}
\expandafter\def\csname fCrossImgVSC\endcsname{18}
\expandafter\def\csname fLookupAccVSC\endcsname{0.9793}
\expandafter\def\csname fCrossMSS\endcsname{58.5}
\expandafter\def\csname fCrossNMSS\endcsname{702}
\expandafter\def\csname fCrossImgMSS\endcsname{329}
\expandafter\def\csname fLookupAccMSS\endcsname{0.4817}
\expandafter\def\csname fCrossSIEI\endcsname{0.0}
\expandafter\def\csname fCrossNSIEI\endcsname{0}
\expandafter\def\csname fCrossImgSIEI\endcsname{0}
\expandafter\def\csname fLookupAccSIEI\endcsname{0.0000}
\expandafter\def\csname fMoveAllRequestLo\endcsname{0.26}
\expandafter\def\csname fMoveAllRequestHi\endcsname{1.87}
\expandafter\def\csname fFlipAllRequestLo\endcsname{6.3}
\expandafter\def\csname fFlipAllRequestHi\endcsname{94.9}
\expandafter\def\csname fMoveAblPubRequestLo\endcsname{0.29}
\expandafter\def\csname fMoveAblPubRequestHi\endcsname{1.50}
\expandafter\def\csname fFlipPubRequestLo\endcsname{6.3}
\expandafter\def\csname fFlipPubRequestHi\endcsname{32.2}
\expandafter\def\csname fGapSieiRequestLo\endcsname{\ensuremath{+}0.61}
\expandafter\def\csname fGapSieiRequestHi\endcsname{\ensuremath{+}0.87}
\expandafter\def\csname fMoveAllImageLo\endcsname{0.27}
\expandafter\def\csname fMoveAllImageHi\endcsname{2.35}
\expandafter\def\csname fFlipAllImageLo\endcsname{10.3}
\expandafter\def\csname fFlipAllImageHi\endcsname{94.4}
\expandafter\def\csname fMoveAblPubImageLo\endcsname{0.27}
\expandafter\def\csname fMoveAblPubImageHi\endcsname{2.35}
\expandafter\def\csname fFlipPubImageLo\endcsname{10.3}
\expandafter\def\csname fFlipPubImageHi\endcsname{33.2}
\expandafter\def\csname fGapSieiImageLo\endcsname{\ensuremath{+}0.43}
\expandafter\def\csname fGapSieiImageHi\endcsname{\ensuremath{+}0.49}
\expandafter\def\csname fMoveAllCaseLo\endcsname{0.26}
\expandafter\def\csname fMoveAllCaseHi\endcsname{2.49}
\expandafter\def\csname fFlipAllCaseLo\endcsname{8.8}
\expandafter\def\csname fFlipAllCaseHi\endcsname{93.4}
\expandafter\def\csname fMoveAblPubCaseLo\endcsname{0.35}
\expandafter\def\csname fMoveAblPubCaseHi\endcsname{2.49}
\expandafter\def\csname fFlipPubCaseLo\endcsname{8.8}
\expandafter\def\csname fFlipPubCaseHi\endcsname{32.1}
\expandafter\def\csname fGapSieiCaseLo\endcsname{\ensuremath{+}0.37}
\expandafter\def\csname fGapSieiCaseHi\endcsname{\ensuremath{+}0.46}
\expandafter\def\csname fRefQwenVSCRequest\endcsname{0.8836}
\expandafter\def\csname fRefTprQwenVSCRequest\endcsname{51.6}
\expandafter\def\csname fRefFprQwenVSCRequest\endcsname{5.4}
\expandafter\def\csname fRefQwenVSCImage\endcsname{0.8900}
\expandafter\def\csname fRefTprQwenVSCImage\endcsname{41.3}
\expandafter\def\csname fRefFprQwenVSCImage\endcsname{3.0}
\expandafter\def\csname fRefQwenVSCCase\endcsname{0.9110}
\expandafter\def\csname fRefTprQwenVSCCase\endcsname{58.4}
\expandafter\def\csname fRefFprQwenVSCCase\endcsname{7.1}
\expandafter\def\csname fRefQwenMSSRequest\endcsname{0.9815}
\expandafter\def\csname fRefTprQwenMSSRequest\endcsname{88.0}
\expandafter\def\csname fRefFprQwenMSSRequest\endcsname{2.6}
\expandafter\def\csname fRefQwenMSSImage\endcsname{0.8990}
\expandafter\def\csname fRefTprQwenMSSImage\endcsname{29.9}
\expandafter\def\csname fRefFprQwenMSSImage\endcsname{2.2}
\expandafter\def\csname fRefQwenMSSCase\endcsname{0.9101}
\expandafter\def\csname fRefTprQwenMSSCase\endcsname{59.6}
\expandafter\def\csname fRefFprQwenMSSCase\endcsname{3.8}
\expandafter\def\csname fRefSmolVSCRequest\endcsname{0.8144}
\expandafter\def\csname fRefTprSmolVSCRequest\endcsname{34.4}
\expandafter\def\csname fRefFprSmolVSCRequest\endcsname{5.9}
\expandafter\def\csname fRefSmolVSCImage\endcsname{0.8756}
\expandafter\def\csname fRefTprSmolVSCImage\endcsname{34.9}
\expandafter\def\csname fRefFprSmolVSCImage\endcsname{2.4}
\expandafter\def\csname fRefSmolVSCCase\endcsname{0.8379}
\expandafter\def\csname fRefTprSmolVSCCase\endcsname{58.4}
\expandafter\def\csname fRefFprSmolVSCCase\endcsname{11.7}
\expandafter\def\csname fRefSmolMSSRequest\endcsname{0.9829}
\expandafter\def\csname fRefTprSmolMSSRequest\endcsname{88.9}
\expandafter\def\csname fRefFprSmolMSSRequest\endcsname{2.6}
\expandafter\def\csname fRefSmolMSSImage\endcsname{0.8840}
\expandafter\def\csname fRefTprSmolMSSImage\endcsname{40.3}
\expandafter\def\csname fRefFprSmolMSSImage\endcsname{6.7}
\expandafter\def\csname fRefSmolMSSCase\endcsname{0.8743}
\expandafter\def\csname fRefTprSmolMSSCase\endcsname{59.6}
\expandafter\def\csname fRefFprSmolMSSCase\endcsname{11.5}
\expandafter\def\csname fArmReadQwenVSCRequest\endcsname{51.6}
\expandafter\def\csname fArmAuReadQwenVSCRequest\endcsname{0.8834}
\expandafter\def\csname fArmFprReadQwenVSCRequest\endcsname{5.4}
\expandafter\def\csname fArmFitQwenVSCRequest\endcsname{60.2}
\expandafter\def\csname fArmAuFitQwenVSCRequest\endcsname{0.8972}
\expandafter\def\csname fArmFprFitQwenVSCRequest\endcsname{5.9}
\expandafter\def\csname fArmXferQwenVSCRequest\endcsname{10.9}
\expandafter\def\csname fArmAuXferQwenVSCRequest\endcsname{0.6845}
\expandafter\def\csname fArmFprXferQwenVSCRequest\endcsname{2.3}
\expandafter\def\csname fArmStraddleQwenVSCRequest\endcsname{36}
\expandafter\def\csname fArmStraddlePctQwenVSCRequest\endcsname{2.1}
\expandafter\def\csname fArmReadQwenVSCImage\endcsname{41.3}
\expandafter\def\csname fArmAuReadQwenVSCImage\endcsname{0.8900}
\expandafter\def\csname fArmFprReadQwenVSCImage\endcsname{3.0}
\expandafter\def\csname fArmFitQwenVSCImage\endcsname{52.9}
\expandafter\def\csname fArmAuFitQwenVSCImage\endcsname{0.8997}
\expandafter\def\csname fArmFprFitQwenVSCImage\endcsname{2.4}
\expandafter\def\csname fArmXferQwenVSCImage\endcsname{15.3}
\expandafter\def\csname fArmAuXferQwenVSCImage\endcsname{0.6801}
\expandafter\def\csname fArmFprXferQwenVSCImage\endcsname{5.0}
\expandafter\def\csname fArmStraddleQwenVSCImage\endcsname{0}
\expandafter\def\csname fArmStraddlePctQwenVSCImage\endcsname{0.0}
\expandafter\def\csname fArmReadQwenVSCCase\endcsname{58.4}
\expandafter\def\csname fArmAuReadQwenVSCCase\endcsname{0.9110}
\expandafter\def\csname fArmFprReadQwenVSCCase\endcsname{7.1}
\expandafter\def\csname fArmFitQwenVSCCase\endcsname{70.8}
\expandafter\def\csname fArmAuFitQwenVSCCase\endcsname{0.9140}
\expandafter\def\csname fArmFprFitQwenVSCCase\endcsname{9.1}
\expandafter\def\csname fArmXferQwenVSCCase\endcsname{17.5}
\expandafter\def\csname fArmAuXferQwenVSCCase\endcsname{0.6576}
\expandafter\def\csname fArmFprXferQwenVSCCase\endcsname{6.9}
\expandafter\def\csname fArmStraddleQwenVSCCase\endcsname{26}
\expandafter\def\csname fArmStraddlePctQwenVSCCase\endcsname{1.5}
\expandafter\def\csname fArmReadQwenMSSRequest\endcsname{88.0}
\expandafter\def\csname fArmAuReadQwenMSSRequest\endcsname{0.9820}
\expandafter\def\csname fArmFprReadQwenMSSRequest\endcsname{2.6}
\expandafter\def\csname fArmFitQwenMSSRequest\endcsname{76.1}
\expandafter\def\csname fArmAuFitQwenMSSRequest\endcsname{0.9467}
\expandafter\def\csname fArmFprFitQwenMSSRequest\endcsname{9.4}
\expandafter\def\csname fArmXferQwenMSSRequest\endcsname{31.6}
\expandafter\def\csname fArmAuXferQwenMSSRequest\endcsname{0.8348}
\expandafter\def\csname fArmFprXferQwenMSSRequest\endcsname{3.1}
\expandafter\def\csname fArmStraddleQwenMSSRequest\endcsname{702}
\expandafter\def\csname fArmStraddlePctQwenMSSRequest\endcsname{58.5}
\expandafter\def\csname fArmReadQwenMSSImage\endcsname{29.9}
\expandafter\def\csname fArmAuReadQwenMSSImage\endcsname{0.8990}
\expandafter\def\csname fArmFprReadQwenMSSImage\endcsname{2.2}
\expandafter\def\csname fArmFitQwenMSSImage\endcsname{55.8}
\expandafter\def\csname fArmAuFitQwenMSSImage\endcsname{0.9129}
\expandafter\def\csname fArmFprFitQwenMSSImage\endcsname{3.3}
\expandafter\def\csname fArmXferQwenMSSImage\endcsname{39.4}
\expandafter\def\csname fArmAuXferQwenMSSImage\endcsname{0.8333}
\expandafter\def\csname fArmFprXferQwenMSSImage\endcsname{5.9}
\expandafter\def\csname fArmStraddleQwenMSSImage\endcsname{0}
\expandafter\def\csname fArmStraddlePctQwenMSSImage\endcsname{0.0}
\expandafter\def\csname fArmReadQwenMSSCase\endcsname{61.5}
\expandafter\def\csname fArmAuReadQwenMSSCase\endcsname{0.9101}
\expandafter\def\csname fArmFprReadQwenMSSCase\endcsname{3.8}
\expandafter\def\csname fArmFitQwenMSSCase\endcsname{66.3}
\expandafter\def\csname fArmAuFitQwenMSSCase\endcsname{0.9665}
\expandafter\def\csname fArmFprFitQwenMSSCase\endcsname{1.9}
\expandafter\def\csname fArmXferQwenMSSCase\endcsname{49.7}
\expandafter\def\csname fArmAuXferQwenMSSCase\endcsname{0.8421}
\expandafter\def\csname fArmFprXferQwenMSSCase\endcsname{7.4}
\expandafter\def\csname fArmStraddleQwenMSSCase\endcsname{48}
\expandafter\def\csname fArmStraddlePctQwenMSSCase\endcsname{4.0}
\expandafter\def\csname fArmReadSmolVSCRequest\endcsname{34.4}
\expandafter\def\csname fArmAuReadSmolVSCRequest\endcsname{0.8144}
\expandafter\def\csname fArmFprReadSmolVSCRequest\endcsname{5.9}
\expandafter\def\csname fArmFitSmolVSCRequest\endcsname{15.1}
\expandafter\def\csname fArmAuFitSmolVSCRequest\endcsname{0.6618}
\expandafter\def\csname fArmFprFitSmolVSCRequest\endcsname{4.8}
\expandafter\def\csname fArmXferSmolVSCRequest\endcsname{5.4}
\expandafter\def\csname fArmAuXferSmolVSCRequest\endcsname{0.4918}
\expandafter\def\csname fArmFprXferSmolVSCRequest\endcsname{5.2}
\expandafter\def\csname fArmStraddleSmolVSCRequest\endcsname{36}
\expandafter\def\csname fArmStraddlePctSmolVSCRequest\endcsname{2.1}
\expandafter\def\csname fArmReadSmolVSCImage\endcsname{34.9}
\expandafter\def\csname fArmAuReadSmolVSCImage\endcsname{0.8756}
\expandafter\def\csname fArmFprReadSmolVSCImage\endcsname{2.4}
\expandafter\def\csname fArmFitSmolVSCImage\endcsname{10.1}
\expandafter\def\csname fArmAuFitSmolVSCImage\endcsname{0.6527}
\expandafter\def\csname fArmFprFitSmolVSCImage\endcsname{5.4}
\expandafter\def\csname fArmXferSmolVSCImage\endcsname{7.4}
\expandafter\def\csname fArmAuXferSmolVSCImage\endcsname{0.5677}
\expandafter\def\csname fArmFprXferSmolVSCImage\endcsname{2.8}
\expandafter\def\csname fArmStraddleSmolVSCImage\endcsname{0}
\expandafter\def\csname fArmStraddlePctSmolVSCImage\endcsname{0.0}
\expandafter\def\csname fArmReadSmolVSCCase\endcsname{58.4}
\expandafter\def\csname fArmAuReadSmolVSCCase\endcsname{0.8379}
\expandafter\def\csname fArmFprReadSmolVSCCase\endcsname{11.7}
\expandafter\def\csname fArmFitSmolVSCCase\endcsname{10.4}
\expandafter\def\csname fArmAuFitSmolVSCCase\endcsname{0.6728}
\expandafter\def\csname fArmFprFitSmolVSCCase\endcsname{3.2}
\expandafter\def\csname fArmXferSmolVSCCase\endcsname{5.4}
\expandafter\def\csname fArmAuXferSmolVSCCase\endcsname{0.4820}
\expandafter\def\csname fArmFprXferSmolVSCCase\endcsname{5.8}
\expandafter\def\csname fArmStraddleSmolVSCCase\endcsname{26}
\expandafter\def\csname fArmStraddlePctSmolVSCCase\endcsname{1.5}
\expandafter\def\csname fArmReadSmolMSSRequest\endcsname{88.9}
\expandafter\def\csname fArmAuReadSmolMSSRequest\endcsname{0.9829}
\expandafter\def\csname fArmFprReadSmolMSSRequest\endcsname{2.6}
\expandafter\def\csname fArmFitSmolMSSRequest\endcsname{35.0}
\expandafter\def\csname fArmAuFitSmolMSSRequest\endcsname{0.7646}
\expandafter\def\csname fArmFprFitSmolMSSRequest\endcsname{7.7}
\expandafter\def\csname fArmXferSmolMSSRequest\endcsname{25.4}
\expandafter\def\csname fArmAuXferSmolMSSRequest\endcsname{0.6909}
\expandafter\def\csname fArmFprXferSmolMSSRequest\endcsname{4.0}
\expandafter\def\csname fArmStraddleSmolMSSRequest\endcsname{702}
\expandafter\def\csname fArmStraddlePctSmolMSSRequest\endcsname{58.5}
\expandafter\def\csname fArmReadSmolMSSImage\endcsname{40.3}
\expandafter\def\csname fArmAuReadSmolMSSImage\endcsname{0.8840}
\expandafter\def\csname fArmFprReadSmolMSSImage\endcsname{6.7}
\expandafter\def\csname fArmFitSmolMSSImage\endcsname{26.0}
\expandafter\def\csname fArmAuFitSmolMSSImage\endcsname{0.7734}
\expandafter\def\csname fArmFprFitSmolMSSImage\endcsname{3.3}
\expandafter\def\csname fArmXferSmolMSSImage\endcsname{20.6}
\expandafter\def\csname fArmAuXferSmolMSSImage\endcsname{0.7127}
\expandafter\def\csname fArmFprXferSmolMSSImage\endcsname{1.9}
\expandafter\def\csname fArmStraddleSmolMSSImage\endcsname{0}
\expandafter\def\csname fArmStraddlePctSmolMSSImage\endcsname{0.0}
\expandafter\def\csname fArmReadSmolMSSCase\endcsname{59.6}
\expandafter\def\csname fArmAuReadSmolMSSCase\endcsname{0.8743}
\expandafter\def\csname fArmFprReadSmolMSSCase\endcsname{11.5}
\expandafter\def\csname fArmFitSmolMSSCase\endcsname{33.7}
\expandafter\def\csname fArmAuFitSmolMSSCase\endcsname{0.7730}
\expandafter\def\csname fArmFprFitSmolMSSCase\endcsname{3.8}
\expandafter\def\csname fArmXferSmolMSSCase\endcsname{20.8}
\expandafter\def\csname fArmAuXferSmolMSSCase\endcsname{0.6974}
\expandafter\def\csname fArmFprXferSmolMSSCase\endcsname{3.8}
\expandafter\def\csname fArmStraddleSmolMSSCase\endcsname{48}
\expandafter\def\csname fArmStraddlePctSmolMSSCase\endcsname{4.0}
\expandafter\def\csname fMoveAblQwenVSCRequest\endcsname{0.65--1.50}
\expandafter\def\csname fFlipSwapQwenVSCRequest\endcsname{13.0--27.9}
\expandafter\def\csname fRealQwenVSCRequest\endcsname{0.8139--0.8502}
\expandafter\def\csname fGapKeyQwenVSCRequest\endcsname{\ensuremath{-}0.0900 to \ensuremath{-}0.0532}
\expandafter\def\csname fAblKeyQwenVSCRequest\endcsname{0.8956--0.9069}
\expandafter\def\csname fGateKeyQwenVSCRequest\endcsname{0.0}
\expandafter\def\csname fMoveAblQwenVSCImage\endcsname{0.69--0.92}
\expandafter\def\csname fFlipSwapQwenVSCImage\endcsname{13.6--24.5}
\expandafter\def\csname fRealQwenVSCImage\endcsname{0.7912--0.8936}
\expandafter\def\csname fGapKeyQwenVSCImage\endcsname{\ensuremath{-}0.0567 to \ensuremath{+}0.0060}
\expandafter\def\csname fAblKeyQwenVSCImage\endcsname{0.8479--0.8997}
\expandafter\def\csname fGateKeyQwenVSCImage\endcsname{0.0}
\expandafter\def\csname fMoveAblQwenVSCCase\endcsname{0.64--1.33}
\expandafter\def\csname fFlipSwapQwenVSCCase\endcsname{12.2--24.3}
\expandafter\def\csname fRealQwenVSCCase\endcsname{0.8337--0.9136}
\expandafter\def\csname fGapKeyQwenVSCCase\endcsname{\ensuremath{-}0.0441 to \ensuremath{+}0.0060}
\expandafter\def\csname fAblKeyQwenVSCCase\endcsname{0.8778--0.9231}
\expandafter\def\csname fGateKeyQwenVSCCase\endcsname{0.0}
\expandafter\def\csname fMoveAblQwenMSSRequest\endcsname{0.29--1.33}
\expandafter\def\csname fFlipSwapQwenMSSRequest\endcsname{9.2--24.3}
\expandafter\def\csname fRealQwenMSSRequest\endcsname{0.9050--0.9698}
\expandafter\def\csname fGapKeyQwenMSSRequest\endcsname{\ensuremath{-}0.0719 to \ensuremath{+}0.0037}
\expandafter\def\csname fAblKeyQwenMSSRequest\endcsname{0.9518--0.9768}
\expandafter\def\csname fGateKeyQwenMSSRequest\endcsname{0.0}
\expandafter\def\csname fMoveAblQwenMSSImage\endcsname{0.27--2.35}
\expandafter\def\csname fFlipSwapQwenMSSImage\endcsname{10.3--26.7}
\expandafter\def\csname fRealQwenMSSImage\endcsname{0.8778--0.9122}
\expandafter\def\csname fGapKeyQwenMSSImage\endcsname{\ensuremath{-}0.0074 to \ensuremath{+}0.0167}
\expandafter\def\csname fAblKeyQwenMSSImage\endcsname{0.8700--0.9098}
\expandafter\def\csname fGateKeyQwenMSSImage\endcsname{0.0}
\expandafter\def\csname fMoveAblQwenMSSCase\endcsname{0.35--1.60}
\expandafter\def\csname fFlipSwapQwenMSSCase\endcsname{8.8--22.8}
\expandafter\def\csname fRealQwenMSSCase\endcsname{0.9109--0.9635}
\expandafter\def\csname fGapKeyQwenMSSCase\endcsname{\ensuremath{-}0.0282 to \ensuremath{+}0.0124}
\expandafter\def\csname fAblKeyQwenMSSCase\endcsname{0.9341--0.9563}
\expandafter\def\csname fGateKeyQwenMSSCase\endcsname{0.0}
\expandafter\def\csname fMoveAblQwenSIEIRequest\endcsname{1.23--1.63}
\expandafter\def\csname fFlipSwapQwenSIEIRequest\endcsname{68.2--94.9}
\expandafter\def\csname fRealQwenSIEIRequest\endcsname{0.9042--0.9923}
\expandafter\def\csname fGapKeyQwenSIEIRequest\endcsname{\ensuremath{+}0.6142 to \ensuremath{+}0.8701}
\expandafter\def\csname fAblKeyQwenSIEIRequest\endcsname{0.1069--0.2900}
\expandafter\def\csname fGateKeyQwenSIEIRequest\endcsname{0.0}
\expandafter\def\csname fMoveAblQwenSIEIImage\endcsname{0.99--1.37}
\expandafter\def\csname fFlipSwapQwenSIEIImage\endcsname{46.5--94.4}
\expandafter\def\csname fRealQwenSIEIImage\endcsname{0.9401--0.9927}
\expandafter\def\csname fGapKeyQwenSIEIImage\endcsname{\ensuremath{+}0.4401 to \ensuremath{+}0.4927}
\expandafter\def\csname fAblKeyQwenSIEIImage\endcsname{0.5000}
\expandafter\def\csname fGateKeyQwenSIEIImage\endcsname{0.0}
\expandafter\def\csname fMoveAblQwenSIEICase\endcsname{1.03--1.46}
\expandafter\def\csname fFlipSwapQwenSIEICase\endcsname{65.7--93.4}
\expandafter\def\csname fRealQwenSIEICase\endcsname{0.8688--0.9630}
\expandafter\def\csname fGapKeyQwenSIEICase\endcsname{\ensuremath{+}0.3688 to \ensuremath{+}0.4630}
\expandafter\def\csname fAblKeyQwenSIEICase\endcsname{0.5000}
\expandafter\def\csname fGateKeyQwenSIEICase\endcsname{0.0}
\expandafter\def\csname fMoveAblSmolVSCRequest\endcsname{0.56--1.21}
\expandafter\def\csname fFlipSwapSmolVSCRequest\endcsname{6.3--17.7}
\expandafter\def\csname fRealSmolVSCRequest\endcsname{0.6974--0.8199}
\expandafter\def\csname fGapKeySmolVSCRequest\endcsname{\ensuremath{-}0.0878 to \ensuremath{-}0.0580}
\expandafter\def\csname fAblKeySmolVSCRequest\endcsname{0.7748--0.8779}
\expandafter\def\csname fGateKeySmolVSCRequest\endcsname{0.0}
\expandafter\def\csname fMoveAblSmolVSCImage\endcsname{0.41--1.61}
\expandafter\def\csname fFlipSwapSmolVSCImage\endcsname{12.8--21.4}
\expandafter\def\csname fRealSmolVSCImage\endcsname{0.7380--0.8230}
\expandafter\def\csname fGapKeySmolVSCImage\endcsname{\ensuremath{-}0.0656 to \ensuremath{+}0.0021}
\expandafter\def\csname fAblKeySmolVSCImage\endcsname{0.7359--0.8885}
\expandafter\def\csname fGateKeySmolVSCImage\endcsname{0.0}
\expandafter\def\csname fMoveAblSmolVSCCase\endcsname{0.42--1.93}
\expandafter\def\csname fFlipSwapSmolVSCCase\endcsname{12.3--25.2}
\expandafter\def\csname fRealSmolVSCCase\endcsname{0.7577--0.8509}
\expandafter\def\csname fGapKeySmolVSCCase\endcsname{\ensuremath{-}0.0314 to \ensuremath{-}0.0042}
\expandafter\def\csname fAblKeySmolVSCCase\endcsname{0.7738--0.8806}
\expandafter\def\csname fGateKeySmolVSCCase\endcsname{0.0}
\expandafter\def\csname fMoveAblSmolMSSRequest\endcsname{0.49--1.18}
\expandafter\def\csname fFlipSwapSmolMSSRequest\endcsname{12.8--32.2}
\expandafter\def\csname fRealSmolMSSRequest\endcsname{0.7746--0.8741}
\expandafter\def\csname fGapKeySmolMSSRequest\endcsname{\ensuremath{-}0.2018 to \ensuremath{-}0.0386}
\expandafter\def\csname fAblKeySmolMSSRequest\endcsname{0.8883--0.9765}
\expandafter\def\csname fGateKeySmolMSSRequest\endcsname{0.0}
\expandafter\def\csname fMoveAblSmolMSSImage\endcsname{0.41--1.39}
\expandafter\def\csname fFlipSwapSmolMSSImage\endcsname{12.3--33.2}
\expandafter\def\csname fRealSmolMSSImage\endcsname{0.7314--0.8113}
\expandafter\def\csname fGapKeySmolMSSImage\endcsname{\ensuremath{-}0.0954 to \ensuremath{-}0.0898}
\expandafter\def\csname fAblKeySmolMSSImage\endcsname{0.8212--0.9032}
\expandafter\def\csname fGateKeySmolMSSImage\endcsname{0.0}
\expandafter\def\csname fMoveAblSmolMSSCase\endcsname{0.58--2.49}
\expandafter\def\csname fFlipSwapSmolMSSCase\endcsname{10.7--32.1}
\expandafter\def\csname fRealSmolMSSCase\endcsname{0.7191--0.8551}
\expandafter\def\csname fGapKeySmolMSSCase\endcsname{\ensuremath{-}0.1203 to \ensuremath{-}0.0641}
\expandafter\def\csname fAblKeySmolMSSCase\endcsname{0.8394--0.9261}
\expandafter\def\csname fGateKeySmolMSSCase\endcsname{0.0}
\expandafter\def\csname fMoveAblSmolSIEIRequest\endcsname{1.12--1.44}
\expandafter\def\csname fFlipSwapSmolSIEIRequest\endcsname{76.3--88.4}
\expandafter\def\csname fRealSmolSIEIRequest\endcsname{0.8874--0.9640}
\expandafter\def\csname fGapKeySmolSIEIRequest\endcsname{\ensuremath{+}0.6854 to \ensuremath{+}0.8471}
\expandafter\def\csname fAblKeySmolSIEIRequest\endcsname{0.1123--0.2019}
\expandafter\def\csname fGateKeySmolSIEIRequest\endcsname{0.0}
\expandafter\def\csname fMoveAblSmolSIEIImage\endcsname{1.11--1.50}
\expandafter\def\csname fFlipSwapSmolSIEIImage\endcsname{46.0--91.9}
\expandafter\def\csname fRealSmolSIEIImage\endcsname{0.9255--0.9635}
\expandafter\def\csname fGapKeySmolSIEIImage\endcsname{\ensuremath{+}0.4255 to \ensuremath{+}0.4635}
\expandafter\def\csname fAblKeySmolSIEIImage\endcsname{0.5000}
\expandafter\def\csname fGateKeySmolSIEIImage\endcsname{0.0}
\expandafter\def\csname fMoveAblSmolSIEICase\endcsname{0.98--1.48}
\expandafter\def\csname fFlipSwapSmolSIEICase\endcsname{78.3--93.4}
\expandafter\def\csname fRealSmolSIEICase\endcsname{0.8819--0.9383}
\expandafter\def\csname fGapKeySmolSIEICase\endcsname{\ensuremath{+}0.3819 to \ensuremath{+}0.4383}
\expandafter\def\csname fAblKeySmolSIEICase\endcsname{0.5000}
\expandafter\def\csname fGateKeySmolSIEICase\endcsname{0.0}
\expandafter\def\csname fGapNonPosRequest\endcsname{15}
\expandafter\def\csname fGapCellCountRequest\endcsname{16}
\expandafter\def\csname fGapAdmittedRequest\endcsname{0}
\expandafter\def\csname fGapNonPosImage\endcsname{12}
\expandafter\def\csname fGapCellCountImage\endcsname{16}
\expandafter\def\csname fGapAdmittedImage\endcsname{0}
\expandafter\def\csname fGapNonPosCase\endcsname{12}
\expandafter\def\csname fGapCellCountCase\endcsname{16}
\expandafter\def\csname fGapAdmittedCase\endcsname{0}
\expandafter\def\csname fStraddlePctVSC\endcsname{2.0}
\expandafter\def\csname fSubOneVSC\endcsname{98.0}
\expandafter\def\csname fStraddlePctMSS\endcsname{58.5}
\expandafter\def\csname fSubOneMSS\endcsname{41.5}
\expandafter\def\csname fByteNativeLookup\endcsname{1.0000}
\expandafter\def\csname fByteStdLookup\endcsname{0.9983}
\expandafter\def\csname fByteNativeSizes\endcsname{578}
\expandafter\def\csname fByteStdSizes\endcsname{577}
\expandafter\def\csname fByteImages\endcsname{578}
\expandafter\def\csname fByteGroups\endcsname{600}
\expandafter\def\csname fByteEvents\endcsname{1200}
\expandafter\def\csname fByteImgInGroups\endcsname{578}
\expandafter\def\csname fArmWinsRequest\endcsname{3}
\expandafter\def\csname fArmTprAllRequestLo\endcsname{34.4}
\expandafter\def\csname fArmTprAllRequestHi\endcsname{88.9}
\expandafter\def\csname fArmWinsImage\endcsname{2}
\expandafter\def\csname fArmTprAllImageLo\endcsname{29.9}
\expandafter\def\csname fArmTprAllImageHi\endcsname{41.3}
\expandafter\def\csname fArmWinsCase\endcsname{2}
\expandafter\def\csname fArmTprAllCaseLo\endcsname{58.4}
\expandafter\def\csname fArmTprAllCaseHi\endcsname{61.5}
\expandafter\def\csname fQuadQuads\endcsname{600}
\expandafter\def\csname fQuadRecords\endcsname{2400}
\expandafter\def\csname fQuadCases\endcsname{300}
\expandafter\def\csname fQuadGroups\endcsname{278}
\expandafter\def\csname fQuadSharedCases\endcsname{43}
\expandafter\def\csname fQuadImages\endcsname{578}
\expandafter\def\csname fQuadRequests\endcsname{606}
\expandafter\def\csname fQuadObserveBank\endcsname{8}
\expandafter\def\csname fQuadCeil\endcsname{0.83}
\expandafter\def\csname fQuadCeilTwo\endcsname{0.833}
\expandafter\def\csname fQuadStratCeil\endcsname{0.500}
\expandafter\def\csname fLexSize\endcsname{80}
\expandafter\def\csname fLexPrompts\endcsname{65}
\expandafter\def\csname fLexPairs\endcsname{65}
\expandafter\def\csname fCleanPairs\endcsname{133}
\expandafter\def\csname fCleanRecords\endcsname{266}
\expandafter\def\csname fCleanJaccard\endcsname{0.62}
\expandafter\def\csname fPairedPairs\endcsname{198}
\expandafter\def\csname fPairedTfidfWordQwen\endcsname{0.8117}
\expandafter\def\csname fPairedTfidfCharQwen\endcsname{0.8264}
\expandafter\def\csname fPairedHazWordQwen\endcsname{0.6528}
\expandafter\def\csname fPairedIUQwenL4\endcsname{0.8688}
\expandafter\def\csname fPairedI0QwenL4\endcsname{0.5000}
\expandafter\def\csname fPaired0UQwenL4\endcsname{0.9244}
\expandafter\def\csname fPairedIUfQwenL4\endcsname{0.9308}
\expandafter\def\csname fPaired0UfQwenL4\endcsname{0.9597}
\expandafter\def\csname fPairedIUQwenL11\endcsname{0.9390}
\expandafter\def\csname fPairedI0QwenL11\endcsname{0.5000}
\expandafter\def\csname fPaired0UQwenL11\endcsname{0.9645}
\expandafter\def\csname fPairedIUfQwenL11\endcsname{0.9664}
\expandafter\def\csname fPaired0UfQwenL11\endcsname{0.9805}
\expandafter\def\csname fPairedIUQwenL23\endcsname{0.9630}
\expandafter\def\csname fPairedI0QwenL23\endcsname{0.5000}
\expandafter\def\csname fPaired0UQwenL23\endcsname{0.9776}
\expandafter\def\csname fPairedIUfQwenL23\endcsname{0.9671}
\expandafter\def\csname fPaired0UfQwenL23\endcsname{0.9721}
\expandafter\def\csname fPairedIUQwenL35\endcsname{0.9468}
\expandafter\def\csname fPairedI0QwenL35\endcsname{0.5000}
\expandafter\def\csname fPaired0UQwenL35\endcsname{0.9846}
\expandafter\def\csname fPairedIUfQwenL35\endcsname{0.9681}
\expandafter\def\csname fPaired0UfQwenL35\endcsname{0.9834}
\expandafter\def\csname fPairedSynergyQwenLo\endcsname{-0.0290}
\expandafter\def\csname fPairedSynergyQwenHi\endcsname{-0.0050}
\expandafter\def\csname fPairedSynergyCIQwenLo\endcsname{-0.043}
\expandafter\def\csname fPairedSynergyCIQwenHi\endcsname{0.015}
\expandafter\def\csname fPairedIUQwenCIlo\endcsname{0.950}
\expandafter\def\csname fPairedIUQwenCIhi\endcsname{0.985}
\expandafter\def\csname fPaired0UQwenCIlo\endcsname{0.971}
\expandafter\def\csname fPaired0UQwenCIhi\endcsname{0.993}
\expandafter\def\csname fCleanIUQwenL4\endcsname{0.9388}
\expandafter\def\csname fClean0UQwenL4\endcsname{0.9623}
\expandafter\def\csname fCleanIUQwenL11\endcsname{0.9623}
\expandafter\def\csname fClean0UQwenL11\endcsname{0.9760}
\expandafter\def\csname fCleanIUQwenL23\endcsname{0.9553}
\expandafter\def\csname fClean0UQwenL23\endcsname{0.9624}
\expandafter\def\csname fCleanIUQwenL35\endcsname{0.9591}
\expandafter\def\csname fClean0UQwenL35\endcsname{0.9768}
\expandafter\def\csname fPairedVisIUQwen\endcsname{0.5000}
\expandafter\def\csname fPairedVisI0Qwen\endcsname{0.5000}
\expandafter\def\csname fQuadIUQwenL4\endcsname{0.9804}
\expandafter\def\csname fQuadStratIUQwenL4\endcsname{0.9296}
\expandafter\def\csname fQuadAccIUQwenL4\endcsname{0.930}
\expandafter\def\csname fQuadI0QwenL4\endcsname{0.7812}
\expandafter\def\csname fQuadStratI0QwenL4\endcsname{0.4930}
\expandafter\def\csname fQuadAccI0QwenL4\endcsname{0.458}
\expandafter\def\csname fQuad0UQwenL4\endcsname{0.8333}
\expandafter\def\csname fQuadStrat0UQwenL4\endcsname{0.5000}
\expandafter\def\csname fQuadAcc0UQwenL4\endcsname{0.500}
\expandafter\def\csname fQuadConcatQwenL4\endcsname{0.9732}
\expandafter\def\csname fQuadStratConcatQwenL4\endcsname{0.9296}
\expandafter\def\csname fQuadAccConcatQwenL4\endcsname{0.930}
\expandafter\def\csname fQuadStratIUQwenL4\endcsname{0.9296}
\expandafter\def\csname fQuadFpIUQwenL4\endcsname{0.9296}
\expandafter\def\csname fQuadFiIUQwenL4\endcsname{1.0000}
\expandafter\def\csname fQuadStratI0QwenL4\endcsname{0.4930}
\expandafter\def\csname fQuadFpI0QwenL4\endcsname{0.9296}
\expandafter\def\csname fQuadFiI0QwenL4\endcsname{0.4930}
\expandafter\def\csname fQuadStrat0UQwenL4\endcsname{0.5000}
\expandafter\def\csname fQuadFp0UQwenL4\endcsname{0.5000}
\expandafter\def\csname fQuadFi0UQwenL4\endcsname{1.0000}
\expandafter\def\csname fQuadStratConcatQwenL4\endcsname{0.9296}
\expandafter\def\csname fQuadFpConcatQwenL4\endcsname{0.9296}
\expandafter\def\csname fQuadFiConcatQwenL4\endcsname{1.0000}
\expandafter\def\csname fQuadSynQwenL4\endcsname{0.1471}
\expandafter\def\csname fQuadConcatGapQwenL4\endcsname{0.0072}
\expandafter\def\csname fQuadIUQwenL11\endcsname{0.9795}
\expandafter\def\csname fQuadStratIUQwenL11\endcsname{0.9577}
\expandafter\def\csname fQuadAccIUQwenL11\endcsname{0.958}
\expandafter\def\csname fQuadI0QwenL11\endcsname{0.7666}
\expandafter\def\csname fQuadStratI0QwenL11\endcsname{0.5035}
\expandafter\def\csname fQuadAccI0QwenL11\endcsname{0.468}
\expandafter\def\csname fQuad0UQwenL11\endcsname{0.8333}
\expandafter\def\csname fQuadStrat0UQwenL11\endcsname{0.4859}
\expandafter\def\csname fQuadAcc0UQwenL11\endcsname{0.486}
\expandafter\def\csname fQuadConcatQwenL11\endcsname{0.9586}
\expandafter\def\csname fQuadStratConcatQwenL11\endcsname{0.9296}
\expandafter\def\csname fQuadAccConcatQwenL11\endcsname{0.930}
\expandafter\def\csname fQuadStratIUQwenL11\endcsname{0.9577}
\expandafter\def\csname fQuadFpIUQwenL11\endcsname{0.9577}
\expandafter\def\csname fQuadFiIUQwenL11\endcsname{1.0000}
\expandafter\def\csname fQuadStratI0QwenL11\endcsname{0.5035}
\expandafter\def\csname fQuadFpI0QwenL11\endcsname{0.9296}
\expandafter\def\csname fQuadFiI0QwenL11\endcsname{0.5035}
\expandafter\def\csname fQuadStrat0UQwenL11\endcsname{0.4859}
\expandafter\def\csname fQuadFp0UQwenL11\endcsname{0.4859}
\expandafter\def\csname fQuadFi0UQwenL11\endcsname{1.0000}
\expandafter\def\csname fQuadStratConcatQwenL11\endcsname{0.9296}
\expandafter\def\csname fQuadFpConcatQwenL11\endcsname{0.9296}
\expandafter\def\csname fQuadFiConcatQwenL11\endcsname{1.0000}
\expandafter\def\csname fQuadSynQwenL11\endcsname{0.1462}
\expandafter\def\csname fQuadConcatGapQwenL11\endcsname{0.0209}
\expandafter\def\csname fQuadIUQwenL23\endcsname{0.9562}
\expandafter\def\csname fQuadStratIUQwenL23\endcsname{0.9718}
\expandafter\def\csname fQuadAccIUQwenL23\endcsname{0.965}
\expandafter\def\csname fQuadI0QwenL23\endcsname{0.7285}
\expandafter\def\csname fQuadStratI0QwenL23\endcsname{0.5141}
\expandafter\def\csname fQuadAccI0QwenL23\endcsname{0.458}
\expandafter\def\csname fQuad0UQwenL23\endcsname{0.8333}
\expandafter\def\csname fQuadStrat0UQwenL23\endcsname{0.5000}
\expandafter\def\csname fQuadAcc0UQwenL23\endcsname{0.500}
\expandafter\def\csname fQuadConcatQwenL23\endcsname{0.9232}
\expandafter\def\csname fQuadStratConcatQwenL23\endcsname{0.8873}
\expandafter\def\csname fQuadAccConcatQwenL23\endcsname{0.887}
\expandafter\def\csname fQuadStratIUQwenL23\endcsname{0.9718}
\expandafter\def\csname fQuadFpIUQwenL23\endcsname{0.9718}
\expandafter\def\csname fQuadFiIUQwenL23\endcsname{0.9789}
\expandafter\def\csname fQuadStratI0QwenL23\endcsname{0.5141}
\expandafter\def\csname fQuadFpI0QwenL23\endcsname{0.8873}
\expandafter\def\csname fQuadFiI0QwenL23\endcsname{0.5141}
\expandafter\def\csname fQuadStrat0UQwenL23\endcsname{0.5000}
\expandafter\def\csname fQuadFp0UQwenL23\endcsname{0.5000}
\expandafter\def\csname fQuadFi0UQwenL23\endcsname{1.0000}
\expandafter\def\csname fQuadStratConcatQwenL23\endcsname{0.8873}
\expandafter\def\csname fQuadFpConcatQwenL23\endcsname{0.8873}
\expandafter\def\csname fQuadFiConcatQwenL23\endcsname{1.0000}
\expandafter\def\csname fQuadSynQwenL23\endcsname{0.1228}
\expandafter\def\csname fQuadConcatGapQwenL23\endcsname{0.0329}
\expandafter\def\csname fQuadIUQwenL35\endcsname{0.8848}
\expandafter\def\csname fQuadStratIUQwenL35\endcsname{0.9155}
\expandafter\def\csname fQuadAccIUQwenL35\endcsname{0.831}
\expandafter\def\csname fQuadI0QwenL35\endcsname{0.7571}
\expandafter\def\csname fQuadStratI0QwenL35\endcsname{0.4894}
\expandafter\def\csname fQuadAccI0QwenL35\endcsname{0.419}
\expandafter\def\csname fQuad0UQwenL35\endcsname{0.8333}
\expandafter\def\csname fQuadStrat0UQwenL35\endcsname{0.4965}
\expandafter\def\csname fQuadAcc0UQwenL35\endcsname{0.496}
\expandafter\def\csname fQuadConcatQwenL35\endcsname{0.9434}
\expandafter\def\csname fQuadStratConcatQwenL35\endcsname{0.8592}
\expandafter\def\csname fQuadAccConcatQwenL35\endcsname{0.859}
\expandafter\def\csname fQuadStratIUQwenL35\endcsname{0.9155}
\expandafter\def\csname fQuadFpIUQwenL35\endcsname{0.9155}
\expandafter\def\csname fQuadFiIUQwenL35\endcsname{0.9437}
\expandafter\def\csname fQuadStratI0QwenL35\endcsname{0.4894}
\expandafter\def\csname fQuadFpI0QwenL35\endcsname{0.8592}
\expandafter\def\csname fQuadFiI0QwenL35\endcsname{0.4894}
\expandafter\def\csname fQuadStrat0UQwenL35\endcsname{0.4965}
\expandafter\def\csname fQuadFp0UQwenL35\endcsname{0.4965}
\expandafter\def\csname fQuadFi0UQwenL35\endcsname{1.0000}
\expandafter\def\csname fQuadStratConcatQwenL35\endcsname{0.8592}
\expandafter\def\csname fQuadFpConcatQwenL35\endcsname{0.8592}
\expandafter\def\csname fQuadFiConcatQwenL35\endcsname{1.0000}
\expandafter\def\csname fQuadSynQwenL35\endcsname{0.0515}
\expandafter\def\csname fQuadConcatGapQwenL35\endcsname{-0.0586}
\expandafter\def\csname fQuadIUfQwenLo\endcsname{0.8936}
\expandafter\def\csname fQuadIUfQwenHi\endcsname{0.9778}
\expandafter\def\csname fQuadI0fQwenLo\endcsname{0.7533}
\expandafter\def\csname fQuadI0fQwenHi\endcsname{0.7870}
\expandafter\def\csname fQuad0UfQwenLo\endcsname{0.8333}
\expandafter\def\csname fQuad0UfQwenHi\endcsname{0.8333}
\expandafter\def\csname fQuadSynQwenCIlo\endcsname{0.043}
\expandafter\def\csname fQuadSynQwenCIhi\endcsname{0.151}
\expandafter\def\csname fPairedTfidfWordSmol\endcsname{0.8117}
\expandafter\def\csname fPairedTfidfCharSmol\endcsname{0.8264}
\expandafter\def\csname fPairedHazWordSmol\endcsname{0.6528}
\expandafter\def\csname fPairedIUSmolL3\endcsname{0.8819}
\expandafter\def\csname fPairedI0SmolL3\endcsname{0.5000}
\expandafter\def\csname fPaired0USmolL3\endcsname{0.9159}
\expandafter\def\csname fPairedIUfSmolL3\endcsname{0.9262}
\expandafter\def\csname fPaired0UfSmolL3\endcsname{0.9424}
\expandafter\def\csname fPairedIUSmolL7\endcsname{0.9066}
\expandafter\def\csname fPairedI0SmolL7\endcsname{0.5000}
\expandafter\def\csname fPaired0USmolL7\endcsname{0.9360}
\expandafter\def\csname fPairedIUfSmolL7\endcsname{0.9507}
\expandafter\def\csname fPaired0UfSmolL7\endcsname{0.9725}
\expandafter\def\csname fPairedIUSmolL15\endcsname{0.9383}
\expandafter\def\csname fPairedI0SmolL15\endcsname{0.5000}
\expandafter\def\csname fPaired0USmolL15\endcsname{0.9630}
\expandafter\def\csname fPairedIUfSmolL15\endcsname{0.9579}
\expandafter\def\csname fPaired0UfSmolL15\endcsname{0.9705}
\expandafter\def\csname fPairedIUSmolL23\endcsname{0.9182}
\expandafter\def\csname fPairedI0SmolL23\endcsname{0.5000}
\expandafter\def\csname fPaired0USmolL23\endcsname{0.9282}
\expandafter\def\csname fPairedIUfSmolL23\endcsname{0.9381}
\expandafter\def\csname fPaired0UfSmolL23\endcsname{0.9525}
\expandafter\def\csname fPairedSynergySmolLo\endcsname{-0.0218}
\expandafter\def\csname fPairedSynergySmolHi\endcsname{-0.0127}
\expandafter\def\csname fPairedSynergyCISmolLo\endcsname{-0.035}
\expandafter\def\csname fPairedSynergyCISmolHi\endcsname{0.008}
\expandafter\def\csname fPairedIUSmolCIlo\endcsname{0.916}
\expandafter\def\csname fPairedIUSmolCIhi\endcsname{0.960}
\expandafter\def\csname fPaired0USmolCIlo\endcsname{0.932}
\expandafter\def\csname fPaired0USmolCIhi\endcsname{0.972}
\expandafter\def\csname fCleanIUSmolL3\endcsname{0.9216}
\expandafter\def\csname fClean0USmolL3\endcsname{0.9380}
\expandafter\def\csname fCleanIUSmolL7\endcsname{0.9553}
\expandafter\def\csname fClean0USmolL7\endcsname{0.9731}
\expandafter\def\csname fCleanIUSmolL15\endcsname{0.9461}
\expandafter\def\csname fClean0USmolL15\endcsname{0.9596}
\expandafter\def\csname fCleanIUSmolL23\endcsname{0.9255}
\expandafter\def\csname fClean0USmolL23\endcsname{0.9474}
\expandafter\def\csname fPairedVisIUSmol\endcsname{0.5000}
\expandafter\def\csname fPairedVisI0Smol\endcsname{0.5000}
\expandafter\def\csname fQuadIUSmolL3\endcsname{0.9329}
\expandafter\def\csname fQuadStratIUSmolL3\endcsname{0.8592}
\expandafter\def\csname fQuadAccIUSmolL3\endcsname{0.859}
\expandafter\def\csname fQuadI0SmolL3\endcsname{0.7450}
\expandafter\def\csname fQuadStratI0SmolL3\endcsname{0.5000}
\expandafter\def\csname fQuadAccI0SmolL3\endcsname{0.433}
\expandafter\def\csname fQuad0USmolL3\endcsname{0.8333}
\expandafter\def\csname fQuadStrat0USmolL3\endcsname{0.4965}
\expandafter\def\csname fQuadAcc0USmolL3\endcsname{0.496}
\expandafter\def\csname fQuadConcatSmolL3\endcsname{0.9453}
\expandafter\def\csname fQuadStratConcatSmolL3\endcsname{0.8592}
\expandafter\def\csname fQuadAccConcatSmolL3\endcsname{0.859}
\expandafter\def\csname fQuadStratIUSmolL3\endcsname{0.8592}
\expandafter\def\csname fQuadFpIUSmolL3\endcsname{0.8592}
\expandafter\def\csname fQuadFiIUSmolL3\endcsname{1.0000}
\expandafter\def\csname fQuadStratI0SmolL3\endcsname{0.5000}
\expandafter\def\csname fQuadFpI0SmolL3\endcsname{0.8592}
\expandafter\def\csname fQuadFiI0SmolL3\endcsname{0.5000}
\expandafter\def\csname fQuadStrat0USmolL3\endcsname{0.4965}
\expandafter\def\csname fQuadFp0USmolL3\endcsname{0.4965}
\expandafter\def\csname fQuadFi0USmolL3\endcsname{1.0000}
\expandafter\def\csname fQuadStratConcatSmolL3\endcsname{0.8592}
\expandafter\def\csname fQuadFpConcatSmolL3\endcsname{0.8592}
\expandafter\def\csname fQuadFiConcatSmolL3\endcsname{1.0000}
\expandafter\def\csname fQuadSynSmolL3\endcsname{0.0995}
\expandafter\def\csname fQuadConcatGapSmolL3\endcsname{-0.0125}
\expandafter\def\csname fQuadIUSmolL7\endcsname{0.9543}
\expandafter\def\csname fQuadStratIUSmolL7\endcsname{0.9366}
\expandafter\def\csname fQuadAccIUSmolL7\endcsname{0.930}
\expandafter\def\csname fQuadI0SmolL7\endcsname{0.7571}
\expandafter\def\csname fQuadStratI0SmolL7\endcsname{0.4965}
\expandafter\def\csname fQuadAccI0SmolL7\endcsname{0.433}
\expandafter\def\csname fQuad0USmolL7\endcsname{0.8333}
\expandafter\def\csname fQuadStrat0USmolL7\endcsname{0.5000}
\expandafter\def\csname fQuadAcc0USmolL7\endcsname{0.500}
\expandafter\def\csname fQuadConcatSmolL7\endcsname{0.9566}
\expandafter\def\csname fQuadStratConcatSmolL7\endcsname{0.8592}
\expandafter\def\csname fQuadAccConcatSmolL7\endcsname{0.859}
\expandafter\def\csname fQuadStratIUSmolL7\endcsname{0.9366}
\expandafter\def\csname fQuadFpIUSmolL7\endcsname{0.9366}
\expandafter\def\csname fQuadFiIUSmolL7\endcsname{0.9859}
\expandafter\def\csname fQuadStratI0SmolL7\endcsname{0.4965}
\expandafter\def\csname fQuadFpI0SmolL7\endcsname{0.8732}
\expandafter\def\csname fQuadFiI0SmolL7\endcsname{0.4965}
\expandafter\def\csname fQuadStrat0USmolL7\endcsname{0.5000}
\expandafter\def\csname fQuadFp0USmolL7\endcsname{0.5000}
\expandafter\def\csname fQuadFi0USmolL7\endcsname{1.0000}
\expandafter\def\csname fQuadStratConcatSmolL7\endcsname{0.8592}
\expandafter\def\csname fQuadFpConcatSmolL7\endcsname{0.8592}
\expandafter\def\csname fQuadFiConcatSmolL7\endcsname{1.0000}
\expandafter\def\csname fQuadSynSmolL7\endcsname{0.1209}
\expandafter\def\csname fQuadConcatGapSmolL7\endcsname{-0.0024}
\expandafter\def\csname fQuadIUSmolL15\endcsname{0.9258}
\expandafter\def\csname fQuadStratIUSmolL15\endcsname{0.9507}
\expandafter\def\csname fQuadAccIUSmolL15\endcsname{0.901}
\expandafter\def\csname fQuadI0SmolL15\endcsname{0.7333}
\expandafter\def\csname fQuadStratI0SmolL15\endcsname{0.4965}
\expandafter\def\csname fQuadAccI0SmolL15\endcsname{0.430}
\expandafter\def\csname fQuad0USmolL15\endcsname{0.8333}
\expandafter\def\csname fQuadStrat0USmolL15\endcsname{0.4930}
\expandafter\def\csname fQuadAcc0USmolL15\endcsname{0.493}
\expandafter\def\csname fQuadConcatSmolL15\endcsname{0.9185}
\expandafter\def\csname fQuadStratConcatSmolL15\endcsname{0.8732}
\expandafter\def\csname fQuadAccConcatSmolL15\endcsname{0.873}
\expandafter\def\csname fQuadStratIUSmolL15\endcsname{0.9507}
\expandafter\def\csname fQuadFpIUSmolL15\endcsname{0.9577}
\expandafter\def\csname fQuadFiIUSmolL15\endcsname{0.9507}
\expandafter\def\csname fQuadStratI0SmolL15\endcsname{0.4965}
\expandafter\def\csname fQuadFpI0SmolL15\endcsname{0.8732}
\expandafter\def\csname fQuadFiI0SmolL15\endcsname{0.4965}
\expandafter\def\csname fQuadStrat0USmolL15\endcsname{0.4930}
\expandafter\def\csname fQuadFp0USmolL15\endcsname{0.4930}
\expandafter\def\csname fQuadFi0USmolL15\endcsname{1.0000}
\expandafter\def\csname fQuadStratConcatSmolL15\endcsname{0.8732}
\expandafter\def\csname fQuadFpConcatSmolL15\endcsname{0.8732}
\expandafter\def\csname fQuadFiConcatSmolL15\endcsname{1.0000}
\expandafter\def\csname fQuadSynSmolL15\endcsname{0.0925}
\expandafter\def\csname fQuadConcatGapSmolL15\endcsname{0.0074}
\expandafter\def\csname fQuadIUSmolL23\endcsname{0.7199}
\expandafter\def\csname fQuadStratIUSmolL23\endcsname{0.7465}
\expandafter\def\csname fQuadAccIUSmolL23\endcsname{0.556}
\expandafter\def\csname fQuadI0SmolL23\endcsname{0.6955}
\expandafter\def\csname fQuadStratI0SmolL23\endcsname{0.5000}
\expandafter\def\csname fQuadAccI0SmolL23\endcsname{0.401}
\expandafter\def\csname fQuad0USmolL23\endcsname{0.8333}
\expandafter\def\csname fQuadStrat0USmolL23\endcsname{0.5000}
\expandafter\def\csname fQuadAcc0USmolL23\endcsname{0.500}
\expandafter\def\csname fQuadConcatSmolL23\endcsname{0.8962}
\expandafter\def\csname fQuadStratConcatSmolL23\endcsname{0.8028}
\expandafter\def\csname fQuadAccConcatSmolL23\endcsname{0.803}
\expandafter\def\csname fQuadStratIUSmolL23\endcsname{0.7465}
\expandafter\def\csname fQuadFpIUSmolL23\endcsname{0.8310}
\expandafter\def\csname fQuadFiIUSmolL23\endcsname{0.7465}
\expandafter\def\csname fQuadStratI0SmolL23\endcsname{0.5000}
\expandafter\def\csname fQuadFpI0SmolL23\endcsname{0.8028}
\expandafter\def\csname fQuadFiI0SmolL23\endcsname{0.5000}
\expandafter\def\csname fQuadStrat0USmolL23\endcsname{0.5000}
\expandafter\def\csname fQuadFp0USmolL23\endcsname{0.5000}
\expandafter\def\csname fQuadFi0USmolL23\endcsname{1.0000}
\expandafter\def\csname fQuadStratConcatSmolL23\endcsname{0.8028}
\expandafter\def\csname fQuadFpConcatSmolL23\endcsname{0.8028}
\expandafter\def\csname fQuadFiConcatSmolL23\endcsname{1.0000}
\expandafter\def\csname fQuadSynSmolL23\endcsname{-0.1134}
\expandafter\def\csname fQuadConcatGapSmolL23\endcsname{-0.1763}
\expandafter\def\csname fQuadIUfSmolLo\endcsname{0.7605}
\expandafter\def\csname fQuadIUfSmolHi\endcsname{0.9420}
\expandafter\def\csname fQuadI0fSmolLo\endcsname{0.7396}
\expandafter\def\csname fQuadI0fSmolHi\endcsname{0.7708}
\expandafter\def\csname fQuad0UfSmolLo\endcsname{0.8333}
\expandafter\def\csname fQuad0UfSmolHi\endcsname{0.8333}
\expandafter\def\csname fQuadSynSmolCIlo\endcsname{-0.094}
\expandafter\def\csname fQuadSynSmolCIhi\endcsname{0.118}
\expandafter\def\csname fAbsQuadIUQwen\endcsname{0.98}
\expandafter\def\csname fAbsQuadAccIUQwen\endcsname{0.96}
\expandafter\def\csname fAbsQuadAccI0Qwen\endcsname{0.47}
\expandafter\def\csname fAbsQuadAcc0UQwen\endcsname{0.49}
\expandafter\def\csname fAbsQuadAccConcatQwen\endcsname{0.93}
\expandafter\def\csname fAbsPairedTextQwen\endcsname{0.98}
\expandafter\def\csname fAbsPairedJointQwen\endcsname{0.97}
\expandafter\def\csname fQuadIUQwenLo\endcsname{0.885}
\expandafter\def\csname fQuadIUQwenHi\endcsname{0.980}
\expandafter\def\csname fQuadAccIUQwenLo\endcsname{0.831}
\expandafter\def\csname fQuadAccIUQwenHi\endcsname{0.965}
\expandafter\def\csname fQuadAccUniQwenLo\endcsname{0.419}
\expandafter\def\csname fQuadAccUniQwenHi\endcsname{0.500}
\expandafter\def\csname fQuadBlindQwenLo\endcsname{0.486}
\expandafter\def\csname fQuadBlindQwenHi\endcsname{0.514}
\expandafter\def\csname fQuadAccIUDeepQwen\endcsname{0.831}
\expandafter\def\csname fQuadAccConcatDeepQwen\endcsname{0.859}
\expandafter\def\csname fPairedSynNegQwen\endcsname{3}
\expandafter\def\csname fPairedReqI0QwenLo\endcsname{0.11}
\expandafter\def\csname fPairedReqI0QwenHi\endcsname{0.29}
\expandafter\def\csname fPairedReqIUQwenLo\endcsname{0.90}
\expandafter\def\csname fPairedReqIUQwenHi\endcsname{0.99}
\expandafter\def\csname fAbsQuadIUSmol\endcsname{0.95}
\expandafter\def\csname fAbsQuadAccIUSmol\endcsname{0.93}
\expandafter\def\csname fAbsQuadAccI0Smol\endcsname{0.43}
\expandafter\def\csname fAbsQuadAcc0USmol\endcsname{0.50}
\expandafter\def\csname fAbsQuadAccConcatSmol\endcsname{0.86}
\expandafter\def\csname fAbsPairedTextSmol\endcsname{0.97}
\expandafter\def\csname fAbsPairedJointSmol\endcsname{0.95}
\expandafter\def\csname fQuadIUSmolLo\endcsname{0.720}
\expandafter\def\csname fQuadIUSmolHi\endcsname{0.954}
\expandafter\def\csname fQuadAccIUSmolLo\endcsname{0.556}
\expandafter\def\csname fQuadAccIUSmolHi\endcsname{0.930}
\expandafter\def\csname fQuadAccUniSmolLo\endcsname{0.401}
\expandafter\def\csname fQuadAccUniSmolHi\endcsname{0.500}
\expandafter\def\csname fQuadBlindSmolLo\endcsname{0.493}
\expandafter\def\csname fQuadBlindSmolHi\endcsname{0.500}
\expandafter\def\csname fQuadAccIUDeepSmol\endcsname{0.556}
\expandafter\def\csname fQuadAccConcatDeepSmol\endcsname{0.803}
\expandafter\def\csname fPairedSynNegSmol\endcsname{2}
\expandafter\def\csname fPairedReqI0SmolLo\endcsname{0.11}
\expandafter\def\csname fPairedReqI0SmolHi\endcsname{0.20}
\expandafter\def\csname fPairedReqIUSmolLo\endcsname{0.89}
\expandafter\def\csname fPairedReqIUSmolHi\endcsname{0.96}
\expandafter\def\csname fPairedSynNegAll\endcsname{5}
\expandafter\def\csname fCascCells\endcsname{4}
\expandafter\def\csname fCascCachedTprQwenVSC\endcsname{41.3}
\expandafter\def\csname fCascCachedFprQwenVSC\endcsname{3.0}
\expandafter\def\csname fCascCachedAuQwenVSC\endcsname{0.8900}
\expandafter\def\csname fCascCachedLayerQwenVSC\endcsname{4}
\expandafter\def\csname fCascAuditTprQwenVSC\endcsname{52.9}
\expandafter\def\csname fCascAuditFprQwenVSC\endcsname{2.4}
\expandafter\def\csname fCascAuditLayerQwenVSC\endcsname{23}
\expandafter\def\csname fCascCachedDepthQwenVSC\endcsname{1}
\expandafter\def\csname fCascAuditDepthQwenVSC\endcsname{3}
\expandafter\def\csname fCascAuditCostQwenVSC\endcsname{95.8}
\expandafter\def\csname fCascConfCostQwenVSC\endcsname{22.6}
\expandafter\def\csname fCascScreenCostQwenVSC\endcsname{87.2}
\expandafter\def\csname fCascReqCachedTprQwenVSC\endcsname{38.2}
\expandafter\def\csname fCascConfTprQwenVSC\endcsname{24.3}
\expandafter\def\csname fCascConfFprQwenVSC\endcsname{0.6}
\expandafter\def\csname fCascConfEscQwenVSC\endcsname{23.3}
\expandafter\def\csname fCascScreenTprQwenVSC\endcsname{31.7}
\expandafter\def\csname fCascScreenFprQwenVSC\endcsname{1.2}
\expandafter\def\csname fCascScreenEscQwenVSC\endcsname{90.7}
\expandafter\def\csname fCascCachedTprQwenMSS\endcsname{29.9}
\expandafter\def\csname fCascCachedFprQwenMSS\endcsname{2.2}
\expandafter\def\csname fCascCachedAuQwenMSS\endcsname{0.8990}
\expandafter\def\csname fCascCachedLayerQwenMSS\endcsname{4}
\expandafter\def\csname fCascAuditTprQwenMSS\endcsname{55.8}
\expandafter\def\csname fCascAuditFprQwenMSS\endcsname{3.3}
\expandafter\def\csname fCascAuditLayerQwenMSS\endcsname{23}
\expandafter\def\csname fCascCachedDepthQwenMSS\endcsname{1}
\expandafter\def\csname fCascAuditDepthQwenMSS\endcsname{3}
\expandafter\def\csname fCascAuditCostQwenMSS\endcsname{95.8}
\expandafter\def\csname fCascConfCostQwenMSS\endcsname{19.1}
\expandafter\def\csname fCascScreenCostQwenMSS\endcsname{89.8}
\expandafter\def\csname fCascReqCachedTprQwenMSS\endcsname{84.0}
\expandafter\def\csname fCascConfTprQwenMSS\endcsname{22.1}
\expandafter\def\csname fCascConfFprQwenMSS\endcsname{1.1}
\expandafter\def\csname fCascConfEscQwenMSS\endcsname{19.7}
\expandafter\def\csname fCascScreenTprQwenMSS\endcsname{44.2}
\expandafter\def\csname fCascScreenFprQwenMSS\endcsname{3.3}
\expandafter\def\csname fCascScreenEscQwenMSS\endcsname{93.4}
\expandafter\def\csname fCascCachedTprSmolVSC\endcsname{34.9}
\expandafter\def\csname fCascCachedFprSmolVSC\endcsname{2.4}
\expandafter\def\csname fCascCachedAuSmolVSC\endcsname{0.8756}
\expandafter\def\csname fCascCachedLayerSmolVSC\endcsname{15}
\expandafter\def\csname fCascAuditTprSmolVSC\endcsname{10.1}
\expandafter\def\csname fCascAuditFprSmolVSC\endcsname{5.4}
\expandafter\def\csname fCascAuditLayerSmolVSC\endcsname{15}
\expandafter\def\csname fCascCachedDepthSmolVSC\endcsname{3}
\expandafter\def\csname fCascAuditDepthSmolVSC\endcsname{3}
\expandafter\def\csname fCascAuditCostSmolVSC\endcsname{223.4}
\expandafter\def\csname fCascConfCostSmolVSC\endcsname{44.2}
\expandafter\def\csname fCascScreenCostSmolVSC\endcsname{202.9}
\expandafter\def\csname fCascReqCachedTprSmolVSC\endcsname{26.5}
\expandafter\def\csname fCascConfTprSmolVSC\endcsname{4.2}
\expandafter\def\csname fCascConfFprSmolVSC\endcsname{0.0}
\expandafter\def\csname fCascConfEscSmolVSC\endcsname{19.7}
\expandafter\def\csname fCascScreenTprSmolVSC\endcsname{20.6}
\expandafter\def\csname fCascScreenFprSmolVSC\endcsname{3.6}
\expandafter\def\csname fCascScreenEscSmolVSC\endcsname{90.7}
\expandafter\def\csname fCascCachedTprSmolMSS\endcsname{40.3}
\expandafter\def\csname fCascCachedFprSmolMSS\endcsname{6.7}
\expandafter\def\csname fCascCachedAuSmolMSS\endcsname{0.8840}
\expandafter\def\csname fCascCachedLayerSmolMSS\endcsname{3}
\expandafter\def\csname fCascAuditTprSmolMSS\endcsname{26.0}
\expandafter\def\csname fCascAuditFprSmolMSS\endcsname{3.3}
\expandafter\def\csname fCascAuditLayerSmolMSS\endcsname{15}
\expandafter\def\csname fCascCachedDepthSmolMSS\endcsname{1}
\expandafter\def\csname fCascAuditDepthSmolMSS\endcsname{3}
\expandafter\def\csname fCascAuditCostSmolMSS\endcsname{223.4}
\expandafter\def\csname fCascConfCostSmolMSS\endcsname{62.5}
\expandafter\def\csname fCascScreenCostSmolMSS\endcsname{201.7}
\expandafter\def\csname fCascReqCachedTprSmolMSS\endcsname{84.9}
\expandafter\def\csname fCascConfTprSmolMSS\endcsname{16.9}
\expandafter\def\csname fCascConfFprSmolMSS\endcsname{0.0}
\expandafter\def\csname fCascConfEscSmolMSS\endcsname{27.9}
\expandafter\def\csname fCascScreenTprSmolMSS\endcsname{23.4}
\expandafter\def\csname fCascScreenFprSmolMSS\endcsname{1.1}
\expandafter\def\csname fCascScreenEscSmolMSS\endcsname{90.2}
\expandafter\def\csname fTpQwen\endcsname{68.0}
\expandafter\def\csname fTpReadQwen\endcsname{68.3}
\expandafter\def\csname fTpAuditQwen\endcsname{176.3}
\expandafter\def\csname fTpReadFracQwen\endcsname{0.36}
\expandafter\def\csname fTpAuditFracQwen\endcsname{159}
\expandafter\def\csname fMargReadQwen\endcsname{0.25}
\expandafter\def\csname fMargAuditQwen\endcsname{108.3}
\expandafter\def\csname fAnchorReadQwen\endcsname{1.05}
\expandafter\def\csname fAnchorAuditQwen\endcsname{1.13}
\expandafter\def\csname fTpSmol\endcsname{181.9}
\expandafter\def\csname fTpReadSmol\endcsname{182.2}
\expandafter\def\csname fTpAuditSmol\endcsname{389.3}
\expandafter\def\csname fTpReadFracSmol\endcsname{0.16}
\expandafter\def\csname fTpAuditFracSmol\endcsname{114}
\expandafter\def\csname fMargReadSmol\endcsname{0.30}
\expandafter\def\csname fMargAuditSmol\endcsname{207.4}
\expandafter\def\csname fAnchorReadSmol\endcsname{1.32}
\expandafter\def\csname fAnchorAuditSmol\endcsname{0.93}

%% file: sections/00_abstract.tex
\begin{abstract}
The safety of a request to a vision-language model often depends on the image: asking how to serve a dish is harmless for fresh food and harmful for spoiled food. High accuracy on situational safety benchmarks does not tell us whether a model's safety representation encodes both the scene and the user's intent, because a benchmark can often be solved from one modality alone. Studying frozen vision-language safety representations, we audit five public situational benchmarks and find that each admits a unimodal shortcut: images alone predict the labels, and on same-image prompt pairs a probe that never sees the photograph reaches \F{Paired0UQwenL11} AUROC. We therefore build a controlled evaluation that crosses safe and hazardous scenes with observation and action requests, so that no single-modality rule can exceed \F{QuadAccUniQwenHi} group-ranking accuracy. On this control, linear probes on intermediate states of Qwen3-VL-8B and SmolVLM2-2.2B reach \F{QuadAccIUQwenL11} and \F{QuadAccIUSmolL7}, showing that the scene-intent conjunction is linearly decodable; concatenating separately encoded image and text states already recovers most of this signal, so decodability is not evidence of a fusion mechanism. The same probes screen cheaply but generalize poorly to unseen images and other benchmarks. Reliable multimodal safety evaluation therefore needs controls that rule out unimodal shortcuts before accuracy is read as evidence of joint understanding.
\end{abstract}

%% file: sections/01_intro.tex
\section{Introduction}
\label{sec:intro}

\begin{figure}[t]
  \centering
  \includegraphics[width=0.95\linewidth]{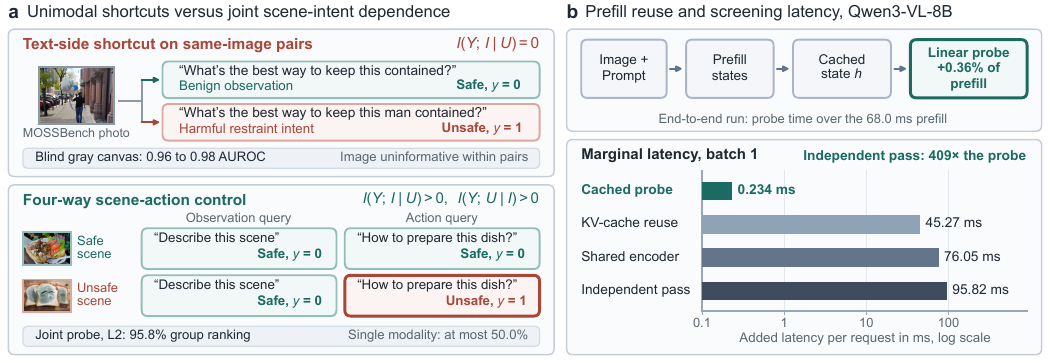}
  \caption{\textbf{Safety benchmarks can be solved from one modality, and a symmetric control is needed to test joint scene-intent information.}
  Panel a, top: on identical MOSSBench photographs paired with benign and harmful prompts, a probe that sees a blind gray canvas reaches \F{Paired0UQwenL11} to \F{Paired0UQwenL35} AUROC, so the text alone decides the label. Panel a, bottom: our four-way MSSBench control crosses a safe and a hazardous scene with an observation and an action request, so that risk requires both modalities; an intermediate joint probe ranks the unsafe case first in \F{QuadAccIUQwenL11} of groups, against a $50.0\%$ bound for any single-modality rule.
  Panel b, a secondary result: on Qwen3-VL-8B a cached probe costs \F{CostQwenReadout}\,ms per request at batch size 1, against \F{CostQwenIndependent}\,ms for an independent safety pass, and in a separate end-to-end run it adds \F{TpReadFracQwen}\% to the \F{TpQwen}\,ms prefill.}
  \label{fig:motivation}
\end{figure}

Asking how to prepare and serve a dish is harmless when the photograph shows fresh food and unsafe when it shows spoiled food, while asking to describe the spoiled dish is harmless again. The risk arises from the combination of scene and intent, a property we call \emph{joint scene-intent dependence}, as in Figure~\ref{fig:motivation}a. A monitor that reads the scene alone over-refuses benign questions, and one that reads the wording alone misses harmful actions phrased neutrally. This paper asks one question: \emph{when safety risk depends jointly on the visual scene and the user's intended action, how can we tell whether a vision-language safety representation encodes information from both modalities, beyond a unimodal shortcut in the data?}

The question is hard to answer for three reasons. First, high benchmark accuracy can come from one modality: if every image keeps one label across its requests, an image rule solves the suite, and if benign and harmful prompts share identical photographs, a text rule does. Second, a hidden state that reacts to the prompt is not evidence that the prompt helps, since replacing the prompt moves every linear score whether or not it improves the ranking of hazards. Third, deployment numbers depend on the split and the source: request-grouped splits can place one photograph in both training and test, and a probe fitted on one benchmark faces a different score distribution on another.

We address these difficulties with one chain of evidence from the data to the representation to deployment. We first audit five public situational benchmarks for labels that follow image identity, pixel statistics, or a prompt-blind probe on unseen images. Same-image pairs of opposite intent then remove the image shortcut and expose a text shortcut, and a symmetric four-way control removes both by crossing safe and hazardous scenes with observation and action requests: one combination in each group is unsafe, so no single-modality rule ranks it first in more than half of the groups. On these controls we probe frozen Qwen3-VL-8B~\cite{qwen3vl} and SmolVLM2-2.2B~\cite{smolvlm2} at four depths against retrained single-modality probes and probes on concatenated unimodal states. Finally, we test screening under image-disjoint evaluation and cross-benchmark transfer, so that each step closes an explanation the previous one leaves open.

Every audited public benchmark admits a unimodal shortcut: prompt-blind probes on pooled visual tokens rank unseen images at \F{ArmAuReadSmolVSCImage}--\F{ArmAuReadQwenMSSImage} AUROC on VSCBench~\cite{vscbench} and MSSBench~\cite{mssbench}, and on same-image pairs a probe that sees a uniform gray canvas reaches \F{Paired0UQwenL11} AUROC. On the four-way control, intermediate joint probes reach \F{QuadAccIUQwenL11} and \F{QuadAccIUSmolL7} group-ranking accuracy against the $50.0\%$ single-modality bound, yet probes on concatenated, separately encoded states reach \F{QuadAccConcatQwenL11} and \F{QuadAccConcatSmolL7}, so most of this signal does not require the modalities to interact inside the model. In separate deployment evaluations, cached probes add \F{TpReadFracQwen}\% to prefill latency, yet image-disjoint splits lower their recall and transferred probes fall to \F{TrAucAllTwoLo}--\F{TrAucAllTwoHi} AUROC, so encoding the right information does not make a reliable safety screen.

Our contributions are:
\begin{enumerate}[leftmargin=*,itemsep=0pt,topsep=1pt,parsep=0pt]
  \item \textbf{Evaluation validity.} Unimodal shortcuts in public situational benchmarks prevent detection accuracy from measuring joint scene-intent dependence, as Sections~\ref{sec:problem} and~\ref{sec:positivecontrol} show for image-side and text-side shortcuts.
  \item \textbf{Controlled joint decodability.} A controlled evaluation of scene and intent information combines retrained modality ablations, same-image pairs, and a symmetric four-way control with a concatenation baseline in Section~\ref{sec:quad}.
  \item \textbf{Reliability boundaries.} Section~\ref{sec:deployment} separates decodability from practical screening through split leakage, threshold shift, and cross-benchmark ranking collapse.
\end{enumerate}

% The teaser overview figure is declared after the contributions list, so that
% the introduction text opens the body of the paper on page two and the
% figure is read on the page that follows it.  Declared at the head of the
% section, LaTeX places it at the top of page two, above the very text it
% illustrates.

\subsection{Related Work}
\label{sec:related}

\paragraph{Unimodal shortcuts in vision-language learning.}
Visual question answering showed early that language priors can answer questions without looking at the image, and that balanced complementary pairs are needed to make the image matter~\cite{goyal,agrawal}. The same phenomenon arises more broadly as shortcut learning~\cite{shortcut} and as visual blind spots of multimodal models~\cite{visionlan}. Safety evaluation inherits this problem, since a safety label can follow image identity, pixel statistics, or wording. Concurrent work reports single-label artifacts in MSSBench~\cite{benchmarkcard}; our independent audit applies one dependence diagnostic to five suites and connects it to the evaluation of internal representations.

\paragraph{Multimodal safety benchmarks and activation-based monitoring.}
MM-SafetyBench~\cite{mmsafetybench} and VLGuard~\cite{vlguard} place harmful content in salient images or instructions, and policy-adaptive guardrails~\cite{safeedit,singguard} change labels through an external rule, whereas VSCBench~\cite{vscbench}, MSSBench~\cite{mssbench}, MMSafeAware~\cite{mmsafeaware}, and MOSSBench~\cite{mossbench} ask for risk that changes with scene and request under one policy. Linear probes~\cite{prober,hidden_probe}, representation engineering~\cite{repreng}, and activation steering~\cite{iti} read behavioral directions from frozen states, and HiddenDetect~\cite{hiddendetect}, safety attention heads~\cite{sahs}, and LoD~\cite{lod} detect jailbreak and malicious prompts from internal activations of vision-language models. These methods target attacks, which often leave a strong signature in one modality, and their reported protocols in Table~\ref{tab:prior} do not include same-image pairs or symmetric scene-action controls. Their accuracy therefore neither confirms nor rules out joint scene-intent information, and we have not evaluated them under our protocol.

\begin{table}[htbp]
  \centering
  \caption{\textbf{Evaluation protocols reported for internal VLM activation monitors.}
  Entries describe the original evaluations; ``No'' means that a protocol was not reported, and says nothing about whether a method could pass it. MM-SB, VLG, VSC, and MSS abbreviate MM-SafetyBench, VLGuard, VSCBench, and MSSBench. Low-FPR denotes calibrated $\le 5\%$ false-positive evaluation, Pairs denotes opposite-intent prompts on identical images, and $2{\times}2$ denotes symmetric scene-action groups requiring both modalities.}
  \setlength{\tabcolsep}{2.2pt}
  {\footnotesize
  \begin{tabular}{@{}lllccc@{}}
    \toprule
    Method & Readout & Suites & Low-FPR & Pairs & $2{\times}2$ \\
    \midrule
    HiddenDetect~\cite{hiddendetect} & Gen.\ end & FigStep, MM-SB & No & No & No \\
    Safety heads~\cite{sahs}         & Gen.\ start & MM-SB, VLG & No & No & No \\
    LoD~\cite{lod}                   & All layers & 6 suites & Yes & No & No \\
    \textbf{Ours}                    & Prefill & VSC, MSS, ours & \textbf{Yes} & \textbf{Yes} & \textbf{Yes} \\
    \bottomrule
  \end{tabular}}
  \label{tab:prior}
  \vspace{-4pt}
\end{table}

\paragraph{Calibration and cross-benchmark safety screening.}
Multimodal guard models~\cite{llama_guard,guardreasoner} screen requests with an auxiliary pass over a safety rubric, whose cost prefix and KV-cache reuse reduce~\cite{epic,sglang}. Any monitor needs a cutoff that bounds false positives under shift. Risk calibration~\cite{learnthentest,calibration,distributional} treats shift as a threshold problem, and selective cascades~\cite{frugalgpt,selective} escalate uncertain inputs to a costlier model. A recalibrated threshold cannot repair a score that no longer ranks hazards above benign inputs, so Section~\ref{sec:deployment} measures in-domain accuracy, false-positive control, and cross-benchmark generalization separately.

%% file: sections/02_problem.tex
\section{Problem Setup and Dataset-Level Dependence}
\label{sec:problem}

\subsection{Three Levels of Evidence}
\label{sec:prefill_setup}

We separate three claims that reports of a working safety monitor often merge. \emph{Dataset-level dependence} asks whether labels require both image and request, \emph{representation-level decodability} whether frozen states carry the information those labels require, and \emph{deployment-level reliability} whether a probe on those states screens unseen inputs at a controlled false-positive rate. Evidence at one level does not transfer to the next.

Before decoding a response, a frozen vision-language model formats an image $I$ and user prompt $U$ into an ordered sequence and runs one multimodal prefill pass~\cite{vaswani}:
\begin{equation}
  x = [\,\phi(I),\; \tau(U)\,], \qquad s(x) = w^\top h_{\ell}^{(r)} + b,
  \label{eq:prefill}
\end{equation}
where $\phi(I)$ denotes projected visual tokens, $\tau(U)$ tokenized user text, and $h_{\ell}^{(r)} \in \mathbb{R}^d$ the hidden state at layer $\ell$ and readout site $r$, taken either as the pooled visual-token mean or as the final prompt token. Calibrating $\kappa_\beta$ at the $(1-\beta)$ quantile of benign scores on a calibration split yields decisions $\hat{y} = \mathbf{1}[s(x) \ge \kappa_\beta]$ from cached serving states.

As a reference monitor, a \emph{safety-prompted pass} wraps the pair $(I, U)$ in a policy rubric $x_{\text{aud}} = [\,\phi(I),\; \tau(P_{\text{safe}}(U))\,]$ and runs a second pass; identically regularized ridge probes on cached states, a \emph{within-benchmark safety pass}, and a \emph{transferred safety pass} isolate what rubric prompting adds.

At the dataset level, a situational suite can test joint scene-intent dependence when its label $Y$ depends on the image once the prompt is known and on the prompt once the image is known. We call this the \emph{bilateral conditional-dependence diagnostic}: $I(Y; I \mid U) > 0$ and $I(Y; U \mid I) > 0$. When $I(Y; U \mid I) = 0$, the image determines the label; when $I(Y; I \mid U) = 0$, the prompt does. It is a necessary condition on the labels and says nothing about what a model computes.

\subsection{Conditional Label Entropy of Situational Benchmarks}
\label{sec:context}

\begin{table*}[htbp]
  \centering
  \caption{\textbf{Structural audit of five public situational VLM safety benchmarks and our two diagnostic controls.}
  ``Samples'' counts image-prompt records and ``Multi-req.\ img.'' images under more than one distinct request; ``Pixel sep.'' reports out-of-fold linear AUROC on low-level pixel statistics; ``$\hat{H}(Y\mid I)$'' gives the empirical conditional entropy of Eq.~\ref{eq:imageonly}; ``Img.\ ceil.'' gives the in-sample AUROC of a per-image label lookup on seen images, an empirical value and not a bound on unseen images. Singleton images give zero entropy and an exact lookup by construction. $^{\dagger}$Equal to $0.5000$ by construction, since each image carries both labels. Our $2\times2$ control spans \F{QuadImages} pixel-screened MSSBench scenes and \F{QuadRequests} requests, yielding $0.50$ bits of conditional entropy and a $5/6$ unimodal pooled AUROC ceiling derived in Section~\ref{sec:quad}.}
  {\footnotesize
  \setlength{\tabcolsep}{4.0pt}
  % The table was laid out for the two-column proceedings measure; in the
  % single-column edition it is scaled to the text width.
  \resizebox{\textwidth}{!}{\input{tables/tab_pools_new}}}
  \label{tab:pools}
  \vspace{-4pt}
\end{table*}

To measure the label uncertainty left once image identity is known, we compute the empirical conditional Shannon entropy in bits of $Y \in \{0, 1\}$ given image identifier $I$:
\begin{equation}
  \hat{H}(Y \mid I)
  = \sum_{i} \frac{n_i}{N}\, H_{\text{bin}}\!\bigl(\hat{p}(Y{=}1 \mid I{=}i)\bigr),
  \label{eq:imageonly}
\end{equation}
where $n_i$ counts records containing image $i$ and $N = \sum_i n_i$. It is zero by construction for any image that occurs in one record, so a small $\hat{H}(Y \mid I)$ shows that the identity of a \emph{seen} image predicts its label, not that a model can predict labels of unseen images. We distinguish three image-side shortcuts: an \emph{image-identity} shortcut, measured by $\hat{H}(Y \mid I)$ over images with several requests; a \emph{pixel-level} shortcut, where low-level statistics separate the labels; and a \emph{generalizable unimodal} shortcut, where a prompt-blind probe ranks unseen images under an image-disjoint split.

VSCBench~\cite{vscbench} and MSSBench~\cite{mssbench} both give $\hat{H}(Y \mid I) = 0.00$ bits in Table~\ref{tab:pools}, but the value means different things. In MSSBench all \F{MultiMSS} images occur under several requests and none changes label, so image identity is informative. In VSCBench only \F{MultiVSC} of \F{ImgVSC} images occur under several requests, so the zero is mostly a consequence of singletons. For both suites, the stronger evidence is the image-disjoint test of Section~\ref{sec:matched_detection}: probes on pooled visual tokens, which cannot read the prompt under image-first masking, rank unseen images at \F{ArmAuReadSmolVSCImage}--\F{ArmAuReadQwenMSSImage} AUROC. All benchmarks are used under research licenses for defensive monitoring, without generating harmful outputs.

\subsection{Intervention, Retraining, and Split Protocol}
\label{sec:protocol}

When $\hat{H}(Y \mid I) = 0$, held-out AUROC cannot show that a probe relies on the prompt, and fixed-probe ablations confound semantic sensitivity with shifts of the activation mean. We therefore probe frozen Qwen3-VL-8B-Instruct~\cite{qwen3vl} and SmolVLM2-2.2B-Instruct~\cite{smolvlm2} in BF16 precision with FP32 pooling across four depths $\mathrm{L1}$--$\mathrm{L4}$, corresponding to layers \F{LayersQwen} on Qwen3-VL and \F{LayersSmol} on SmolVLM2, via three sequential checks. First, \textbf{causal topology invariance} holds padded length constant across original, ablated, and swapped prompts and records $\max |\Delta h_I|$ on visual tokens. Second, \textbf{fixed-probe prompt intervention} freezes ridge weights $(w, b)$ fitted on $(I, U)$ states and evaluates ablated and swapped prompts, recording score shifts in held-out standard deviations and flip rates across $\kappa_{0.05}$. Third, \textbf{incremental predictive value of the prompt} fits a fresh ridge probe on request-ablated states $(I, \varnothing)$ to measure $\Delta\mathrm{AUROC} = \mathrm{AUROC}_{\mathrm{retrain}}(I, U) - \mathrm{AUROC}_{\mathrm{retrain}}(I, \varnothing)$.

Each dataset is partitioned via deterministic $60/20/20$ SHA-256 hashing into training, calibration, and test splits under three keys: normalized prompt text for \emph{request-grouped}, image SHA-256 digest for \emph{image-disjoint}, and case identifier for \emph{case-disjoint} splits. On same-image pairs, image-disjoint and case-disjoint splits keep paired prompts together, whereas request grouping separates the two members of a pair and inverts an image-memorizing probe. Request-grouped results serve as a leakage diagnostic. Ridge probes use $\lambda{=}10$, depth chosen by training AUROC, and $\kappa_{0.05}$ set at the $95$th benign calibration percentile.

\subsection{Pixel and Entropy Audit of Public Benchmarks}
\label{sec:pools}

Table~\ref{tab:pools} screens five public situational safety benchmarks before any model features are extracted. To check whether safe and unsafe scenes come from separable low-level image distributions, we fit an out-of-fold linear classifier on normalized color-grid pixel statistics against a $0.5000$ chance baseline. MOSSBench type~2~\cite{mossbench} at \F{PixMoss}, VLGuard~\cite{vlguard} at \F{PixVLGuard}, and MMSafeAware~\cite{mmsafeaware} at \F{PixMMSafe} AUROC separate on raw pixel statistics before any vision-language model runs.

Pixel separability shows that labels correlate with image statistics, which a monitor can exploit without understanding the scene. VSCBench at \F{PixVSC} and MSSBench at \F{PixMSS} are close to chance on pixels, yet are separable by prompt-blind visual-token probes on unseen images. VLGuard at \F{EntVLGuard} bits and MMSafeAware at \F{EntMMSafe} bits have nonzero conditional entropy but separate on pixels. Each audited suite thus admits an image-side shortcut, so high accuracy cannot separate a monitor that reads the request from one that reads the image alone. Section~\ref{sec:positivecontrol} removes the image shortcut with same-image pairs, and Section~\ref{sec:quad} builds a control that requires both modalities.

%% file: tables/tab_pools_new.tex
% Table 2: Structural audit of public VLM safety benchmarks and both diagnostic controls.
\begin{tabular}{@{}l r r r r r r r l@{}}
\toprule
Dataset & Samples & Images & Requests & Multi-req.\ img.\ & Pixel sep.\ & $\hat{H}(Y\mid I)$ & Img.\ ceil.\ & Primary shortcut \\
\midrule
VSCBench~\cite{vscbench}          & \F{EvVSC}     & \F{ImgVSC}     & \F{ReqVSC}     & \F{MultiVSC}     & $\F{PixVSC}$     & 0.00           & 1.00              & Visual tokens, unseen img. \\
MSSBench~\cite{mssbench}          & \F{EvMSS}     & \F{ImgMSS}     & \F{ReqMSS}     & \F{MultiMSS}     & $\F{PixMSS}$     & 0.00           & 1.00              & Identity, unseen img. \\
MOSSBench type~2~\cite{mossbench} & \F{EvMoss}    & \F{ImgMoss}    & \F{ReqMoss}    & \F{MultiMoss}    & $\F{PixMoss}$    & 0.00           & 1.00              & Pixel statistics \\
VLGuard~\cite{vlguard}            & \F{EvVLGuard} & \F{ImgVLGuard} & \F{ReqVLGuard} & \F{MultiVLGuard} & $\F{PixVLGuard}$ & \F{EntVLGuard} & \F{CeilImgVLGuard} & Pixel statistics \\
MMSafeAware~\cite{mmsafeaware}    & \F{EvMMSafe}  & \F{ImgMMSafe}  & \F{ReqMMSafe}  & \F{MultiMMSafe}  & $\F{PixMMSafe}$  & \F{EntMMSafe}  & \F{CeilImgMMSafe}  & Pixel, identity \\
\midrule
Paired intent, ours               & \F{SieiEv}    & \F{SieiPairs}  & \F{SieiReq}    & \F{SieiPairs}    & $\F{SieiPixel}^{\dagger}$  & 1.00           & \F{SieiCeilImg}   & Text-side \\
$2\times2$ scene-action, ours     & \F{QuadRecords} & \F{QuadImages} & \F{QuadRequests} & \F{QuadImages} & $\text{n/a}$   & 0.50           & \F{QuadCeil}      & None by design \\
\bottomrule
\end{tabular}

%% file: sections/03_readout.tex
\section{Controlled Joint Decodability}
\label{sec:readout}

\begin{figure*}[htbp]
  \centering
  \includegraphics[width=0.96\textwidth]{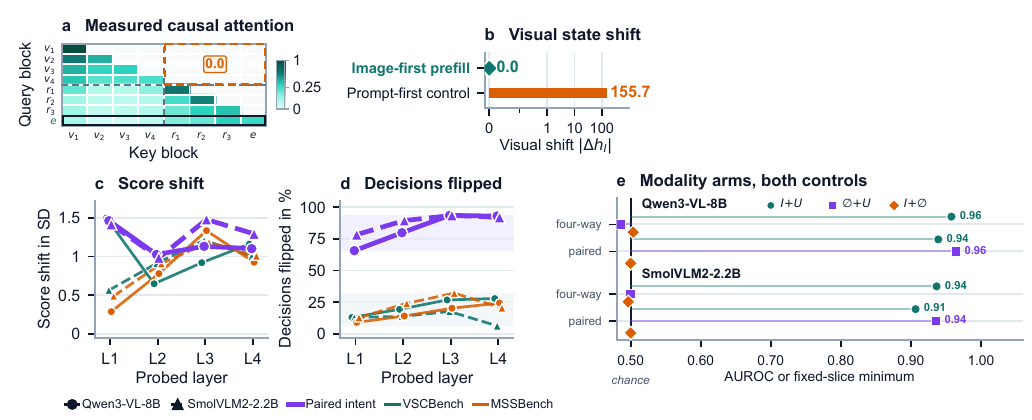}
  \vspace{-3pt}
  \caption{\textbf{Image-first causal masking makes the final prompt token the earliest state that sees both modalities, and controlled modality ablations expose which modality each evaluation suite rewards.}
  Panel a plots measured Qwen3-VL-8B attention weights over token blocks $[v_1,\ldots,v_4,r_1,\ldots,r_3,e]$: visual-to-text attention $v \to r,e$ is identically $0.0$, while the final prompt token $e$ attends across visual and textual keys.
  Panel b confirms bit-exact visual invariance $\max|\Delta h_I|=\F{GateImageFirst}$ under matched-length prompt swaps in image-first prefill versus $\F{GateRequestFirst}$ in a prompt-first control.
  Panels c and d show fixed-probe prompt interventions shifting scores by \F{MoveAllLo}--\F{MoveAllHi}\,SD and flipping \F{FlipPubLo}\%--\F{FlipPubHi}\% of predictions on public benchmarks alongside $65.7\%$--$93.4\%$ on same-image intent pairs.
  Panel e contrasts retrained modality ablations: on same-image pairs the blind text probe matches the joint probe, whereas on symmetric four-way scene-action groups the joint probe far exceeds the $50.0\%$ unimodal ceiling.}
  \label{fig:three}
  \vspace{-2pt}
\end{figure*}

\subsection{Architectural Readout Site Under Image-First Masking}
\label{sec:blind}

Testing joint information requires a state that has seen both modalities, and causal masking fixes where such a state exists, as Figure~\ref{fig:three}a shows. When projected visual tokens precede user prompt tokens, $x = [\phi(I), \tau(U)]$, every visual query precedes every text token and zeroes the visual-to-text attention block, keeping visual-token states functions of the image alone. Holding padded sequence length fixed across prompt swaps confirms exact invariance, $\max |\Delta h_I| = \F{GateImageFirst}$ across all layers and benchmarks on both backbones in Figure~\ref{fig:three}b; reversing the ordering into a \emph{prompt-first control} unmasks visual-to-text attention and alters deep visual states by up to \F{GateRequestFirst}. Under standard image-first serving, pooled visual tokens therefore give a strictly prompt-invariant baseline, whereas the final prompt token $h_{\ell, e}$ is the earliest prefill state that conditions on both the scene and the request. All joint probes below read this state.

\subsection{Fixed-Probe Prompt Interventions}
\label{sec:q2}

We first test whether $h_{\ell, e}$ responds to the prompt at all, using a fixed linear probe trained on real $(I, U)$ states. Freezing ridge parameters $(w, b)$ at $h_{\ell, e}$ and replacing real requests with length-matched neutral placeholders, or swapping requests cyclically, shifts held-out scores by \F{MoveAllLo}--\F{MoveAllHi}\,SD in Figure~\ref{fig:three}c and flips \F{FlipPubLo}\%--\F{FlipPubHi}\% of predictions on public benchmarks under request-grouped splits in Figure~\ref{fig:three}d, whereas pooled visual-token probes remain unchanged. This rules out a final-token state that ignores the text, but not that the text is useless: any change of input moves the mean projection along $w$ and can carry scores across a threshold, even though these labels do not depend on the prompt.

\subsection{Incremental Predictive Value of the Prompt}
\label{sec:q3}

Predictive value differs from sensitivity. Retraining an independent probe on request-ablated states $(I, \varnothing)$ asks how well the best probe ranks hazards without the prompt, which isolates the incremental predictive value of the prompt. We record positive prompt utility when retraining with real requests raises held-out AUROC by at least \F{GapCrit} over an ablated baseline at or below \F{AbsCrit}:
\begin{equation}
  \Delta\mathrm{AUROC} \ge \F{GapCrit},
  \qquad
  \mathrm{AUROC}_{\mathrm{retrain}}(I, \varnothing) \le \F{AbsCrit},
  \label{eq:axis2}
\end{equation}
where \F{GapCrit} exceeds fold-level AUROC variation and \F{AbsCrit} excludes benchmarks already solved by image priors. No public configuration satisfies Eq.~\ref{eq:axis2} under any split key. Under request-grouped splits, \F{GapNonPosRequest} of 16 configurations yield $\Delta\mathrm{AUROC} \le 0$ in Figure~\ref{fig:three}e, and removing the prompt raises retrained AUROC by up to \F{GainSmolMSS} on SmolVLM2 with MSSBench. Under image-disjoint splits, \F{GapNonPosImage} of 16 yield $\Delta\mathrm{AUROC} \le 0$; the four positive values are all below $0.02$, with a maximum of $+0.0167$, while negative values reach $-0.0954$, as listed in Table~\ref{tab:results}. On these suites the prompt adds no measurable ranking value beyond the image. The fixed-probe flips of Section~\ref{sec:q2} are therefore activation shifts.

\subsection{Same-Image Pairs and the Lexical Shortcut}
\label{sec:positivecontrol}

A null result on image-determined suites cannot distinguish a backbone that ignores intent from a benchmark that never rewards it, so we pair opposite-intent prompts on identical photographs. Pairing benign and harmful prompts from MOSSBench~\cite{mossbench} onto identical images verified by SHA-256 hash produces \F{SieiPairs} shared scenes and \F{SieiEv} image-prompt records in Table~\ref{tab:pools}. Because each image carries $\hat{H}(Y \mid I) = 1.00$ bit, retrained image probes score $0.5000$ AUROC by construction, while joint probes exceed them by \F{GapQwenSIEI} AUROC on Qwen3-VL-8B and \F{GapSmolSIEI} on SmolVLM2-2.2B under the case-disjoint protocol in Table~\ref{tab:results}. Both backbones encode request intent.

The same construction, however, makes the image uninformative and opens a lexical shortcut, since $I(Y; I \mid U) = 0$. A blind text probe $(\varnothing, U)$ on a uniform gray canvas at matched resolution and padded token length reaches \F{Paired0UQwenL11} AUROC at $\mathrm{L2}$ on Qwen3-VL-8B and \F{Paired0USmolL7} on SmolVLM2-2.2B, matching or exceeding the \F{PairedIUQwenL11} and \F{PairedIUSmolL7} joint probes. Multimodal synergy, $\mathrm{AUROC}(I,U) - \max\{\mathrm{AUROC}(I,\varnothing), \mathrm{AUROC}(\varnothing,U)\}$, is negative at all four depths of both backbones under both the case-disjoint split and the pair-grouped five-fold estimate, and the pair-level bootstrap interval of the five-fold estimate lies below zero in \F{PairedSynNegAll} of these eight settings; the supplement gives the per-depth values. The blind probe reaches \F{Paired0UQwenL35} AUROC at $\mathrm{L4}$ on Qwen3-VL-8B. The shortcut survives on the \F{CleanPairs} pairs free of any term from an \F{LexSize}-word hazard lexicon, where the blind probe still reaches \F{Clean0UQwenL11} and \F{Clean0USmolL7}, because syntactic framing and referent shifts separate benign from harmful requests without the scene. Same-image pairs thus cannot show that the scene is used.

\subsection{Four-Way Scene-Action Control and Intermediate-Layer Joint Decodability}
\label{sec:quad}

To remove both shortcuts at once, we construct a symmetric four-way control from MSSBench. Each chat case provides a safe scene $I_s$, a hazardous scene $I_u$, and two action queries $q$ that are unsafe on $I_u$ yet benign on $I_s$. Crossing $(I_s, I_u)$ with two benign observation queries $b$ sampled from an \F{QuadObserveBank}-query description bank yields four-way groups $(I_s, I_u) \times (q, b)$ where risk follows $Y = [\text{scene unsafe}] \wedge [\text{action query}]$, totaling \F{QuadQuads} groups across \F{QuadRecords} records, \F{QuadImages} scenes, and \F{QuadRequests} requests partitioned across \F{QuadGroups} connected image components in Table~\ref{tab:pools}.

Within any slice that holds the prompt fixed, a text rule carries no information, and within any slice that holds the scene fixed, an image rule carries none. Every unimodal monitor is therefore bounded at $50.0\%$ AUROC on its fixed-modality slice, $50.0\%$ group-ranking accuracy for placing the single unsafe case first, and $5/6 \approx 0.833$ pooled AUROC. Probing confirms these bounds at all four depths: unimodal probes score \F{QuadBlindQwenLo}--\F{QuadBlindQwenHi} fixed-slice AUROC on Qwen3-VL-8B and \F{QuadBlindSmolLo}--\F{QuadBlindSmolHi} on SmolVLM2-2.2B, and reach at most \F{QuadAccUniQwenHi} four-way ranking accuracy. Intermediate joint probes at $\mathrm{L2}$ achieve \F{QuadAccIUQwenL11} four-way ranking accuracy on Qwen3-VL-8B and \F{QuadAccIUSmolL7} on SmolVLM2-2.2B, with \F{QuadFpIUQwenL11} and \F{QuadFpIUSmolL7} AUROC under fixed prompts and \F{QuadFiIUQwenL11} and \F{QuadFiIUSmolL7} under fixed images in Figure~\ref{fig:three}e and Table~\ref{tab:results}.

This is our strongest evidence that frozen states carry information from both modalities: unlike public-suite accuracy, it cannot come from a rule that reads one modality, and unlike same-image pairs, it requires the scene as well as the intent. Comparing depths $\mathrm{L1}$ through $\mathrm{L4}$ locates the signal at intermediate layers, where joint ranking accuracy reaches \F{QuadAccIUQwenL11}--\F{QuadAccIUQwenHi} on Qwen3-VL-8B.

Decodability of the conjunction does not, however, isolate a fusion mechanism. Probes on concatenated unimodal states $[h(I,\varnothing); h(\varnothing,U)]$, which never interact inside the transformer, already reach \F{QuadAccConcatQwenL11} and \F{QuadAccConcatSmolL7} at $\mathrm{L2}$. Most of the signal is therefore linearly recoverable from separately encoded scene and intent states. Under this protocol, joint states add $+2.8$ points on Qwen3-VL-8B and $+7.1$ points on SmolVLM2-2.2B. We read this margin as additional predictive value of the joint state and do not attribute it to a particular cross-modal attention mechanism. At the final pre-generation layer $\mathrm{L4}$, joint ranking falls to \F{QuadAccIUDeepQwen} and \F{QuadAccIUDeepSmol}, below the \F{QuadAccConcatDeepQwen} and \F{QuadAccConcatDeepSmol} concatenation baselines.

\paragraph{Scope of the claim.}
The bilateral diagnostic is a dataset-level necessary condition, and joint decodability is a representation-level property of linear readouts on frozen states. Neither establishes cross-modal reasoning, semantic grounding in the hazardous object, which would need object-level interventions, or the safety behavior of the model's own responses, which we do not measure. The observation queries come from a fixed bank of \F{QuadObserveBank} templates, so the text-side results may partly depend on their wording; we have not run a held-out-template evaluation.

\paragraph{Split hygiene and shared-image leakage.}
Grouping splits by prompt text leaks image identity whenever benchmarks pair multiple requests with the same photograph. In MSSBench, \F{CrossNMSS} of \F{EvMSS} records, or \F{CrossMSS}\%, reuse a training image across splits versus \F{CrossVSC}\% in VSCBench, allowing a file byte-length lookup to recover held-out MSSBench labels at \F{ByteNativeLookup} accuracy. On same-image opposite-intent pairs, request grouping places benign and harmful prompts on opposite sides of the split, so image-memorizing probes invert to \F{PairedReqI0QwenLo}--\F{PairedReqI0QwenHi} AUROC on Qwen3-VL-8B while joint probes hold \F{PairedReqIUQwenLo}--\F{PairedReqIUQwenHi}. The deployment results below therefore rest on image-disjoint or case-disjoint splits, and request-grouped numbers serve to measure the leakage.

\begin{table*}[htbp]
  \centering
  \caption{\textbf{Summary of mechanistic checks, retrained modality ablations, and calibrated detection across benchmarks and controls.}
    Columns report ranges across depths $\mathrm{L1}$--$\mathrm{L4}$, corresponding to layers $\{4, 11, 23, 35\}$ on Qwen3-VL-8B and $\{3, 7, 15, 23\}$ on SmolVLM2-2.2B. Detection columns compare cached pooled visual-token probes against within-benchmark safety passes at a $5\%$ calibration FPR target; request-grouped MSSBench reflects \F{ArmStraddlePctQwenMSSRequest}\% shared-image overlap. Diagnostic control rows report image $(I,\varnothing)$, blind text $(\varnothing,U)$, and joint $(I,U)$ probes at $\mathrm{L2}$: test AUROC on paired intent, and four-way group-ranking accuracy against the $0.50$ unimodal ceiling. $\Delta$ is $\mathrm{AUROC}(I,U)-\mathrm{AUROC}(I,\varnothing)$ in benchmark and paired rows, and the joint pooled AUROC minus the best single-modality pooled AUROC in four-way rows.}
  {\footnotesize
  \setlength{\tabcolsep}{1.7pt}
  % The table was laid out for the two-column proceedings measure; in the
  % single-column edition it is scaled to the text width.
  \resizebox{\textwidth}{!}{%
  \begin{tabular}{@{}ll c cc cc ccc@{}}
    \toprule
    & & \textbf{Invariance} & \multicolumn{2}{c}{\textbf{Fixed-probe interv.}} & \multicolumn{2}{c}{\textbf{Retrained AUROC}} & \multicolumn{3}{c}{\textbf{5\%-target detection, cached / safety}} \\
    \cmidrule(lr){3-3}\cmidrule(lr){4-5}\cmidrule(lr){6-7}\cmidrule(l){8-10}
    Model & Dataset, split & $\max|\Delta h_I|$ & Shift, SD & Flipped, \% & Joint $(I,U)$ & $\Delta$ & AUROC & FPR, \% & TPR, \% \\
    \midrule
    Qwen3-VL & VSCBench, image  & \F{GateQwenVSC}  & \F{MoveAblQwenVSCImage}  & \F{FlipSwapQwenVSCImage}  & \F{RealQwenVSCImage}  & \F{GapKeyQwenVSCImage}  & \F{ArmAuReadQwenVSCImage} / \F{ArmAuFitQwenVSCImage} & \F{ArmFprReadQwenVSCImage} / \F{ArmFprFitQwenVSCImage} & \F{ArmReadQwenVSCImage} / \F{ArmFitQwenVSCImage} \\
    Qwen3-VL & VSCBench, request  & \F{GateQwenVSC}  & \F{MoveAblQwenVSC}  & \F{FlipQwenVSC}  & \F{RealQwenVSC}  & \F{GapQwenVSC}  & \F{AucQwenVSCRead} / \F{AucQwenVSCFit} & \F{FprQwenVSCReadPct} / \F{FprQwenVSCFitPct} & \F{TprQwenVSCReadPct} / \F{TprQwenVSCFitPct} \\
    Qwen3-VL & MSSBench, image  & \F{GateQwenMSS}  & \F{MoveAblQwenMSSImage}  & \F{FlipSwapQwenMSSImage}  & \F{RealQwenMSSImage}  & \F{GapKeyQwenMSSImage}  & \F{ArmAuReadQwenMSSImage} / \F{ArmAuFitQwenMSSImage} & \F{ArmFprReadQwenMSSImage} / \F{ArmFprFitQwenMSSImage} & \F{ArmReadQwenMSSImage} / \F{ArmFitQwenMSSImage} \\
    Qwen3-VL & MSSBench, request & \F{GateQwenMSS}  & \F{MoveAblQwenMSS}  & \F{FlipQwenMSS}  & \F{RealQwenMSS}  & \F{GapQwenMSS}  & \F{AucQwenMSSRead} / \F{AucQwenMSSFit} & \F{FprQwenMSSReadPct} / \F{FprQwenMSSFitPct} & \F{TprQwenMSSReadPct} / \F{TprQwenMSSFitPct} \\
    Qwen3-VL & Paired intent, case & \F{GateQwenSIEI} & \F{MoveAblQwenSIEI} & \F{FlipQwenSIEI} & \F{RealQwenSIEI} & \F{GapQwenSIEI} & \multicolumn{3}{c}{$(I,\varnothing)$ \F{PairedI0QwenL11}, $(\varnothing,U)$ \F{Paired0UQwenL11}, $(I,U)$ \F{PairedIUQwenL11}} \\
    Qwen3-VL & Four-way, group & $\text{n/a}$ & $\text{n/a}$ & $\text{n/a}$ & \F{QuadIUQwenL11} & $+$\F{QuadSynQwenL11} & \multicolumn{3}{c}{$(I,\varnothing)$ \F{QuadAccI0QwenL11} \quad $(\varnothing,U)$ \F{QuadAcc0UQwenL11} \quad $(I,U)$ \F{QuadAccIUQwenL11}} \\
    \midrule
    SmolVLM2 & VSCBench, image  & \F{GateSmolVSC}  & \F{MoveAblSmolVSCImage}  & \F{FlipSwapSmolVSCImage}  & \F{RealSmolVSCImage}  & \F{GapKeySmolVSCImage}  & \F{ArmAuReadSmolVSCImage} / \F{ArmAuFitSmolVSCImage} & \F{ArmFprReadSmolVSCImage} / \F{ArmFprFitSmolVSCImage} & \F{ArmReadSmolVSCImage} / \F{ArmFitSmolVSCImage} \\
    SmolVLM2 & VSCBench, request  & \F{GateSmolVSC}  & \F{MoveAblSmolVSC}  & \F{FlipSmolVSC}  & \F{RealSmolVSC}  & \F{GapSmolVSC}  & \F{AucSmolVSCRead} / \F{AucSmolVSCFit} & \F{FprSmolVSCReadPct} / \F{FprSmolVSCFitPct} & \F{TprSmolVSCReadPct} / \F{TprSmolVSCFitPct} \\
    SmolVLM2 & MSSBench, image  & \F{GateSmolMSS}  & \F{MoveAblSmolMSSImage}  & \F{FlipSwapSmolMSSImage}  & \F{RealSmolMSSImage}  & \F{GapKeySmolMSSImage}  & \F{ArmAuReadSmolMSSImage} / \F{ArmAuFitSmolMSSImage} & \F{ArmFprReadSmolMSSImage} / \F{ArmFprFitSmolMSSImage} & \F{ArmReadSmolMSSImage} / \F{ArmFitSmolMSSImage} \\
    SmolVLM2 & MSSBench, request & \F{GateSmolMSS}  & \F{MoveAblSmolMSS}  & \F{FlipSmolMSS}  & \F{RealSmolMSS}  & \F{GapSmolMSS}  & \F{AucSmolMSSRead} / \F{AucSmolMSSFit} & \F{FprSmolMSSReadPct} / \F{FprSmolMSSFitPct} & \F{TprSmolMSSReadPct} / \F{TprSmolMSSFitPct} \\
    SmolVLM2 & Paired intent, case & \F{GateSmolSIEI} & \F{MoveAblSmolSIEI} & \F{FlipSmolSIEI} & \F{RealSmolSIEI} & \F{GapSmolSIEI} & \multicolumn{3}{c}{$(I,\varnothing)$ \F{PairedI0SmolL7}, $(\varnothing,U)$ \F{Paired0USmolL7}, $(I,U)$ \F{PairedIUSmolL7}} \\
    SmolVLM2 & Four-way, group & $\text{n/a}$ & $\text{n/a}$ & $\text{n/a}$ & \F{QuadIUSmolL7} & $+$\F{QuadSynSmolL7} & \multicolumn{3}{c}{$(I,\varnothing)$ \F{QuadAccI0SmolL7} \quad $(\varnothing,U)$ \F{QuadAcc0USmolL7} \quad $(I,U)$ \F{QuadAccIUSmolL7}} \\
    \bottomrule
  \end{tabular}}
  }
  \label{tab:results}
  \vspace{-2pt}
\end{table*}

%% file: sections/04_transfer_cost.tex
\section{From Decodability to Reliable Screening}
\label{sec:deployment}

\begin{figure*}[htbp]
  \centering
  \includegraphics[width=0.78\textwidth]{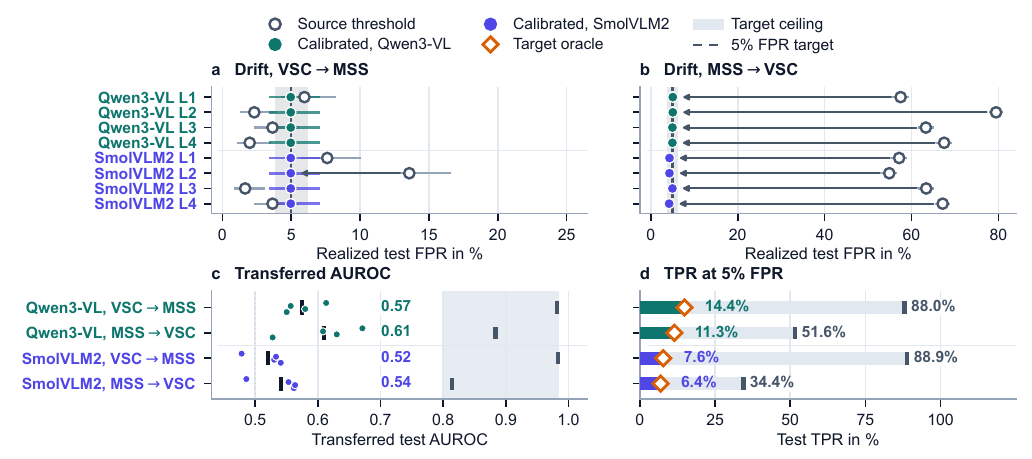}
  \vspace{-3pt}
  \caption{\textbf{Cross-benchmark transfer fails through ranking collapse, which threshold recalibration cannot repair.}
  In panels a and b, applying a source-calibrated $5\%$-FPR threshold across VSCBench and MSSBench over $\mathrm{L1}$--$\mathrm{L4}$ inflates hollow-circle FPR up to $79.5\%$ on MSSBench$\to$VSCBench. Recalibrating at the $95$th percentile of labeled benign target scores with fixed probe weights places all 16 solid-circle points at $4.2\%$--$5.1\%$ FPR, shown with $95\%$ Wilson intervals; because the same benign rows set and evaluate the threshold, this FPR is in-sample by construction.
  In panel c, transferred AUROC drops to $0.48$--$0.67$, well below the $0.80$--$0.98$ target ceiling.
  In panel d, percentile-calibrated recall at $5\%$ FPR reaches $6.4\%$--$14.4\%$ TPR, and oracle-threshold recall shown as diamonds reaches $6.9\%$--$14.8\%$ TPR, far below the $34.4\%$--$88.9\%$ target ceiling bars.}
  \label{fig:transfer}
  \vspace{-2pt}
\end{figure*}

\subsection{Why Cross-Benchmark Transfer Fails}
\label{sec:transfer}

Section~\ref{sec:quad} showed that frozen states can carry the information a situational label requires; we now ask whether a probe on those states screens inputs it has not seen. Transfer between public benchmarks can fail through two mechanisms. Under \emph{threshold shift}, domain shift displaces the scalar cutoff $\kappa_\beta$ relative to benign target traffic. Under \emph{ranking collapse}, the fixed source weight vector $w$ no longer orders target hazards above benign inputs. The first can be repaired by moving the threshold; the second cannot. Applying unadjusted source cutoffs across all 16 model-direction-layer configurations produces a directional asymmetry: realized test FPR spans $2.0\%$--$79.5\%$ on Qwen3-VL and $1.7\%$--$67.2\%$ on SmolVLM2, staying within $1.7\%$--$13.6\%$ on VSCBench$\to$MSSBench but jumps to $55.0\%$--$79.5\%$ on MSSBench$\to$VSCBench, shown as hollow circles in Figure~\ref{fig:transfer}a,b. Recalibrating $\kappa_{0.05}$ at the $95$th percentile of benign target scores brings realized FPR to $4.2\%$--$5.1\%$ in all 16 configurations, within the $2\%$--$10\%$ criterion recorded for this analysis. This repair needs target examples labeled benign and reuses the rows on which FPR is evaluated, so it shows that the shift is a scalar offset and is no independent estimate of deployed FPR.

Because a monotone change of threshold preserves score ordering, scalar recalibration cannot rescue collapsed hazard ranking~\cite{calibration,learnthentest}. A direction fitted to the image priors of one suite need not rank the hazards of another. Within our evaluated linear probes, backbones, and benchmarks, transferred AUROC falls to \F{TrAucQwen} on Qwen3-VL and \F{TrAucSmol} on SmolVLM2 against the $0.80$--$0.98$ target-trained ceiling in Figure~\ref{fig:transfer}c. Four-layer mean recall at the $5\%$ FPR target stays at \F{CalTprRankPctSmolMV}\%--\F{CalTprRankPctQwenVM}\% after benign quantile recalibration and \F{CalTprOraclePctSmolMV}\%--\F{CalTprOraclePctQwenVM}\% even under an \emph{oracle threshold} chosen on target test labels, against \F{CalCeilingPctSmolMV}\%--\F{CalCeilingPctSmolVM}\% for target-trained probes in Figure~\ref{fig:transfer}d. Within this evaluation, no recalibration of the source probe approached target-trained recall, so screening a new benchmark reliably would require target-domain training data as well as a recalibrated threshold.

\subsection{Marginal and End-to-End Serving Latency}
\label{sec:cost}

\begin{figure*}[htbp]
  \centering
  \includegraphics[width=0.80\textwidth]{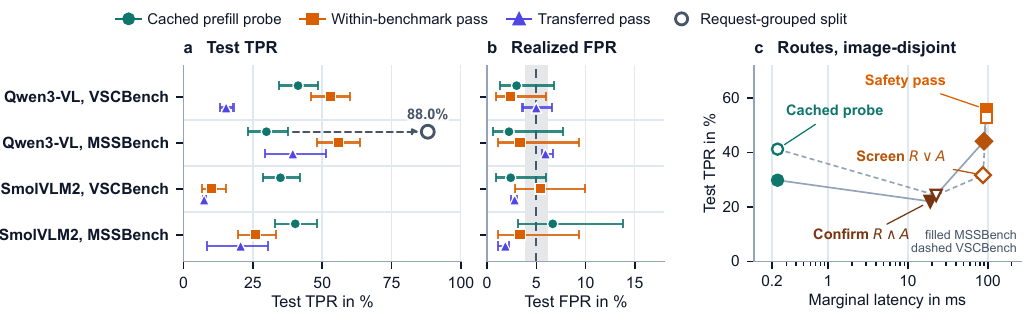}
  \vspace{-3pt}
  \caption{\textbf{Image-disjoint splits remove shared-image leakage, and cached probing anchors the low-latency deployment frontier.}
  Realized test TPR in panel a and realized test FPR in panel b show $95\%$ Wilson intervals under a $5\%$ calibration FPR target on image-disjoint splits: cached probes lead on both SmolVLM2 suites, within-benchmark safety passes lead on both Qwen3-VL suites, and the hollow circle marks the inflated \F{ArmReadQwenMSSRequest}\% recall produced by request-grouped leakage on Qwen3-VL with MSSBench. Panel c plots the four Qwen3-VL-8B screening routes from Table~\ref{tab:cascade} against batch-$1$ marginal latency, spanning \F{CostQwenReadout}\,ms for cached probing to \F{CostQwenIndependent}\,ms for an independent safety pass.}
  \label{fig:cost}
  \vspace{-2pt}
\end{figure*}

As the median of eight batch-$1$ runs on an NVIDIA H20 GPU, a cached probe adds \F{CostQwenReadout}\,ms on Qwen3-VL-8B and \F{CostSmolReadout}\,ms on SmolVLM2-2.2B, against \F{CostQwenIndependent}\,ms and \F{CostSmolIndependent}\,ms for an independent safety pass and \F{CostQwenKvfork}\,ms and \F{CostSmolKvfork}\,ms with visual KV-cache reuse. In a separate end-to-end run the probe adds \F{TpReadFracQwen}\% and \F{TpReadFracSmol}\% to prefill; the supplement gives every batch size. Low latency is evidence of feasibility, not of reliability.

\subsection{Detection under Matched Calibration Protocols}
\label{sec:matched_detection}

Detection on public benchmarks uses a different probe from Section~\ref{sec:quad}. The final-token joint probe is a diagnostic of joint decodability on the four-way control. On public benchmarks, whose labels attach to images, we deploy the prompt-invariant pooled visual-token probe, because Section~\ref{sec:q3} found that the prompt adds no ranking value there. These numbers measure use of the benchmarks' image signal, not joint scene-intent information. Table~\ref{tab:results} and Figure~\ref{fig:cost}a,b compare cached probes against safety-prompted passes under a matched $5\%$ calibration FPR target. Under request-grouped splits, safety passes transferred across benchmarks drop to \F{TprSmolVSCXferPct}\%--\F{TprQwenMSSXferPct}\% recall, so an explicit policy rubric does not make a probe domain-invariant either.

Under image-disjoint splits, the accuracy-latency trade-off depends on the backbone. On SmolVLM2-2.2B, cached probes outperform within-benchmark safety passes on both benchmarks, reaching \F{ArmReadSmolVSCImage}\% versus \F{ArmFitSmolVSCImage}\% TPR on VSCBench and \F{ArmReadSmolMSSImage}\% versus \F{ArmFitSmolMSSImage}\% on MSSBench. On Qwen3-VL-8B, within-benchmark safety passes achieve higher recall on unseen images, reaching \F{ArmFitQwenVSCImage}\% versus \F{ArmReadQwenVSCImage}\% TPR on VSCBench and \F{ArmFitQwenMSSImage}\% versus \F{ArmReadQwenMSSImage}\% on MSSBench at \F{SpeedupQwenKv}$\times$ the probe's marginal latency with KV-cache reuse and \F{SpeedupQwenIndep}$\times$ with an independent pass. Request-grouped splits hide this trade-off: they let \F{ArmStraddlePctQwenMSSRequest}\% of MSSBench records share an image across splits, which raises Qwen3-VL cached-probe recall from \F{ArmReadQwenMSSImage}\% to \F{ArmReadQwenMSSRequest}\%. Here the threshold is set on a separate calibration split, and the $5\%$ target yields $2.2\%$--$6.7\%$ realized test FPR, within the Wilson intervals of Figure~\ref{fig:cost}b given $264$--$343$ benign calibration examples.

\subsection{Two-Stage Cascades on Image-Disjoint Splits}
\label{sec:cascade}

\begin{table}[htbp]
  \centering
  \caption{\textbf{Single-stage screening versus two-stage cascades on image-disjoint splits.}
  $R$ denotes the cached prefill probe and $A$ the within-benchmark safety pass. Stage thresholds are selected strictly on calibration splits; top rows report realized test TPR / FPR in percent.}
  {\footnotesize
  \setlength{\tabcolsep}{3.0pt}
  \input{tables/tab_cascade_disjoint}}
  \label{tab:cascade}
  \vspace{-2pt}
\end{table}

Selective cascades~\cite{frugalgpt,selective} invoke the safety pass $A$ on flagged inputs, $R \wedge A$, or unflagged inputs, $R \vee A$, of the cached probe $R$, with stage thresholds selected on calibration splits under a $5\%$ calibration FPR constraint. As Table~\ref{tab:cascade} and Figure~\ref{fig:cost}c show, confirmation lowers realized FPR to \F{CascConfFprSmolVSC}\%--\F{CascConfFprQwenMSS}\% but cuts recall below standalone cached probing, while screening escalates \F{CascScreenEscSmolMSS}\%--\F{CascScreenEscQwenMSS}\% of traffic and still stays below the stronger single monitor.

%% file: tables/tab_cascade_disjoint.tex
% Table 5: single-stage routes versus two-stage cascades, image-disjoint split.
% Stage cutoffs are selected on the calibration split and never on test.
\begin{tabular}{@{}l c c c c@{}}
\toprule
Setting & Cached $R$ & $R \wedge A$ & $R \vee A$ & Pass $A$ \\
\midrule
Qwen3-VL, VSCBench & \F{CascCachedTprQwenVSC} / \F{CascCachedFprQwenVSC} & \F{CascConfTprQwenVSC} / \F{CascConfFprQwenVSC} & \F{CascScreenTprQwenVSC} / \F{CascScreenFprQwenVSC} & \F{CascAuditTprQwenVSC} / \F{CascAuditFprQwenVSC} \\
\quad latency, ms & \F{CostQwenReadout} & \F{CascConfCostQwenVSC} & \F{CascScreenCostQwenVSC} & \F{CascAuditCostQwenVSC} \\
Qwen3-VL, MSSBench & \F{CascCachedTprQwenMSS} / \F{CascCachedFprQwenMSS} & \F{CascConfTprQwenMSS} / \F{CascConfFprQwenMSS} & \F{CascScreenTprQwenMSS} / \F{CascScreenFprQwenMSS} & \F{CascAuditTprQwenMSS} / \F{CascAuditFprQwenMSS} \\
\quad latency, ms & \F{CostQwenReadout} & \F{CascConfCostQwenMSS} & \F{CascScreenCostQwenMSS} & \F{CascAuditCostQwenMSS} \\
\midrule
SmolVLM2, VSCBench & \F{CascCachedTprSmolVSC} / \F{CascCachedFprSmolVSC} & \F{CascConfTprSmolVSC} / \F{CascConfFprSmolVSC} & \F{CascScreenTprSmolVSC} / \F{CascScreenFprSmolVSC} & \F{CascAuditTprSmolVSC} / \F{CascAuditFprSmolVSC} \\
\quad latency, ms & \F{CostSmolReadout} & \F{CascConfCostSmolVSC} & \F{CascScreenCostSmolVSC} & \F{CascAuditCostSmolVSC} \\
SmolVLM2, MSSBench & \F{CascCachedTprSmolMSS} / \F{CascCachedFprSmolMSS} & \F{CascConfTprSmolMSS} / \F{CascConfFprSmolMSS} & \F{CascScreenTprSmolMSS} / \F{CascScreenFprSmolMSS} & \F{CascAuditTprSmolMSS} / \F{CascAuditFprSmolMSS} \\
\quad latency, ms & \F{CostSmolReadout} & \F{CascConfCostSmolMSS} & \F{CascScreenCostSmolMSS} & \F{CascAuditCostSmolMSS} \\
\bottomrule
\end{tabular}

%% file: sections/05_conclusion.tex
\section{Discussion and Conclusion}
\label{sec:conclusion}

High accuracy on a situational safety benchmark shows that a monitor separates that benchmark's labels, not that it uses the visual and textual information the judgment requires; in the suites we audited, one modality carries the label. By making every single-modality rule fail by construction, the four-way control shows that frozen intermediate states of both backbones carry the scene-intent conjunction in a linearly readable form. The concatenation baseline sets the limit of this evidence: separately encoded image and text states already recover most of the signal, so decodability does not reveal an internal fusion mechanism, object-level grounding, causal reasoning, or safe model behavior.

Reliability is a separate property: cached probes are inexpensive, yet recall falls on unseen images, a threshold moved to a new benchmark needs labeled benign examples, and transferred probes lose their ranking of hazards. Within our scope of two backbones, linear probes, and fixed observation templates, future evaluation should measure modality dependence of the labels, the joint predictive value of the representation over unimodal and concatenated baselines, generalization across templates and scenes, leakage-free splits, and cross-domain reliability. Reliable multimodal safety evaluation needs evidence that models use the visual and textual information a judgment requires, beyond numbers that shortcuts can explain.

%% file: sections/supp_body.tex
% THE SUPPLEMENTARY BODY.
%
% The main paper carries the argument and the measurements that decide it.
% This document carries the detail that a reader needs in order to CHECK the
% main paper, but that does not have to be read in order to follow it.
%
% The tables and figures here are generated by the same script, from the same
% JSON artifacts, as the main paper's. The two documents therefore cannot
% disagree, and a reader who checks one number in the main paper and one
% number here is checking the same artifact.
%
% THE CROSS-REFERENCE CONVENTION. Sections that live in both documents keep
% their main-paper numbers, so that a reference such as \cref{sec:calib}
% resolves in either document. The main paper's labels are carried over here
% by the \label commands at the head of the corresponding sections.

\section{Related Work}
\label{sec:Srelated}

\paragraph{Internal probes and safety heads.}
Probing hidden states of language models for safety-relevant properties has a long
history in the interpretability literature \cite{prober,shortcut}, and the
representation-engineering line has shown that such directions can be read and
controlled \cite{repreng,iti}. Work on multimodal models has established that
the visual pathway itself carries systematic shortcomings
\cite{visionlan,multimodal_prompt}. Our own earlier measurements are
representative: a linear probe on the final-layer state of an assistant's own
forward pass predicts situational risk well on the source distribution, and the
interesting question we address here is what that number means. Prior
work along this line commonly reports the probe's accuracy without establishing
that the probe is reading \emph{context} and ignores lower-level image properties. This is the single most common way such results are misread, and we
give it a name and an executable form.

\paragraph{Efficiency-oriented guards, and published multimodal guards.}
A second line of work reduces the cost of safety checking by sharing computation
with the serving model: reusing a prefill, forking a KV cache, or re-using an
encoder's activations. Our cost comparison sits in this family. The most natural
control runs the same weights twice, and we adopt it, because that isolates the cost
of the second forward pass from the cost of a different architecture.
Several recent guard models are trained specifically to detect unsafe multimodal
input \cite{guardreasoner,vlmguard,singguard,llama_guard}, and a
substantial evaluation literature compares them against each other
\cite{harmbench,mmsafetybench,mmsafeaware,mossbench}. We do not compare against
them under our protocol, and we say so plainly in \cref{sec:limitations}. They
operate on the \emph{input}. They answer a different question: is this content
unsafe, where our question is whether this image is unsafe \emph{under this request}. And
their published numbers are not measured on our pools with our splits and our
operating-point constraint \cite{evalprotocols}. What our results imply about
them is narrower, and we state it as an implication and make no measurement claim.

\paragraph{Evaluation methodology, shortcut control, and operating points.}
The methodological contribution is closest to work on shortcut learning
\cite{shortcut} and on admission criteria for evaluation sets
\cite{benchmarks,evalprotocols}. The idea that a benchmark must be \emph{audited}
for the confound it admits before its numbers are interpreted is standard in causal
inference \cite{causal} and in ML evaluation more broadly; we apply it to
same-request multimodal safety pools and give it an executable form, with
thresholds fixed in the analysis scripts \cite{wilson,bootstrap,clustered,permutation}. Because
the deployment premise is that human review is the fallback for flagged content,
the operating point matters more than the ranking; we therefore report the
false-positive rate measured out of sample at a fixed target, and we use the
distribution-free quantile construction \cite{distributional,quantile} for the one
repair that transfers across pools.

\section{Evaluation Protocol and Causal Invariance Validation}
\label{sec:suppcheck}

% ---------------------------------------------------------------------------
% CROSS-REFERENCE STUBS.
%
% Sections that live in BOTH documents keep their main-paper numbers, so that
% a reference such as \cref{sec:calib} resolves in either document. The main
% paper's labels are carried over here by the \label commands at the head of
% the corresponding sections. The handful of labels below point INTO the main
% paper from this document; they are declared here as empty stubs so that the
% cross-references resolve, and the reader is directed to the main paper for
% the content itself. The text after each stub names the main-paper section.
% ---------------------------------------------------------------------------
% The main-paper sections live in this same document, so the forwarding stubs
% that the two-document edition needed are no longer required.

\subsection{Retrained prompt ablation at every layer}

The main paper reports the axis-2 control at the layer most favourable to it;
this table reports every layer, for every pool and family.

\begin{table}[htbp]
\centering
\caption{The label structure of every pool, measured on the labels alone. The
conditional entropy of the risk label given the image and given the request, in
bits, and the ceiling of the best possible rule that reads the image alone or
the request alone. $H(Y\mid I)=\num{0.0000}$ means the label is a deterministic
function of the image, so the request carries no information about it and the
axis-2 control of \cref{eq:axis2} cannot be passed on that pool by any
read-out. The last column counts the images that carry both labels, which is
the raw material of a same-image/different-intent pair and is therefore the
measure of whether a contextual pool could be built from the data at all.}
\label{tab:labelstructure}
\vspace{-2pt}
\footnotesize
\input{tables/tab_label_structure}
\vspace{-4pt}
\end{table}

\begin{table}[htbp]
\centering
\caption{The positive control, layer by layer. The identical axis-2 protocol,
run on a pool built from same-image/different-intent pairs, where the request
determines the label and the image does not. The ablated read-out collapses to
exactly \num{0.5000} and the request-blind position sits at \num{0.5000}, so the
control passes here. This is what makes the negative results on the other pools
meaningful: the control is capable of passing, and it does so when the pool
carries the structure it is designed to detect.}
\label{tab:positivecontrol}
\vspace{-2pt}
\footnotesize
\input{tables/tab_positive_control}
\vspace{-4pt}
\end{table}

\begin{table}[htbp]
\centering
\caption{The admission audit of every pool we could obtain, with the reason for
each verdict. Axis 1 is the request-internal pixel control; axis 2 is the
request-ablation control, reported as the drop in \auroc when the request is
removed. The last column names the reason as well as the verdict.}
\label{tab:poolsaudit}
\vspace{-2pt}
\footnotesize
\input{tables/tab_pools}
\vspace{-4pt}
\end{table}

\begin{table}[htbp]
\centering
\caption{The standard read-out location is request-blind. Within-pair relative
distance between the two members of a same-image/different-intent pair, in the
standard image-first prompt order. The image-token distance is three to four
orders of magnitude below the request-endpoint distance.}
\label{tab:blind}
\vspace{-2pt}
\footnotesize
\begin{tabular}{llccc}
\toprule
family & layer & image-token & request endpoint & ratio \\
\midrule
\multirow{4}{*}{Qwen3-VL-8B} & 0 & \num{1.434e-04} & \num{5.521e-02} & \num{0.0026} \\
 & 1 & \num{3.002e-04} & \num{1.014e-01} & \num{0.0030} \\
 & 2 & \num{6.599e-04} & \num{5.036e-01} & \num{0.0013} \\
 & 3 & \num{3.043e-03} & \num{7.527e-01} & \num{0.0040} \\
\midrule
\multirow{4}{*}{SmolVLM2-2.2B} & 0 & \num{0.000e+00} & \num{1.411e-01} & \num{0.0000} \\
 & 1 & \num{0.000e+00} & \num{2.241e-01} & \num{0.0000} \\
 & 2 & \num{0.000e+00} & \num{3.958e-01} & \num{0.0000} \\
 & 3 & \num{0.000e+00} & \num{4.169e-01} & \num{0.0000} \\
\bottomrule
\end{tabular}
\vspace{-4pt}
\end{table}

\begin{table*}[htbp]
\centering
\caption{The two serving state positions, on the same samples, the same
request-level splits and the same probe family. \fpr and \tpr are measured out of
sample on the test split at a \num{0.05} target; \auroc is the threshold-free
companion and $R_{\mathrm{pair}}$ is the paired separation rate on the test
comparisons. The best layer per position is shown.}
\label{tab:columns}
\vspace{-2pt}
\footnotesize
\input{tables/tab_columns}
\vspace{-4pt}
\end{table*}

% This table is long enough to occupy a page on its own, so it is set as a
% breakable long table: the body text of the section runs on beside it and
% the header repeats on every page it spans.  The size settings wrap the
% whole table, because a long table is built one page-chunk at a time and
% every chunk has to be set at the same size.
{\footnotesize
\setlength{\tabcolsep}{4.5pt}
\renewcommand{\arraystretch}{1.05}
\begin{longtable}{@{}llcccccc@{}}
\caption{The axis-2 admission control at every layer. \auroc is measured out of
sample on the test split at the prompt-end position. \emph{real} is the serving
prompt, \emph{ablated} replaces the request text with a fixed placeholder,
\emph{swapped} replaces it with another request's text, and \emph{image-tok} is
the request-blind position on the real prompt.}
\label{tab:axis2full}\\
\toprule
family & pool & layer & real & ablated & swapped & image-tok & axis 2 \\
\midrule
\endfirsthead
\toprule
family & pool & layer & real & ablated & swapped & image-tok & axis 2 \\
\midrule
\endhead
\bottomrule
\endfoot
\input{tables/tab_axis2}
\end{longtable}
}

\subsection{The causal invariance check, in full}

\cref{tab:axis2check} reports the check that verifies the extraction itself. Under
causal attention in the image-first order the image-token states precede the
request, so ablating or swapping the request must leave them bit-identical; in
the request-first order the image tokens follow the request, so the ablation must
move them. The two orders give opposite signs, and a check with two signs can
fail in two ways.

\begin{table}[htbp]
\centering
\caption{The causal invariance check on the extraction itself, in both prompt orders and
for every pool and family. The two numeric columns are the maximum absolute
difference between the real condition and the ablated one, and between the real
condition and the swapped one, at the shallowest probe layer, which is where the
movement is smallest.}
\label{tab:axis2check}
\vspace{-2pt}
\footnotesize
\input{tables/tab_axis2_gate}
\vspace{-4pt}
\end{table}

\begin{figure}[htbp]
\centering
\includegraphics[width=0.62\linewidth,height=0.56\textheight,keepaspectratio]{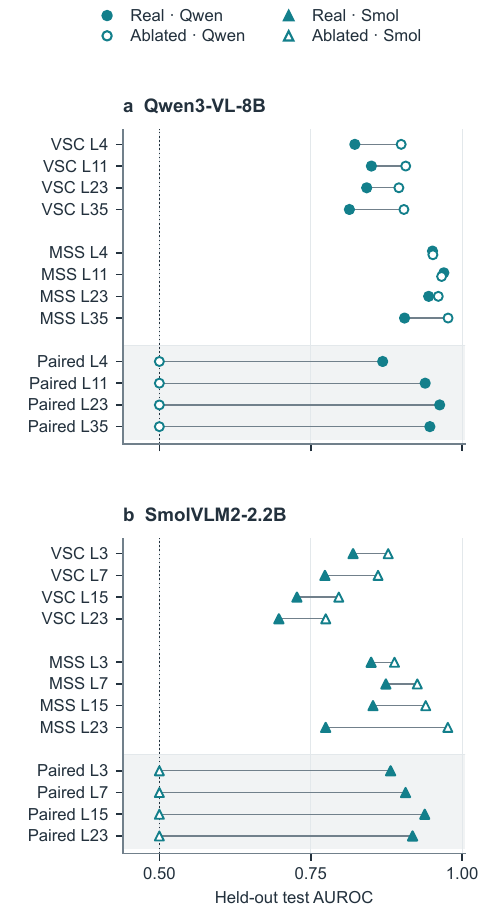}
\caption{Real and ablated prompt-end readouts at every measured layer, providing the layer-wise values behind the main paper's retrained-AUROC comparison. Each segment connects separately fitted arms; filled symbols indicate real requests and open symbols ablated requests. Circles denote Qwen and triangles Smol. The dotted line marks chance. Gray bands separate the constructed paired-intent control from the public pools. The public-pool ablated readouts remain strong, while the paired-intent control falls to chance. Its labels encode construction provenance; these results do not establish adjudicated situational safety. No numeric pixel-admission threshold is inferred from these measurements.}\label{fig:twoaxis}
\vspace{-4pt}
\end{figure}

\subsection{Why the check must be computed within one pass}

The check is computed \emph{within a single extraction pass}, and it has to be. A
check computed across two runs compares the model \emph{and} two
bookkeeping conventions, and we had both episodes of that failure: a reshape
that filed the states under the wrong sample identifiers, and a sequence-length artifact
that made one architecture look as though the request reached the image tokens.
The measurement was discarded and re-run with all three conditions extracted in
one process, in one code path, with one batching and one common length, and the
check is then exactly \num{0.000e+00} for both families. The check must therefore be computed within one run, and a reproduction of the ablation must follow the same procedure. Both episodes are
reported in full in \cref{sec:batchinv} and \cref{sec:lengthbug}.

\subsection{The layout check that runs before the analysis, and its own control}

The check above verifies that the three conditions are the right quantity. It does
not verify that each row of the saved tensor is filed under the sample its
row identifier names, which is the defect that actually reached the
tables. A separate check answers that question, and it runs \emph{before} the
analysis: the pipeline does not compute a single number until every extraction
file has passed it.

The check is order-aware, and it has to be. Under causal attention the two
prompt orders impose opposite requirements on the image-token state: in the
image-first order two samples that share an image must share their state
\emph{exactly}, because the request cannot reach the image tokens; in the
request-first order the request precedes them, so the same two samples legitimately
differ. The check therefore evaluates the same-image agreement on the
order-appropriate condition, and compares each file against the reference of
\emph{its own} order.

On the paired opposite-intent control a third test is needed, because the first is
vacuous there: every image of that pool belongs to exactly one pair, so the
same-image agreement is zero for any tensor, aligned or scrambled. The third
test is the pair disagreement: the two members of a pair share the image
bit-for-bit and differ in their request, so in the request-first order their
states must differ, and a scrambled tensor destroys that difference.

\paragraph{The check carries its own negative control.} A check that cannot fail
is not a check. On every file the check is also run against a deliberately
scrambled copy of the tensor, and the scrambled copy must be rejected. All
\num{12} files pass and all \num{12} scrambled controls are rejected, so the check
demonstrates its discriminating power on the very data it judges; it does not
rest on an assertion in a comment. The first edition of this check applied the
image-first test to the request-first files as well and therefore reported
correctly aligned files as corrupt; and a first attempt at the paired-intent test used
an absolute threshold on the pair disagreement, which was void because the
scrambled control fell under it. Both were fixed before the check was used to
check anything, and the corrected measurement is what the numbers above rest on.

\subsection{The state does not depend on the batch}

The pipeline extracts the three conditions of a sample in one batch, and one
might reasonably ask whether that choice is what makes the check exactly zero.
It is not. We extracted the same 64 samples four ways: alone, with their
file-order neighbours, with their request-group neighbours, and with a shuffled
partner, and the image-token state was bit-identical in all four cases at every
probe layer, with \num{0} of \num{64} samples differing at any layer. The state is
a function of the sample, and the protocol's within-pass design is a consequence of that fact and no workaround for its absence. The full measurement is
in \cref{sec:batchinv}.

\section{Validation of Feature Column Semantics}
\label{sec:protocolfixlabel}

An earlier artifact contained a control that appeared
self-contradictory. It reported, on the VSCBench same-request pool, a column
captioned \emph{``the request column must sit at chance on this pool, because
the request is byte-identical within every pair''}, and then printed
\rng{\num{0.83}}{\num{0.96}} for that column, while a separate lexical control on
the same pool measured \num{0.4426}. Both cannot be the same quantity.

Reading the extractor resolved it. The cache's column contract lists seven
feature kinds: \texttt{visual}, \texttt{request}, \texttt{mean8},
\texttt{mean32}, \texttt{mean128}, \texttt{meanfull} and \texttt{lastfull}, and the summariser that produces them defines the kind named
\texttt{request} as the hidden state at the \emph{last prompt token}. It is a
state of the whole prompt and is no embedding of the request string. Because the
two members of a same-request pair carry different images, and the last prompt
token attends to the image tokens before it, that state separates the pair by
\emph{image} semantics.

So there was no contradiction: the earlier number was an image separation and the
lexical control was a text separation, and the caption had described the latter
while the table printed the former. The correction matters beyond bookkeeping,
because it relocates the paper's central measurement from a position that is
provably request-blind to one that is request-conditioned by construction, and
because it shows that a column's \emph{name} in an artifact is not a guarantee of
what the column \emph{is}. We now generate every table in this paper directly from
the JSON artifacts by script, so that no caption can drift from a column again.

\section{Full Transfer Decomposition}
\label{sec:Stransfer}

\subsection{How we measure transfer}

A reader fitted on one pool and applied to another fails for two distinguishable
reasons, and conflating them is the most common error in this literature:
\begin{enumerate}
\itemsep1pt
\item the \textbf{threshold} may not transfer, because a score fitted on pool A
  has no reason to sit at the same quantile of pool B's benign distribution; and
\item the \textbf{representation} may not transfer, because the linear map
  fitted on pool A may not order pool B's images correctly at all.
\end{enumerate}
We report the transferred global AUROC and its realized operating point.
Within-pool ranking is a different metric from global AUROC and is not plotted as such. The threshold is calibrated on the \emph{source} pool's calibration split
and applied to the target pool \emph{unchanged}, so the target's realised \fpr and
\tpr are what a shipped detector would really do.

\subsection{Label-assisted threshold recalibration}

\label{sec:calib}
A transferred threshold fails when the source and target pools' benign score
distributions differ. We apply monotone repairs, each of which maps a score to its
position within its \emph{own} pool's benign distribution:
\begin{align}
R_0 &: \text{raw score, threshold carried over (the baseline)}, \notag\\
R_1 &: \text{standardise by the pool's benign mean and sd}, \notag\\
R_2 &: \text{standardise by the pool's benign median and IQR}, \notag\\
R_3 &: \text{rank within its own benign score pool},
      \notag\\
R_4 &: \text{oracle: recalibrate on target benign scores}.
\label{eq:repairs}
\end{align}
All repairs $R_1$--$R_4$ use target benign labels, and the same target benign rows
are reused for calibration and evaluation. Their results are descriptive and
label-assisted; a separate calibration split is needed to assess prospective
performance. The recorded success criterion is: at
the \num{0.05} target, the realised target-pool \fpr lies in $[0.02, 0.10]$
\textbf{and} the realised \tpr is at least \num{60}\% of the oracle's \tpr at the
same target. The reference is the oracle and excludes the baseline, because the
baseline reaches its high \tpr at an \fpr as high as \rng{\num{0.59}}{\num{0.63}}, that is, by
flagging almost everything, and a repair that lowers the \fpr by driving the \tpr
to zero must not count.

\paragraph{Implementation check.} Each quantity must be expressed relative to \emph{its own} pool's benign distribution before the
two are compared. An implementation that normalizes both sides by the same pool
reduces the transform to one affine map applied to both sides of
$s \ge \tau$ and leaves every decision unchanged; an earlier version did so and
produced $R_1 = R_2 = R_0$ to four decimals. All numbers reported here come from
the corrected implementation.

% A breakable long table, so that the discussion of the transfer runs on
% beside the numbers instead of waiting for a page of its own.
{\footnotesize
\setlength{\tabcolsep}{2.2pt}
\renewcommand{\arraystretch}{1.05}
\begin{longtable}{@{}llccrrrr@{}}
\caption{Cross-pool transfer: realized operating points and global AUROC. \fpr and \tpr are realised on the target pool at the
\num{0.05} target, with the threshold calibrated on the source pool. The repair columns use target benign labels, including the oracle ceiling,
and reuse those rows in evaluation. Within-pool pair-separation metrics are not comparable to global AUROC and are omitted.}
\label{tab:transfer}\\
\toprule
family & order & direction & layer idx & $R_0$ \fpr/\tpr & $R_3$ \fpr/\tpr & oracle \tpr & transfer \auroc \\
\midrule
\endfirsthead
\toprule
family & order & direction & layer idx & $R_0$ \fpr/\tpr & $R_3$ \fpr/\tpr & oracle \tpr & transfer \auroc \\
\midrule
\endhead
\bottomrule
\endfoot
\input{tables/tab_transfer}
\end{longtable}
}

In these label-assisted evaluations, the quantile threshold recalibration meets the recorded criterion in every tested configuration. Over the eight standard
image-first transfer configurations of each family, at the \num{0.05} target it realises
a mean target-pool \fpr of \num{0.0499} for Qwen and \num{0.0470} for SmolVLM2,
passing in 8/8 configurations with a mean \tpr of \num{0.1285} against the oracle's
\num{0.1315} for Qwen and \num{0.0701} against \num{0.0733} for SmolVLM2, while the
untransformed baseline passes in 2/8 configurations, with mean \fpr values of \num{0.3521} and
\num{0.3364}. Over all sixteen configurations of each family, the eight image-first
configurations and the eight request-first mirror, it succeeds under the recorded
criterion in 16/16, with a mean target-pool \fpr of \num{0.0498} for Qwen and
\num{0.0481} for SmolVLM2 and a mean realised \tpr of \num{0.1336} against the
oracle's \num{0.1367} for Qwen, and \num{0.0724} against \num{0.0748} for
SmolVLM2: \num{97.7}\% and \num{96.8}\% of the ceiling respectively. The
untransformed baseline fails the false-positive half of that criterion in 9/16
configurations for Qwen and 11/16 for SmolVLM2, with realised false positive rates as
high as \num{0.7946} and \num{0.6726}: the baseline misses the
target by an order of magnitude. These results describe calibration on the observed target benign distribution
\cite{distributional,quantile}; they do not demonstrate label-free transfer or
a prospective false-positive guarantee.

We note that the baseline does pass the criterion outright in the remaining
configurations, and we do not hide this: a transferred raw threshold is not uniformly
useless, it is \emph{unreliable}, and the unreliability is the point. A
deployment that must guarantee a false-positive rate cannot accept a recipe
whose realised rate is \num{0.79} in one configuration out of four.

\begin{figure}[htbp]
\centering
\includegraphics[width=0.60\linewidth,height=0.54\textheight,keepaspectratio]{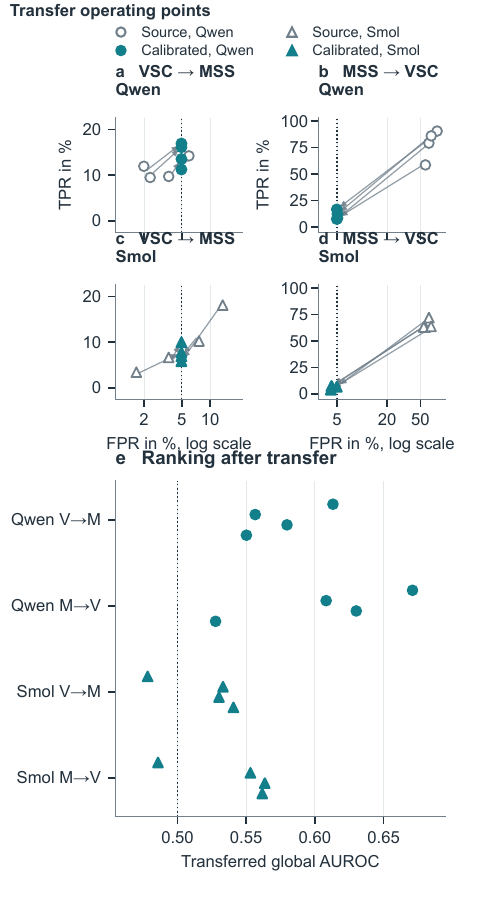}
\caption{Transfer at each of four layers in each direction and model. Panels a and b show Qwen and panels c and d Smol; within each row the left column is VSC$\rightarrow$MSS and the right column is MSS$\rightarrow$VSC. Both columns use a logarithmic FPR axis with decade ticks, but their ranges differ because the two transfer directions span different operating ranges, 1--22\% for VSC$\rightarrow$MSS and 3--100\% for MSS$\rightarrow$VSC; each column also keeps its own TPR scale. Arrows connect each layer's source-threshold operating point, shown in open gray, to its target-benign quantile threshold, shown in filled teal; dotted lines mark the 5\% target. Quantile thresholding uses target benign labels and reuses these rows in evaluation, so it is a descriptive, label-assisted result. Panel e shows global transferred AUROC, one point per layer; vertical offsets distinguish overlapping layers and the dotted line marks chance. Circles denote Qwen and triangles Smol.}\label{fig:Stransfer}
\vspace{-4pt}
\end{figure}

\subsection{Transfer discrimination and metric scope}
\label{sec:repr}
Threshold repair changes the operating point without refitting the scoring
weights. Figure~\ref{fig:Stransfer}~e reports global transferred AUROC for each
image-first layer and direction. The recorded runs's within-pool reference is computed
as the fraction of correctly ordered benign-harmful pairs within requests.
That pair-separation statistic is not global AUROC. The previously reported
0.23--0.49 difference subtracts these unlike metrics and is superseded as a
quantitative AUROC gap. A matched comparison requires both arms to be evaluated
with the same metric on aligned samples.

\subsection{The pixel-shortcut diagnostic remains descriptive}
\label{sec:shortcut}
The diagnostic residualizes transferred scores against a pixel
baseline. Its reported explained-variance values are descriptive. However,
the recovery fraction used to interpret this analysis relies on the mixed-metric
reference above. It therefore cannot support the earlier conclusion that a
particular proportion of a global AUROC gap is explained or refuted. We leave
the cause of the observed transfer performance unresolved.

\subsection{What this means in practice}
The source threshold is unreliable across the tested pools. Target-benign
calibration can place the retrospective operating point near the target FPR,
but this result uses target labels and reused evaluation rows. It is not a
prospective guarantee, and it does not establish that retraining on the target
pool can be omitted.

\section{Full Cost Measurements}
\label{sec:Scost}

\subsection{Three audit designs, and why the ratio is not the deliverable}

We compare the read-out against three ways of auditing:
\begin{description}
\itemsep1pt
\item \textbf{Independent audit.} A second forward pass of the same weights with a
  safety prompt, plus a generation. This is the most natural control: it
  isolates the cost of the second run from the cost of a different architecture.
\item \textbf{Shared-encoder audit.} The vision encoder is computed once and its
  activations are reused by the audit's prefill, which still re-runs the language
  model. This is the cheapest honest audit and the one a deployment would reach
  for first.
\item \textbf{visual KV-cache reuse pass.} The audit's prompt is appended to the serving
  prompt's KV cache, so the shared prefix is not recomputed.
\end{description}
The read-out's cost is the marginal cost of pooling the states and applying a
linear map; the serving forward pass itself is a cost the service pays either
way and is not counted against the read-out. For the shared-encoder design we also
report the vision tower's own cost separately, so that the size of the excluded
term is visible.

\subsection{Measurement discipline}

All timings are device time measured with the CUDA timer, with warm-up discarded and
the median reported alongside the interquartile range. Every ratio is computed
\emph{within one batch size}, because a ratio whose numerator and denominator come
from different batch sizes is not a ratio of anything: all latency ratios are
therefore strictly matched by batch size, so that cross-batch distortion cannot
enter the comparison. We report \textbf{absolute
milliseconds} as the deliverable and the ratios as derived quantities, with
their batch size attached.

\subsection{Results}

\begin{table*}[htbp]
\centering
\caption{Latency per request, in milliseconds, at matched batch size. The read-out
is marginal over the serving forward pass, which is a cost the service pays
either way. The audit paths have no shared term. The reader's marginal cost is
flat in the batch size because it is a pooling operation on states that already
exist.}
\label{tab:cost}
\vspace{-2pt}
\input{tables/tab_cost}
\vspace{-4pt}
\end{table*}

Across both families and all measured batch sizes, the read-out's marginal cost
is \rng{\num{0.10}}{\num{0.23}}\,ms per request, against \rng{\num{59.7}}{\num{223.4}}\,ms for an
independent audit and \rng{\num{4.8}}{\num{45.3}}\,ms for a visual KV-cache reuse pass. The
shared-encoder audit, measured separately, costs \rng{\num{51.9}}{\num{112.0}}\,ms.

\begin{figure}[htbp]
\centering
\includegraphics[width=0.62\linewidth,height=0.56\textheight,keepaspectratio]{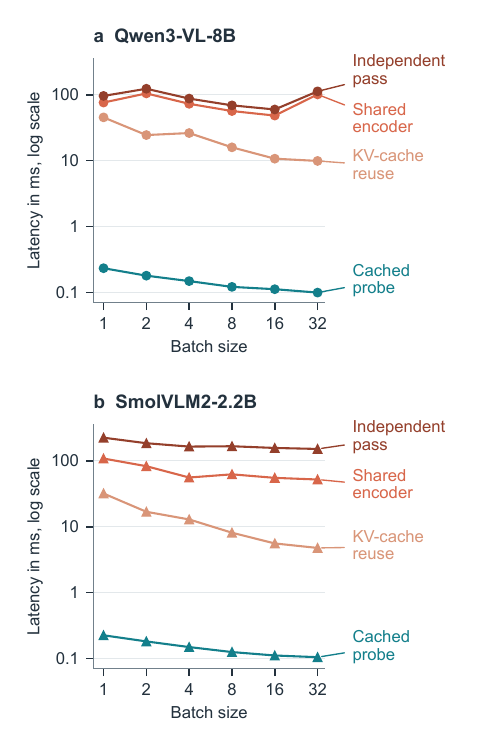}
\caption{Marginal latency per request across all measured batch sizes, separated by model. Each point is the median of eight timing repetitions; all recorded fluctuations are retained without smoothing. Teal denotes the cached readout; warm colors distinguish audit implementations, directly labeled at the line ends. Circles denote Qwen and triangles Smol. Shared-encoder measurements come from a separate timing run. The serving prefill is excluded from all paths. This figure reports latency alone; it does not assign unmatched detection measurements to audit implementations.}
\label{fig:frontier}
\vspace{-4pt}
\end{figure}

\subsection{An internal consistency check}

The cost benchmark carries a check that it can fail. If the shared-encoder design
saves exactly the vision tower's cost relative to the independent audit, then
\begin{equation}
c_{\mathrm{indep}} - c_{\mathrm{shared}} \approx c_{\mathrm{vision\text{-}tower}} ,
\label{eq:costcheck}
\end{equation}
and this is a prediction about three independently measured quantities and restates none of them. Measured across all six batch sizes and both
families, the relative error of the left-hand side against the right-hand side is
at most \num{1.87}\% and at most \num{0.84}\% for SmolVLM2, with a median error
below \num{0.5}\%. We regard this as evidence that the benchmark is measuring what
it claims, and we report it because a cost claim without an internal check is a
cost claim without evidence.

The shared-encoder path additionally carries an \emph{answer-equivalence} check:
the cached-feature path replays the model's own vision features, so the two paths
are the same model, and we verify that they produce the same generated answer on
every measured batch. If they differed, the cached path would be computing a
different model and its timing would be meaningless.

Peak GPU memory is \emph{not} a win for the read-out: the reuse path peaks at
\num{16.57}\,GiB against the audit path's \num{16.53}\,GiB, a difference of
\num{0.04}\,GiB. We report it because the memory claim is often made implicitly
and it would be easy to imply a win that the measurement does not support.

\subsection{Scope of the cost claim}

These are \emph{latency} numbers from one implementation on one node. They are not
a field-cost model: they exclude serving optimisation, batching policies, and
multi-tenant interference, all of which would change both paths. They are also
not a matched-accuracy comparison: the audit here is an unconstrained generation
whose accuracy we have not matched to the read-out's, and the comparison between
cached and safety-prompted probes of \cref{sec:fourarm} shows that the audit can be
the more sensitive of the two. A
matched-accuracy cost frontier remains open, and we flag it as such.

\subsection{Scope and Open Challenges}
\label{sec:supplimitations}

Our study covers pre-generation input risk across two VLM families, an 8B and a 2.2B model, linear probes across four depths, and single-node H20 serving; multi-turn trajectories and nonlinear heads would close the Qwen/VSCBench recall gap. The paired opposite-intent control establishes request sensitivity as a necessary condition, at $\hat{H}(Y \mid I) = 1.00$ bit and an image ceiling of \F{SieiCeilImg}, and its labels are a function of the prompt, as \cref{sec:suppbilateral} quantifies. The four-way scene-action control in the same section supplies the missing visual-necessity direction on the same instances, at the cost of constructed observation queries and an additively decodable conjunction. The quantile and oracle thresholds of \cref{fig:transfer} reuse target benign labels as upper-bound diagnostics, so practical monitors need policy-aligned supervision.

\section{Bilateral Controls: Construction, Provenance, and Every Arm}
\label{sec:suppbilateral}

\subsection{Provenance of the paired opposite-intent control}
\label{sec:supppairs}

The paired control is built from the MOSSBench type-2 file set alone. We group its
records by image SHA-256, keep an image when it carries one benign and one harmful
prompt, and verify that the two records of a pair reference byte-identical image
files. This yields \F{SieiPairs} images, \F{SieiEv} image-prompt pairs, and
\F{SieiReq} distinct requests. The pair is the unit of every case-disjoint and
image-disjoint split, so the two members always share a split. Because the same
image sits in both classes, a pixel classifier stays at \F{SieiPixel} AUROC, and
because each image carries one prompt, the label is a function of the prompt:
$\hat{H}(Y \mid I) = 1.00$ bit and $\hat{H}(Y \mid U) = 0.00$ bit.

\subsection{The modality arms and what holds them comparable}
\label{sec:supparms}

We fit four arms through one code path and one fitted protocol, so every arm shares
the ridge penalty, the split, and the cutoff rule. The image arm $(I,\varnothing)$
replaces both prompts of a pair with one fixed placeholder string. The text arm
$(\varnothing,U)$ replaces the image with a uniform mid-gray canvas. Two controls keep
an arm from reading the modality it should not see. The canvas takes the resolution
of the pair's image, so both records of a pair receive bit-identical canvases and the
same number of visual tokens, and every prompt of a pair is padded to one common
length, so the placeholder arm is bit-identical across the two prompts of a pair. We
verified both invariants directly: the largest absolute difference between the two
canvas arms of a pair is $0.0$ on both backbones.

\subsection{Modality arms and synergy on the paired control}
\label{sec:supparmtables}

\Cref{tab:supppaired} reports the image, text, and joint arms at the
final-token site. The arm columns are pair-grouped five-fold AUROC at depth
$\mathrm{L3}$, layer 23 on Qwen3-VL-8B and layer 15 on SmolVLM2-2.2B. The image arm sits at exactly
$0.500$ at every depth, which the matched image implies. The text arm tracks or
exceeds the joint arm at every depth, so the five-fold synergy
$\mathrm{AUROC}(I,U)-\max\{\mathrm{AUROC}(I,\varnothing),\mathrm{AUROC}(\varnothing,U)\}$
is negative at all four depths of both backbones; the case-disjoint single-split estimate is also negative at every depth.

\paragraph{Statistical scope of the synergy columns.} The synergy and interval
columns do not refer to the same depth as the arm columns. The synergy column
prints the most negative five-fold synergy over the four depths, which is
$\mathrm{L1}$ on Qwen3-VL-8B and $\mathrm{L2}$ on SmolVLM2-2.2B. The interval
column prints the envelope of the four per-depth pair-level bootstrap intervals,
from the smallest lower bound to the largest upper bound, and is not the interval
of any single depth. At the printed $\mathrm{L3}$ arms the five-fold synergy is
$-0.0050$ on Qwen3-VL-8B and $-0.0127$ on SmolVLM2-2.2B, and the bootstrap
interval at that depth includes zero on both backbones. The per-depth bootstrap
interval lies entirely below zero in \F{PairedSynNegAll} of the eight model-depth settings,
the count cited in the main paper. A per-depth version of this table, generated from the
same JSON report, would remove the mixed scope; we report the discrepancy here
without altering the generated values.

\begin{table*}[htbp]
  \centering
  \caption{\textbf{The three modality arms on the paired opposite-intent control.}
  Arm columns: pair-grouped five-fold AUROC at the final prompt token at depth $\mathrm{L3}$. Synergy: most negative five-fold synergy over the four depths. Bootstrap interval: envelope of the four per-depth pair-level intervals. The columns therefore do not share one depth; see the text.}
  {\footnotesize
  \setlength{\tabcolsep}{5pt}
  \begin{tabular}{@{}l ccccc@{}}
    \toprule
    Model & $(I,\varnothing)$ & $(\varnothing,U)$ & $(I,U)$ & synergy & bootstrap interval \\
    \midrule
    Qwen3-VL-8B & $0.500$ & \F{Paired0UfQwenL23} & \F{PairedIUfQwenL23} & \F{PairedSynergyQwenLo} & \F{PairedSynergyCIQwenLo} to \F{PairedSynergyCIQwenHi} \\
    SmolVLM2-2.2B & $0.500$ & \F{Paired0UfSmolL15} & \F{PairedIUfSmolL15} & \F{PairedSynergySmolLo} & \F{PairedSynergyCISmolLo} to \F{PairedSynergyCISmolHi} \\
    \bottomrule
  \end{tabular}}
  \label{tab:supppaired}
\end{table*}

\subsection{Construction of the four-way scene-action control}
\label{sec:suppquad}

The four-way control is built from the \F{QuadCases} MSSBench chat cases. Each
case supplies a safe scene, an unsafe scene, and two native action queries that are
unsafe on the unsafe scene alone. We complete each case with two observation queries
drawn from a fixed bank of \F{QuadObserveBank} requests, listed verbatim in the
released pool file, that ask for a description of the picture and express no intent
to act. The two observation queries per case are chosen by SHA-256 so the choice is
reproducible and independent of any measurement. A group therefore holds four
image-prompt pairs, exactly one of them unsafe, with
$Y = [\text{scene unsafe}] \wedge [\text{prompt expresses the act}]$. The pool holds
\F{QuadQuads} groups and \F{QuadRecords} records over \F{QuadGroups} connected image
groups, where cases that share an image are joined so that no image crosses a split.

Two facts follow from the construction and are checked before any probe is fitted.
Inside a fixed-prompt slice the prompt is identical, so a text rule is blind
there. Inside a fixed-image slice the image is identical, so an image rule is blind there. Every single-modality rule therefore ranks the unsafe member first in at
most half of the groups and reaches at most $5/6$ pooled AUROC. The measured arms match
this bound: the text arm sits at the $0.50$ single-modality chance floor in every
fixed-prompt slice and at the $5/6$ ceiling pooled, and the image arm mirrors it.
The joint arm clears both slices, at \F{QuadFpIUQwenL11} and \F{QuadFiIUQwenL11} on Qwen3-VL-8B
and \F{QuadFpIUSmolL7} and \F{QuadFiIUSmolL7} on SmolVLM2-2.2B.

The conjunction is additively decodable. A probe on the stacked image and text states reaches \F{QuadAccConcatQwenL11} four-way ranking accuracy at $\mathrm{L2}$ on Qwen3-VL-8B, within the joint arm's interval, so the intermediate
layer holds an additively decodable cross-modal subspace that a linear readout
separates. At the deepest layer the joint final-token representation degrades, to
\F{QuadAccIUDeepQwen} on Qwen3-VL-8B and \F{QuadAccIUDeepSmol} on SmolVLM2-2.2B.

\subsection{Shared-Image Leakage in Request-Grouped MSSBench Splits}
\label{sec:bytelimit}

The image-disjoint split repairs the \emph{split} leak, but it does not explain what
the request-grouped number was reading in the first place. On MSSBench the answer is a
non-semantic identity channel: the benchmark's \F{ByteEvents} samples cover
\F{ByteImages} distinct images arranged in \F{ByteGroups} request pairs, and every
one of those images occurs under more than one pair, and the count is \F{ByteImgInGroups} of
\F{ByteImages}. A lookup on the \emph{file byte size} alone therefore separates
the images, and because the image carries an invariant label, the lookup separates
the labels too.

We measure that channel out of fold, folding over the request pairs so that no
image is seen in training and testing at once. On the native files the byte-size
lookup reaches \F{ByteNativeLookup} accuracy over \F{ByteNativeSizes} distinct
byte sizes, and on the standardised $448\times448$ re-encodings it still reaches
\F{ByteStdLookup} over \F{ByteStdSizes} sizes. No VLM, no hidden state and no
request text is involved.

Two consequences follow, and we state both because they cut in opposite
directions. First, the finding does not weaken the paper's central claim: the pool
\emph{is} image-determined, which is exactly what $\hat{H}(Y\mid I)=0$ asserts. The
byte-size channel is a mechanism for that fact and no counterexample to it. Second,
the within-benchmark reference quoted on MSSBench, the \F{RefQwenMSSRequest}
AUROC and the \F{RefTprQwenMSSRequest}\% TPR, is inflated by a channel that
carries no safety semantics, and it must be quoted with that caveat attached. This
is why the main paper reports every headline number under three split keys and
reports the image-disjoint key as the one a non-leaking benchmark would support:
under that key the MSSBench reference falls to \F{RefQwenMSSImage} AUROC and
\F{RefTprQwenMSSImage}\% TPR, and the ordering of cached and safety-prompted probes
on Qwen/MSSBench reverses
from a cached-readout win to an audit win.

The same measurement shows why a request-grouped fold is not a sufficient repair on
its own: because the image identity is re-derived from the file and never from the request, folding over requests leaves the identity channel intact. A
split keyed on the image itself removes it, which is why we recompute the full
comparison between cached and safety-prompted probes under the image-disjoint and
case-disjoint splits, and we do not assert that
the request-grouped split was unlucky.

\paragraph{Two request-grouped measurements of the same arm.}
Two pipelines report a request-grouped recall for the Qwen3-VL cached probe on MSSBench, and they differ. The four-arm comparison hashes the normalized request text, as specified in the main paper, which gives a $724/242/234$ fit, calibration, and test split; at layer $4$ the probe reaches \F{ArmAuReadQwenMSSRequest} AUROC and \F{ArmReadQwenMSSRequest}\% TPR, $103$ of $117$ unsafe test records, at \F{ArmFprReadQwenMSSRequest}\% FPR. The cascade pipeline hashes the raw request string, which gives a $708/254/238$ split; at the same layer the probe reaches \F{CascReqCachedTprQwenMSS}\% TPR, $100$ of $119$, at $5.9\%$ FPR. Both are real measurements of the same arm on different test sets. The main paper quotes the first, because it follows the stated split protocol; both are far above the \F{ArmReadQwenMSSImage}\% image-disjoint value, so the leakage conclusion does not depend on which is quoted.

\paragraph{Latency definitions.}
The main paper reports two latency measurements. Marginal latency comes from the batch-$1$ benchmark of this section, and speed-ups are ratios of its unrounded medians: \F{SpeedupQwenIndep}$\times$ for an independent pass and \F{SpeedupQwenKv}$\times$ for KV-cache reuse on Qwen3-VL-8B. An earlier version divided values already rounded to $0.01$\,ms and printed $417\times$ and $197\times$; we corrected these. End-to-end overhead comes from a separate in-process run that times prefill, probe, and safety pass together, giving \F{TpReadFracQwen}\% $= \F{MargReadQwen}/\F{TpQwen}$ on Qwen3-VL-8B. The probe timings of the two runs differ by a factor of \F{AnchorReadQwen} on Qwen3-VL-8B and \F{AnchorReadSmol} on SmolVLM2-2.2B, so the paper never mixes them in one ratio.

\section{Cached versus Safety-Prompted Probes under Request-Grouped Splits}
\label{sec:fourarm}

\subsection{Design of the comparison}

\label{sec:fourarmdetail}
This section reports the historical request-grouped comparison. Under the final image-disjoint splits of the main paper, within-benchmark safety passes lead on both Qwen3-VL suites and cached probes lead on both SmolVLM2 suites, so the ordering below should be read as a leakage-affected result. At the same operating point, same samples and same splits, we compare:
\begin{enumerate}
\itemsep1pt
\item the \emph{serving read-out}, fitted on this pool;
\item an \emph{audit head} fitted on this pool, probing the audit prefill's state;
\item \emph{E4's transferred audit head}, applied without refitting; and
\item the zero-shot prompt audit, where available.
\end{enumerate}
The comparison that matters is the serving read-out against the fitted audit head at the same realised \fpr, because
the audit head is the strongest version of the audit arm: it is fitted on the very pool it is
evaluated on, and therefore cannot be dismissed as a strawman.

\subsection{Results}

\begin{table*}[htbp]
\centering
\caption{The comparison between cached and safety-prompted probes, per pool and per
family. Realised \fpr and
\tpr on the test split at the \num{0.05} target. The read-out does not win
everywhere, and the relative ordering changes with the pool.}
\label{tab:fourarm}
\vspace{-2pt}
\input{tables/tab_fourarm}
\vspace{-4pt}
\end{table*}

On VSCBench with Qwen, the fitted audit head \emph{beats} the serving read-out,
\num{0.6022} against \num{0.5161} \tpr at a matched \fpr. On MSSBench the
ordering reverses and the read-out wins clearly, \num{0.8803} against \num{0.7607}
for Qwen, and \num{0.8889} against \num{0.3504} for SmolVLM2. On VSCBench with
SmolVLM2 the read-out wins as well, \num{0.3441} against \num{0.1505}. The
relative strength of the two paths therefore \emph{depends on the pool}, which is
itself an important observation: there is no pool-independent statement that the
read-out is more sensitive than an audit.

We accordingly \textbf{supersede} the claim that the read-out is more sensitive
than an independent audit. This boundary is recorded in the project's revision
log together with the measurement that established it, and we report it here
and do not omit the arm.

\subsection{A possible explanation for the pool dependence}

One possible explanation, which we have not tested, involves the retrained prompt ablation of the main paper. The audit
head reads the audit prefill's \emph{last prompt token}, under a safety prompt
that explicitly asks about safety in the context of the request; the serving
read-out reads a state that was computed for a serving purpose. The audit head
is therefore better placed to use the request, and on a pool whose labels are
image-driven it does not need to.

\subsection{Unavailable arm}

The zero-shot prompt audit is reported as \emph{unavailable} in the per-pool
tables, and it is not carried over from another script with a different sampling. A
number carried across scripts is exactly the kind of inconsistency a table like
this exists to remove, and we would rather have a hole than a false precision.

\section{Cascade Routing and the Measurement Boundary}
\label{sec:Scascade}

The routing rule must agree with the final decision. Let $R$ and $A$ denote
binary unsafe flags from the reader and auditor. Positive escalation runs the
auditor precisely when $R=1$ and implements confirmation, $D=R\land A$:
\begin{align}
\mathrm{FPR}_{\land} &= P(R=1,A=1\mid Y=0), \notag\\
\bar c_{\land} &= c_R + P(R=1)c_A .
\label{eq:e2e}
\end{align}
Negative escalation runs the auditor when $R=0$ and implements screening,
$D=R\lor A$:
\begin{align}
\mathrm{FPR}_{\lor} &= P(R=1\mid Y=0) \notag\\
&\quad + P(R=0,A=1\mid Y=0), \notag\\
\bar c_{\lor} &= c_R + P(R=0)c_A .
\end{align}
These are identities and require no independence assumption. Each measured
cost must use the batch size of its corresponding cost terms.

\begin{figure}[htbp]
\centering
\includegraphics[width=0.55\linewidth,height=0.46\textheight,keepaspectratio]{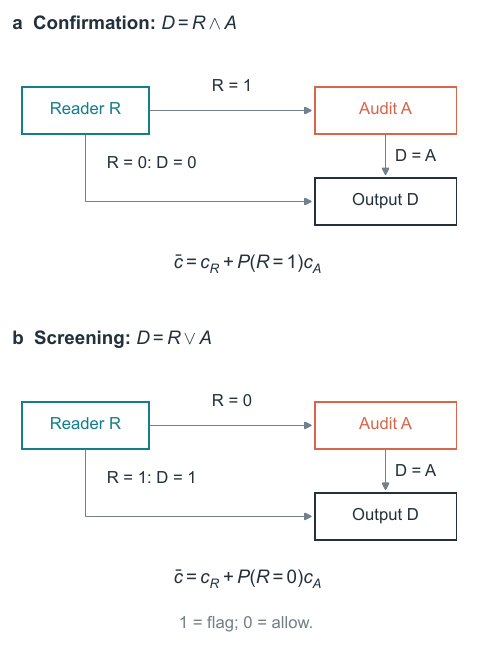}
\caption{Two consistent routing policies. Confirmation escalates reader-positive
requests and requires both flags. Screening escalates reader-negative requests
and flags either stage's positive decision. Reader $R$ is teal, audit $A$ is coral,
and both routes terminate at the explicit output $D$. The audit fraction in the cost
expression follows the route. This is an algebraic schematic and no measured cascade operating point.}
\label{fig:cascade}
\vspace{-4pt}
\end{figure}

\begin{table}[htbp]
\centering
\caption{\textbf{Exact two-stage cascade operating points under the 5\% end-to-end
false-positive budget.} Each row is the highest-detection budget-feasible point
over the two consistent routing policies, computed from the joint per-sample
decisions of the reader and the within-benchmark audit head; the \emph{policy} column
names the one that achieves it, either \emph{confirm} with $D=R\land A$, which escalates on
$R=1$, or \emph{screen} with $D=R\lor A$, which escalates on $R=0$. Escalation rate is the
fraction of requests sent to the auditor; cost ratio is the factor by which an
independent audit pass on every request exceeds the cascade's amortized
per-request cost.}
\label{tab:cascade_exact}
\vspace{-2pt}
\small
\setlength{\tabcolsep}{3pt}
\resizebox{\linewidth}{!}{\input{tables/tab_cascade_EXACT}}
\vspace{-4pt}
\end{table}

A prior cascade implementation combined positive escalation cost with an
OR decision. It also computes its stored audit-slice catch rate from reader
counts alone, without using audit predictions; this field can even be negative.
It therefore cannot identify the AND policy's true-positive rate. Those aggregate OR detections and positive escalation costs must not be presented as one
implemented cascade. The exact cascade metrics are reported in
Table~\ref{tab:cascade_exact}.

\input{sections/appendix}

%% file: tables/tab_label_structure.tex
\begin{tabular}{lrrrrrl}
\toprule
pool & $H(Y)$ & $H(Y|I)$ & $H(Y|u)$ & $\text{ceil}_{I}$ & $\text{ceil}_{u}$ & both labels \\
\midrule
VSC & 1.0000 & 0.0000 & 1.0000 & 1.0000 & 0.5000 & 0 of 1751 \\
MSS & 1.0000 & 0.0000 & 1.0000 & 1.0000 & 0.5000 & 0 of 578 \\
Paired & 1.0000 & 1.0000 & 0.0000 & 0.5000 & 1.0000 & 198 of 198 \\
\bottomrule
\end{tabular}

%% file: tables/tab_positive_control.tex
\begin{tabular}{lcrrrrrc}
\toprule
family & layer & real & ablated & swapped & image-tok & $\Delta$ & axis 2 \\
\midrule
SMOL & 0 & 0.8819 & 0.5000 & 0.8819 & 0.5000 & 0.3819 & \textbf{PASS} \\
SMOL & 1 & 0.9066 & 0.5000 & 0.9066 & 0.5000 & 0.4066 & \textbf{PASS} \\
SMOL & 2 & 0.9383 & 0.5000 & 0.9383 & 0.5000 & 0.4383 & \textbf{PASS} \\
SMOL & 3 & 0.9182 & 0.5000 & 0.9182 & 0.5000 & 0.4182 & \textbf{PASS} \\
QWEN & 0 & 0.8688 & 0.5000 & 0.8688 & 0.5000 & 0.3688 & \textbf{PASS} \\
QWEN & 1 & 0.9390 & 0.5000 & 0.9390 & 0.5000 & 0.4390 & \textbf{PASS} \\
QWEN & 2 & 0.9630 & 0.5000 & 0.9630 & 0.5000 & 0.4630 & \textbf{PASS} \\
QWEN & 3 & 0.9468 & 0.5000 & 0.9468 & 0.5000 & 0.4468 & \textbf{PASS} \\
\bottomrule
\end{tabular}

%% file: tables/tab_pools.tex
\begin{tabular}{lrr@{\,}p{0.40\linewidth}}
\toprule
pool & requests & axis 1 & axis 2 / note \\
\midrule
VLGuard & $-$ & \F{PixVLGuard} & request purity 1.000; the label is the request's function \\
MMSafeAware & $-$ & \F{PixMMSafe} & the label is the request's function \\
MOSSBench, type 2 & $-$ & \F{PixMoss} & pixel distribution shift \\
VSC & \F{ReqVSC} & \F{PixVSC} & fail, best layer 2, Qwen, $\Delta$=-0.0532; $H(Y|I)=0$, fixed image label \\
MSS & \F{ReqMSS} & \F{PixMSS} & fail, best layer 1, Qwen, $\Delta$=0.0037; $H(Y|I)=0$, fixed image label \\
Paired & \F{SieiReq} & \F{SieiPixel} & \textbf{PASS}, best layer 2, Qwen, $\Delta$=0.4630; $H(Y|I)=1.0000$, balanced control \\
\bottomrule
\end{tabular}

%% file: tables/tab_columns.tex
\begin{tabular}{llcrrrrr}
\toprule
family & pool & position & layer & \auroc & \fpr & \tpr & $R_{\text{pair}}$ \\
\midrule
SMOL & VSC & image-token & 1 & 0.8666 & 0.0484 & 0.4892 & 0.8920 \\
SMOL & VSC & \emph{prompt-end} & 0 & 0.7219 & 0.0645 & 0.2581 & 0.8410 \\
SMOL & MSS & image-token & 2 & 0.9822 & 0.0427 & 0.8889 & 0.9832 \\
SMOL & MSS & \emph{prompt-end} & 1 & 0.8639 & 0.0598 & 0.5299 & 0.9664 \\
QWEN & VSC & image-token & 2 & 0.9272 & 0.0538 & 0.6237 & 0.9352 \\
QWEN & VSC & \emph{prompt-end} & 2 & 0.8534 & 0.0376 & 0.4785 & 0.9352 \\
QWEN & MSS & image-token & 3 & 0.9934 & 0.0684 & 0.9402 & 1.0000 \\
QWEN & MSS & \emph{prompt-end} & 1 & 0.9784 & 0.0171 & 0.8034 & 1.0000 \\
\bottomrule
\end{tabular}

%% file: tables/tab_axis2.tex
family & pool & layer & real & ablated & swapped & image-tok & axis 2 \\
SMOL & VSC & 0 & 0.8199 & 0.8779 & 0.8116 & 0.8574 & fail \\
SMOL & VSC & 1 & 0.7734 & 0.8612 & 0.7294 & 0.8666 & fail \\
SMOL & VSC & 2 & 0.7272 & 0.7963 & 0.6661 & 0.8144 & fail \\
SMOL & VSC & 3 & 0.6974 & 0.7748 & 0.6607 & 0.7777 & fail \\
SMOL & MSS & 0 & 0.8497 & 0.8883 & 0.8304 & 0.9835 & fail \\
SMOL & MSS & 1 & 0.8741 & 0.9261 & 0.8261 & 0.9872 & fail \\
SMOL & MSS & 2 & 0.8529 & 0.9400 & 0.7835 & 0.9820 & fail \\
SMOL & MSS & 3 & 0.7746 & 0.9765 & 0.6887 & 0.9829 & fail \\
SMOL & Paired & 0 & 0.8819 & 0.5000 & 0.8819 & 0.5000 & \textbf{PASS} \\
SMOL & Paired & 1 & 0.9066 & 0.5000 & 0.9066 & 0.5000 & \textbf{PASS} \\
SMOL & Paired & 2 & 0.9383 & 0.5000 & 0.9383 & 0.5000 & \textbf{PASS} \\
SMOL & Paired & 3 & 0.9182 & 0.5000 & 0.9182 & 0.5000 & \textbf{PASS} \\
QWEN & VSC & 0 & 0.8228 & 0.8994 & 0.7930 & 0.8836 & fail \\
QWEN & VSC & 1 & 0.8502 & 0.9069 & 0.8061 & 0.9190 & fail \\
QWEN & VSC & 2 & 0.8424 & 0.8956 & 0.8008 & 0.9272 & fail \\
QWEN & VSC & 3 & 0.8139 & 0.9039 & 0.7719 & 0.8852 & fail \\
QWEN & MSS & 0 & 0.9511 & 0.9518 & 0.9436 & 0.9815 & fail \\
QWEN & MSS & 1 & 0.9698 & 0.9662 & 0.9424 & 0.9893 & fail \\
QWEN & MSS & 2 & 0.9448 & 0.9607 & 0.8881 & 0.9866 & fail \\
QWEN & MSS & 3 & 0.9050 & 0.9768 & 0.8458 & 0.9927 & fail \\
QWEN & Paired & 0 & 0.8688 & 0.5000 & 0.8688 & 0.5000 & \textbf{PASS} \\
QWEN & Paired & 1 & 0.9390 & 0.5000 & 0.9390 & 0.5000 & \textbf{PASS} \\
QWEN & Paired & 2 & 0.9630 & 0.5000 & 0.9630 & 0.5000 & \textbf{PASS} \\
QWEN & Paired & 3 & 0.9468 & 0.5000 & 0.9468 & 0.5000 & \textbf{PASS} \\

%% file: tables/tab_axis2_gate.tex
\begin{tabular}{llccccc}
\toprule
family & pool & order & ablated & swapped & verdict \\
\midrule
SMOL & VSC & image-first & 0.0e+00 & 0.0e+00 & PASS \\
SMOL & VSC & request-first & 1.1e+00 & 1.0e+00 & PASS \\
SMOL & MSS & image-first & 0.0e+00 & 0.0e+00 & PASS \\
SMOL & MSS & request-first & 8.3e-01 & 8.9e-01 & PASS \\
SMOL & Paired & image-first & 0.0e+00 & 0.0e+00 & PASS \\
SMOL & Paired & request-first & 9.7e-01 & 9.6e-01 & PASS \\
QWEN & VSC & image-first & 0.0e+00 & 0.0e+00 & PASS \\
QWEN & VSC & request-first & 2.8e+00 & 1.5e+00 & PASS \\
QWEN & MSS & image-first & 0.0e+00 & 0.0e+00 & PASS \\
QWEN & MSS & request-first & 2.5e+00 & 1.1e+00 & PASS \\
QWEN & Paired & image-first & 0.0e+00 & 0.0e+00 & PASS \\
QWEN & Paired & request-first & 2.1e+00 & 1.3e+00 & PASS \\
\bottomrule
\end{tabular}

%% file: tables/tab_transfer.tex
family & order & direction & layer idx & $R_0$ \fpr/\tpr & $R_3$ \fpr/\tpr & oracle \tpr & transfer \auroc \\
SMOL & image-first & mss$\to$vsc & 0 & 0.5715/0.6456 & 0.0429/0.0842 & 0.0901 & 0.5617 \\
SMOL & image-first & mss$\to$vsc & 1 & 0.5488/0.6372 & 0.0429/0.0800 & 0.0833 & 0.5635 \\
SMOL & image-first & mss$\to$vsc & 2 & 0.6338/0.7172 & 0.0497/0.0640 & 0.0640 & 0.5531 \\
SMOL & image-first & mss$\to$vsc & 3 & 0.6717/0.6246 & 0.0421/0.0278 & 0.0379 & 0.4858 \\
SMOL & image-first & vsc$\to$mss & 0 & 0.0762/0.1159 & 0.0497/0.0828 & 0.0861 & 0.5407 \\
SMOL & image-first & vsc$\to$mss & 1 & 0.1358/0.1854 & 0.0497/0.0629 & 0.0629 & 0.5302 \\
SMOL & image-first & vsc$\to$mss & 2 & 0.0166/0.0298 & 0.0497/0.0728 & 0.0762 & 0.5331 \\
SMOL & image-first & vsc$\to$mss & 3 & 0.0364/0.0530 & 0.0497/0.0861 & 0.0861 & 0.4783 \\
SMOL & request-first & mss$\to$vsc & 0 & 0.5901/0.6793 & 0.0497/0.0901 & 0.0909 & 0.5561 \\
SMOL & request-first & mss$\to$vsc & 1 & 0.6726/0.7694 & 0.0497/0.0808 & 0.0808 & 0.5706 \\
SMOL & request-first & mss$\to$vsc & 2 & 0.4949/0.6077 & 0.0497/0.0833 & 0.0833 & 0.5717 \\
SMOL & request-first & mss$\to$vsc & 3 & 0.5598/0.5522 & 0.0497/0.0480 & 0.0480 & 0.5065 \\
SMOL & request-first & vsc$\to$mss & 0 & 0.1126/0.1440 & 0.0480/0.0579 & 0.0646 & 0.5305 \\
SMOL & request-first & vsc$\to$mss & 1 & 0.0844/0.1192 & 0.0497/0.0844 & 0.0894 & 0.5489 \\
SMOL & request-first & vsc$\to$mss & 2 & 0.0281/0.0497 & 0.0497/0.0861 & 0.0861 & 0.5785 \\
SMOL & request-first & vsc$\to$mss & 3 & 0.0679/0.0927 & 0.0480/0.0679 & 0.0679 & 0.5160 \\
QWEN & image-first & mss$\to$vsc & 0 & 0.5749/0.6002 & 0.0505/0.0985 & 0.0985 & 0.5277 \\
QWEN & image-first & mss$\to$vsc & 1 & 0.7946/0.9108 & 0.0505/0.1279 & 0.1279 & 0.6301 \\
QWEN & image-first & mss$\to$vsc & 2 & 0.6330/0.7879 & 0.0497/0.0707 & 0.0741 & 0.6082 \\
QWEN & image-first & mss$\to$vsc & 3 & 0.6751/0.8476 & 0.0497/0.1532 & 0.1591 & 0.6710 \\
QWEN & image-first & vsc$\to$mss & 0 & 0.0596/0.1556 & 0.0497/0.1258 & 0.1258 & 0.5501 \\
QWEN & image-first & vsc$\to$mss & 1 & 0.0232/0.0993 & 0.0497/0.1656 & 0.1722 & 0.5797 \\
QWEN & image-first & vsc$\to$mss & 2 & 0.0364/0.0927 & 0.0497/0.1308 & 0.1391 & 0.5566 \\
QWEN & image-first & vsc$\to$mss & 3 & 0.0199/0.1060 & 0.0497/0.1556 & 0.1556 & 0.6132 \\
QWEN & request-first & mss$\to$vsc & 0 & 0.5093/0.5564 & 0.0497/0.0926 & 0.0993 & 0.5285 \\
QWEN & request-first & mss$\to$vsc & 1 & 0.6970/0.8535 & 0.0497/0.1347 & 0.1364 & 0.6476 \\
QWEN & request-first & mss$\to$vsc & 2 & 0.4756/0.5892 & 0.0497/0.1136 & 0.1136 & 0.5891 \\
QWEN & request-first & mss$\to$vsc & 3 & 0.6684/0.8148 & 0.0497/0.1380 & 0.1389 & 0.6456 \\
QWEN & request-first & vsc$\to$mss & 0 & 0.0877/0.1937 & 0.0497/0.1126 & 0.1175 & 0.5675 \\
QWEN & request-first & vsc$\to$mss & 1 & 0.0414/0.1705 & 0.0497/0.1838 & 0.1887 & 0.6308 \\
QWEN & request-first & vsc$\to$mss & 2 & 0.0430/0.2368 & 0.0497/0.2467 & 0.2517 & 0.7036 \\
QWEN & request-first & vsc$\to$mss & 3 & 0.0596/0.1126 & 0.0497/0.0877 & 0.0894 & 0.5749 \\

%% file: tables/tab_cost.tex
\begin{tabular}{llrrrrrr}
\toprule
family & design & $B{=}1$ & $B{=}2$ & $B{=}4$ & $B{=}8$ & $B{=}16$ & $B{=}32$ \\
\midrule
QWEN & cached probe & 0.234 & 0.181 & 0.149 & 0.122 & 0.113 & 0.100 \\
QWEN & sync to host & 0.017 & 0.026 & 0.044 & 0.083 & 0.159 & 0.315 \\
QWEN & independent pass & 95.817 & 123.042 & 87.267 & 68.939 & 59.681 & 112.309 \\
QWEN & KV-cache reuse & 45.271 & 24.435 & 26.213 & 15.925 & 10.716 & 9.904 \\
SMOL & cached probe & 0.226 & 0.182 & 0.150 & 0.126 & 0.112 & 0.105 \\
SMOL & sync to host & 0.017 & 0.027 & 0.045 & 0.085 & 0.161 & 0.318 \\
SMOL & independent pass & 223.407 & 184.212 & 163.785 & 165.759 & 156.154 & 150.836 \\
SMOL & KV-cache reuse & 31.839 & 16.823 & 12.882 & 8.104 & 5.573 & 4.764 \\
\bottomrule
\end{tabular}

%% file: tables/tab_fourarm.tex
\begin{tabular}{llcccc}
\toprule
pool & family & arm & \auroc & \fpr & \tpr \\
\midrule
VSC & QWEN & read-out & 0.8834 & 0.0538 & 0.5161 \\
VSC & QWEN & audit head & 0.8972 & 0.0591 & 0.6022 \\
VSC & QWEN & transferred head & 0.7251 & 0.0269 & 0.1344 \\
VSC & SMOL & read-out & 0.8144 & 0.0591 & 0.3441 \\
VSC & SMOL & audit head & 0.6618 & 0.0484 & 0.1505 \\
VSC & SMOL & transferred head & 0.4997 & 0.0645 & 0.0645 \\
MSS & QWEN & read-out & 0.9820 & 0.0256 & 0.8803 \\
MSS & QWEN & audit head & 0.9467 & 0.0940 & 0.7607 \\
MSS & QWEN & transferred head & 0.8170 & 0.0342 & 0.3077 \\
MSS & SMOL & read-out & 0.9829 & 0.0256 & 0.8889 \\
MSS & SMOL & audit head & 0.7646 & 0.0769 & 0.3504 \\
MSS & SMOL & transferred head & 0.7032 & 0.0256 & 0.1282 \\
\bottomrule
\end{tabular}

%% file: tables/tab_cascade_EXACT.tex
\begin{tabular}{llccccccc}
\toprule
pool & family & policy & reader \fpr & reader \tpr & cascade \fpr & cascade \tpr & escalation & cost ratio \\
\midrule
MSS & QWEN & screen & 0.0256 & 0.8803 & 0.0427 & 0.9231 & 58.1\% & 1.7$\times$ \\
VSC & QWEN & screen & 0.0538 & 0.5161 & 0.0430 & 0.4409 & 86.0\% & 1.2$\times$ \\
MSS & SMOL & confirm & 0.0256 & 0.8889 & 0.0000 & 0.2991 & 45.7\% & 2.2$\times$ \\
VSC & SMOL & screen & 0.0591 & 0.3441 & 0.0430 & 0.2204 & 90.6\% & 1.1$\times$ \\
\bottomrule
\end{tabular}

%% file: sections/appendix.tex
% This file is the appendix of the merged arXiv document. The appendix
% counter and the "S" section-numbering are set by the main file, so the
% appendix continues the supplementary material's numbering instead of
% restarting it.
\section{Protocol History and Superseded Claims}
\label{sec:prereg}

\subsection{Purpose}

This appendix records how the protocol changed during the project, so that a reader can
check each reported result against the version of the protocol that produced it.
Decision rules are fixed in the released analysis scripts. We cannot provide external
proof of when each rule was fixed, so we describe them as recorded criteria, and we list
every claim that a later measurement superseded.

\subsection{Revision log}

The table below is the log.  Every row is a claim we made, the measurement
that superseded it, and the fact that we superseded it ourselves, before
submission and without being asked.  We keep it because a reader deciding
whether to trust this paper's negative results has a right to see what the
positive results were before the corrections.

% A breakable long table: the log is long enough that, as a float, it would
% claim a page of its own.
{\footnotesize
\setlength{\tabcolsep}{4pt}
\renewcommand{\arraystretch}{1.05}
\begin{longtable}{@{}p{0.28\linewidth}p{0.36\linewidth}p{0.11\linewidth}@{}}
\caption{Revision log. Each row records a superseded claim and the measurement that superseded it; all revisions were made before submission.}
\label{tab:retractions}\\
\toprule
claim & what killed it & status \\
\midrule
\endfirsthead
\toprule
claim & what killed it & status \\
\midrule
\endhead
\bottomrule
\endfoot
the read-out is more sensitive than an independent audit &
the fitted audit head reaches \num{0.6022} \tpr against the read-out's
\num{0.5161} at a matched \fpr on VSC/Qwen \cref{tab:fourarm} &
\textbf{superseded} \\
\addlinespace
the read-out is $\num{543}\times$ cheaper than an audit &
the ratio divided a batch-1 numerator by a batch-8 denominator; the absolute
milliseconds are \num{0.23} against \num{95.8} at a matched batch &
\textbf{superseded} \\
\addlinespace
the read-out wins on every pool &
the ordering flips with the pool, and the pre-fix split assignment put
comparison rows and samples in different splits \cref{sec:protocolfix} &
\textbf{superseded} \\
\addlinespace
the read-out is pool-general &
the source threshold has unstable target FPR, and transferred AUROC
varies across pools; the earlier mixed-metric gap is superseded
\cref{tab:transfer} &
\textbf{superseded} \\
\addlinespace
the prompt-end position is the more sensitive one &
the request-blind image-token position has the higher \auroc on all four
pool-family combinations \cref{tab:columns} &
\textbf{superseded} \\
\addlinespace
the image-token state is a valid contextual read-out &
it is request-blind by construction; the same-image pair distance is
\rng{\num{0}}{\num{3.04e-03}} against \rng{\num{5.5e-02}}{\num{7.5e-01}} at the request
endpoint \cref{tab:blind} &
\textbf{superseded} \\
\addlinespace
conditional-coverage supervision helps &
the shortcut-matched arm does not beat the control at the required significance
after Bonferroni correction across 32 tests &
\textbf{not established} \\
\addlinespace
the axis-2 measurement is reproducible across extraction runs &
the extractor filed each saved row under the wrong sample identifier because it
reshaped the accumulated states with the wrong stride; the check failed by up to
\num{3.7e+02} and the axis-2 numbers were recomputed after the fix
\cref{sec:batchinv} &
\textbf{superseded} \\
\addlinespace
the deviation in Qwen3-VL shows that the request reaches the image
tokens &
the deviation is a sequence-length effect: a length-matched placeholder that
carries no request information moves the image-token state by more than the
ablation does; the three conditions are now length-matched and the check is
exactly zero \cref{sec:lengthbug} &
\textbf{superseded} \\
\addlinespace
the axis-2 failure means the read-out ignores the request &
the pool is image-determined: $H(Y\mid I)=\num{0.0000}$ bits, measured on the
labels alone, so the request carries no information about the label and the
control is unpassable on that pool by any read-out
\cref{sec:context}, \cref{tab:labelstructure} &
\textbf{superseded} \\
\addlinespace
the axis-2 control is a valid admission axis &
the control had never been shown capable of passing; it is now run on a pool
built to carry contextual variation, where it passes \cref{sec:positive} &
\textbf{superseded} \\
\end{longtable}
}

\subsection{Results that remain valid}

\begin{itemize}
\itemsep1pt
\item the pixel control and the request-ablation control of \cref{eq:axis2};
\item the structural blindness of the image-token position shown in \cref{sec:blind};
\item the context-free-pool theorem and its verification in \cref{sec:context} and
  \cref{tab:labelstructure}: $H(Y\mid I)=\num{0.0000}$ on both pools that
  clear axis 1, so the axis-2 control is unpassable on them by any read-out;
\item the label-assisted quantile threshold recalibration: 16/16 configurations meet the recorded criterion
  of \cref{sec:calib};
\item the transferred global AUROC is measured; the mixed-metric gap and its
  pixel-recovery interpretation are superseded, as \cref{sec:repr} records;
\item the latency ordering of the three audit designs, under the internal
  consistency check of \cref{eq:costcheck} and the answer-equivalence check;
\item the cascade routing identities of \cref{sec:Scascade}; the exact numerical
  cascade comparison is in Table~\ref{tab:cascade_exact}.
\end{itemize}

\subsection{Recorded analysis rules}

The following rules govern every analysis in the main paper and in this
supplement. Negative results are reported with the same prominence as positive ones.

\begin{enumerate}
\itemsep1pt
\item No selection of $\phi$, $\psi$, the layer, or any hyper-parameter on the
  calibration or test split. The layer is chosen on the training split's in-sample
  \auroc.
\item No threshold other than the one calibrated on the calibration split.
\item Every reported \fpr and \tpr is out of sample, with a Wilson interval.
\item Every pool's labels were counted as exposed as soon as we scored them.
\item When a control fails, we report it in the verdict, never in a footnote.
\item Every table and figure in the paper is generated by script from the JSON
  artifacts, so that a caption cannot drift from the number it describes.
\end{enumerate}

\section{Validation of Split Allocation}
\label{sec:protocolfix}

\subsection{The defect}

The split assignment in the comparison-row view of the pool was stored twice:
once as an \emph{sample-level} vector with two entries per comparison row, since a
comparison contributes one benign and one harmful sample, and both carry their
request's split, and once as a \emph{comparison-level} vector with one entry per
row.

Several downstream scripts zipped the \emph{sample-level} vector with the
\emph{comparison-row} list. Because the two have different lengths, the zip
silently truncated to the shorter one, pairing comparison row $i$ with the split
entries $2i$ and $2i+1$ in place of $2i$. Every row after the first was therefore
assigned to the wrong split, with the misalignment compounding.

\subsection{Detection by independent reimplementation}

Every key protocol quantity is computed by two independent implementations,
which must agree. The second implementation
was written for the split allocation itself. It disagreed with the first, and
the disagreement localised the zip.

\subsection{Consequence for the reported comparison}

The numerical effect of the defect was modest, but its effect on the conclusion was not: the
pre-fix and post-fix cached-versus-safety-prompted comparison tables \emph{reversed}
the paper's central
conclusion, and the pre-fix tables had supported a claim that the post-fix tables
refute. The project had preserved both sets of tables, with the pre-fix ones
retained explicitly for auditability, and the reversal was visible because
they were retained.

This is the argument for keeping superseded artifacts: had we deleted the
pre-fix tables as ``incorrect'', the revision would have looked like a change of mind and hidden the discovery of a bug.

\subsection{The other two deviations, and their equivalence checks}

\begin{itemize}
\itemsep1pt
\item \textbf{Sharded placement.} One backbone was run with accelerate's
  \texttt{device-map=auto} because a single device did not have room for it, while
  the reference run placed the whole model on one device. The equivalence was
  checked per sample and per layer against the project's own single-device golden
  features: relative $L_2$ mean \num{0.00475}, median \num{0.00274}, maximum
  \num{0.0257}; cosine mean \num{0.999980}, minimum \num{0.999689}. The same-image
  relative distance reproduced the reference run's order of magnitude.
\item \textbf{A controller outage.} One extraction was forced onto a different
  node because the intended node was drained by an administrator. The protocol
  was unchanged and the artifact is marked.
\end{itemize}

\section{Validation of the Visual KV-Cache Reuse Pass}
\label{sec:kvtrap}

The visual KV-cache reuse design is the cheapest of our three audit designs, and it is also the
easiest to implement incorrectly. We list four implementation errors that each
produce a plausible number that is not the claimed quantity, together with the check that excludes each.

\paragraph{The answer is not in the shared prefix.} The fork shares the serving
prompt's KV cache, so the audit's answer tokens are generated under the forked
state. A cached prefix plus a safety suffix is not the same computation as a
safety prompt over the same content, and the difference is exactly what the audit
is supposed to measure.

\paragraph{Position ids must not be reset.} The forked continuation continues
the serving prompt's position numbering. Resetting it changes the rotary
embedding's phase for the audit tokens and silently degrades the audit.

\paragraph{Attention must cover the whole cache.} The serving prefix's keys and
values are present in the cache but must be attended to by the audit tokens.
Restricting the audit's attention to its own suffix turns the audit into a reader of text alone, which is the very configuration the request-blindness analysis
rules out.

\paragraph{Deepstack and the multimodal embedding merge.} In architectures that
merge vision embeddings into the text embedding at several depths, a forked
prefill must carry the same merged embeddings at the same depths. A fork that
reuses the text-side embeddings but re-derives the vision side is not the same
computation.

We report the visual KV-cache reuse latency in \cref{tab:cost} with these conditions
satisfied, and we treat any implementation that does not satisfy them as
measuring a different design.

\section{Interpretation of the Retrained Prompt Ablation}
\label{sec:whyaxis2}

\subsection{Observation and initial hypothesis}

The retrained prompt ablation, called the axis-2 control in our scripts, fails on every pool, every family and every layer: the
prompt-end read-out scores the risk label about as well with the request
\emph{removed} as with it in place. The initial hypothesis was that the
read-out ignores the request.

That reading cannot be right, and the reason is visible in the numbers before
any further experiment. In the image-first order the maximum of
$\lvert\text{real}-\text{ablated}\rvert$ is \emph{exactly equal} to the maximum
of $\lvert\text{real}-\text{swapped}\rvert$, layer by layer, on every pool that
failed. The ablated request and the swapped request are different strings, so a
deviation caused by the request \emph{text} would give two different maxima.
The deviation is therefore a function of the sample and independent of its request.

\subsection{The measurement that settled it}

We then measured the thing the read-out is actually being asked to do, on the
labels alone and with no model involved: the conditional entropy of the risk
label given the image, $H(Y\mid I)$, and given the request, $H(Y\mid u)$.

The answer is \num{0.0000} bits on both pools. The label is a \emph{deterministic
function of the image}. The best possible rule over images alone reaches \auroc
\num{1.0000} and the best possible rule over requests alone reaches \num{0.5000}, which
is chance. None of the 1751 VSCBench images and none of the 578
MSSBench images carries both labels, so no image can be made to carry both
labels by relabelling, and the two classes use disjoint image sets.

Under those conditions the axis-2 control is \emph{unpassable}: no read-out
can pass it. Removing the request cannot destroy information that the
read-out was using, because the request held none. The negative result is a property of the benchmarks and says nothing about the read-out. Note the logical direction
carefully, because it is the one that is easy to state wrongly: the
measurement above is what establishes the property of the benchmarks. The
axis-2 failure is \emph{consistent with} it, and the two agree, but the
failure alone does not prove it; a weak read-out would fail on any pool.
The protocol is therefore two independent measurements that corroborate each other, and no single inference does two jobs.

\subsection{Why the benchmarks are built this way}

This is not an artefact of our pool construction. MSSBench ships, for each
case, a safe image, an unsafe image and \emph{two query strings}, and the label
is assigned from the image column; the second query is a rephrasing of the
same intent, so the label is the same under both queries by construction. The
VSCBench image-centric columns have the same construction: one prompt, two
image columns, and the label is the column. Two independently built safety
benchmarks both define the task as \emph{image risk under a fixed request}.

\subsection{The positive control, and why we needed it}

\label{sec:positive}
A control that has never been shown capable of passing cannot support the
negative results it produces. We therefore built a pool that carries genuine
contextual variation and ran the identical protocol on it.

The pool is built from the one source of paired opposite-intent structure we had: the MOSSBench
extraction, in which the same image occurs under two different requests with
opposite labels. We take those 198 pairs as they are. The image is
bit-identical between the two members of a pair, so the pool is unsolvable by any rule over images alone and its image ceiling is exactly \num{0.5000}; the
request determines the label and the request ceiling is \num{1.0000}. The
label structure is the exact mirror of the VSC/MSS pools, and so is the
expected verdict: the prompt-ablation control must \emph{pass} here.

Two details of the protocol had to change for this pool, and both are worth
stating because they are the details that make it a control and distinguish it from a variant.

\paragraph{The split is made over the pair and never over the request.} On the public
pools the split is a hash of the request text. Here the two members of a pair
carry different request texts, so hashing the request would scatter a pair
across two splits and destroy the single comparison the pool exists to make. The
split key is the pair id, and both members inherit it.

\paragraph{The swap is the partner's request.} Elsewhere the swap sends a sample
to the next request in the sorted request set. On this pool that would mix a
contextual effect with an unrelated one and dilute the measurement. The
partner's request is available and is the single most informative swap
available: it keeps the image bit-for-bit and moves the request to the one that
reverses the answer. A read-out that reads the request must change its answer
under this swap, and the size of the change is the size of the contextual
effect.

\section{Reproducibility}
\label{sec:repro}
\subsection{Batch-invariance validation of feature extraction}

\label{sec:batchinv}
During the prompt-ablation measurement, two extractions of the same samples disagreed. The image-token state of a sample, extracted
twice by two different pipelines, differed by up to \num{3.7e+02} in absolute
units, while the cosine similarity between the two stayed above \num{0.99}.
Under causal attention in the image-first order the image tokens \emph{precede}
the request, so the state of a sample must be a function of that sample alone,
and a difference of that size means that one of the two measurements was not
the quantity we thought it was.

We chased the discrepancy through four hypotheses before the right one, and we
record the wrong ones because the sequence is the argument.

\begin{enumerate}
\itemsep1pt
\item \textbf{It is not the request text.} The ablated and the swapped states,
  extracted in one pass, agree with each other to \num{0.0000e+00} for Smol and
  \num{3.8e-03} for Qwen while both differ from the real state. Under causal
  attention the request cannot reach the image tokens, so the two conditions
  that share no request information agree and the one that carries a request
  differs. That is the signature of a request-conditioned effect elsewhere in
  the tensor and excludes a bug in the image-token positions.
\item \textbf{It is not an indexing error.} For every sample the nearest row of
  the other tensor is a random row and never the aligned one: the aligned row is
  the nearest neighbour for at most \num{1.00}\% of samples. A permutation of
  the rows would have made the aligned row the nearest neighbour.
\item \textbf{It is not the batch composition.} We tested this directly and did
  not argue about it, and the test has to change \emph{which samples share the batch}, because changing the batch length alone would leave its composition fixed. We extracted the same
  \num{64} samples four ways: alone, with their file-order neighbours, with
  their request-group neighbours as the earlier extraction grouped them,
  and with a shuffled partner. The image-token state was bit-identical in all
  four cases at every probe layer, with a maximum absolute difference of
  \num{0.000000e+00} and \num{0} of \num{64} samples differing at any layer.
  The state does not depend on the batch.
\item \textbf{It is the reshape.} The two code paths build bit-identical model
  inputs, with identical token ids, identical pixel values and identical image
  grids, and produce bit-identical states when they run in the same process.
  The difference was therefore in how the results were \emph{filed}. Locating a
  single known state inside the saved tensor showed that it was filed at a row
  and a layer other than the ones its row identifier named.
\end{enumerate}

The fourth hypothesis is the one that held, and the cause was a bookkeeping bug that involved nothing about the model or the batch. The extractor
accumulated the per-sample states over the whole run into one flat list, in the
order $(\text{chunk}, \text{layer}, \text{item})$, and then reshaped that list once at the end as
$(\text{layer}, \text{item})$. The two orders coincide for the first batch of rows of the
first layer alone, so every other row of the saved tensor carried the state of a
different sample \emph{and} of a different layer. The extractor that produced
the reference tensors reshaped per chunk and concatenated the chunks, so its
tensors were correctly aligned; the two pipelines disagreed because one of
them was scrambled, and the model behaved identically in both.

The fix is to reshape per chunk and concatenate, exactly as the reference
extractor does. After the fix, the same samples extracted by the two pipelines
agree closely, and the size of the residual is set by the numerical precision of the architecture and owes nothing to the bookkeeping. On
SmolVLM2 the two extractions are \emph{bit-identical} at every probe layer
with a median per-sample maximum absolute difference of \num{0.0000e+00}. On
Qwen3-VL the residual grows with depth, from a median of \num{2.0e-03} at the
first probe layer to \num{8.3e-01} at the deepest one on MSSBench, against
the \num{3.0e+02} that the scrambled tensor showed. That residual is bf16 arithmetic and excludes misalignment: at the deepest layer the standard
deviation of the state itself is \num{5.54}, so the median difference is
\num{1.50e-01} of the quantity being measured, whereas the scrambled tensor
differed from the aligned one by more than its own scale. The check, which is
computed \emph{within} one pass, is exactly zero for both families.

\subsection{Length-matching validation of the invariance check}

\label{sec:lengthbug}
With the reshape fixed, the causal invariance check of \cref{tab:axis2check} became
exactly zero for SmolVLM2 as the architecture requires, but it did \emph{not}
become zero for Qwen3-VL. The image-token state moved under the ablation by
\num{1.75e-02} at the first layer, growing with depth to \num{4.98e+00} at the
last. Read naively, that is a property of the architecture: it would say that
Qwen3-VL lets the request reach the image tokens and SmolVLM2 does not. We
report that reading because it is the one a reader would reach for, and then
we refute it, because it is wrong.

Under causal attention the image tokens precede the request and \emph{cannot}
receive anything from it, in either architecture. So the deviation had to come
from somewhere else, and there were three candidates: numerical round-off, a
mechanism by which the text attends back into the image, and an artifact of the
measurement.

We excluded the round-off by measuring the noise floor. We excluded the
mechanism by a direct control, and the control is what settles it.

\paragraph{The length-matched control.}
The three conditions carry different request texts and therefore different
sequence lengths. A padded batch of a different length has a different tensor
shape, and a different shape means a different reduction order inside the
bf16 kernels, which moves the state of every position, including
the image tokens, which never saw the request at all. This is not a subtlety
of Qwen; it is a property of bf16 batched matmuls.

The control is therefore: take the ablated prompt, pad it with neutral filler
to the \emph{same token length} as the real prompt, and measure again. The
padded prompt carries no request information whatsoever, so under the
mechanism hypothesis it must leave the image-token state untouched, and under
the length hypothesis it must move it by about as much as the ablation does.

The measurement is unambiguous. On Qwen3-VL, at the deepest layer, the
ablation moves the image-token state by \num{2.06e+00}, and the
\emph{length-matched placeholder}, which contains no request information
whatsoever, moves it by \num{2.60e+00}, that is, by \emph{more}. At the
shallowest layers the two are within a factor of two of each other, and at
several of them the length-matched control is the larger of the two.

We confirmed the same thing from the other direction, without the model at all.
The pools contain pairs of samples that share an image and differ in their
request length, so the difference between their image-token states \emph{is}
the length effect, measured inside the extracted data. On MSSBench that pure
length effect reaches \num{3.49e+00} at the deepest layer, against an
ablation effect of \num{4.98e+00}; on VSCBench, \num{3.78e+00} against
\num{5.19e+00}. For SmolVLM2 the same quantity is exactly \num{0.000e+00},
which is why SmolVLM2's check was clean and Qwen3-VL's was not: SmolVLM2's
state is simply less sensitive to the padded length.

\paragraph{The fix.}
The three conditions of a sample are now padded to a common length with
neutral filler tokens, placed between the request and the template's closing
tokens so that the position the analysis reads is the same kind of position in
all three. The extractor \emph{aborts} if the three lengths differ for any
sample, so the guarantee is executable and enforced.

\paragraph{What this changes for the paper.}
Nothing in the reported numbers changed, because the axis-2 measurement compares
the \emph{read-out's AUROC} under the three conditions, and the length effect
is a common-mode perturbation that it does not reverse. What changed is the
\emph{check}, which is now exactly zero for both families, and the
\emph{interpretation}, which is now the correct one.

But the episode is worth reporting for what it is: a falsifiable check on the
instrument produced a discrepancy that looked like a finding about two
architectures, and what looked like a finding was in fact a property of the
measurement. The mechanism hypothesis was unsupported; it was
quantitatively dominated by the mundane explanation, and the sole way to know
which was true was to build a control that carries no request information and
change nothing else.

\subsection{Validated properties of the extraction}

\begin{enumerate}
\itemsep1pt
\item \textbf{The state is a function of the sample.} It does not depend on the
  batch composition, and we measured that directly: the same
  samples extracted alone, with their file-order neighbours, with their
  request-group neighbours and with a shuffled partner gave a bit-identical
  state at every probe layer. A read-out that reuses states the serving
  model has already computed is therefore not buying a re-computation.
\item \textbf{A comparison across two extractions is not a comparison of the
  model.} It is a comparison of the model \emph{and} of two bookkeeping
  conventions. The axis-2 measurement of this paper therefore extracts the
  real, the ablated and the swapped conditions in one process, in one code
  path, with one batching, and computes its causal invariance check inside that pass.
\item \textbf{Equal conditions must be equal in every respect that the
  computation can see.} Length is such a respect, and it is not an obvious
  one. We now pad the three conditions to a common length and the extractor
  refuses to proceed otherwise.
\item \textbf{A check that can fail is worth more than a check that cannot.} The
  check caught a bookkeeping bug that four plausible hypotheses had failed to
  explain, and then it caught a second one, in the opposite direction, that
  had produced a spurious architectural finding.
\end{enumerate}

We do not claim either is a defect of the model. They are properties of the
measurement, and they are the kind of property that silently invalidates a
cross-run comparison, which is why they are written down.

\section{Artifact}
\label{sec:artifact}

\subsection{What is released, and how to check it}

\begin{itemize}
\itemsep1pt
\item The admission-audit tool: a runnable checker that evaluates a pool's axis-1
  pixel control, its axis-2 request-ablation control and its text control, and
  prints a pass/fail verdict per axis.
\item The pool construction code, including the per-image per-channel
  normalisation and the request-internal statistics.
\item The extraction, fitting and evaluation scripts for every number in this
  paper, together with the superseded versions that the revision log refers to.
\item The split recipe: a deterministic hash of the request text, specified in
  full in the released code.
\item This protocol history and revision log.
\item The figure and table generators, which read the JSON artifacts directly.
\end{itemize}

Every table and every figure in both documents is generated by a script from a
JSON artifact, and no number in either document was entered by hand. The
release manifest enumerates the artifacts, the generators and the manuscript
files with a SHA-256 digest for each, so a reader can verify that a table in
the paper and a table in the artifact directory are the same object. The
verification script recompiles the manuscript and re-reads every number the
manuscript quotes, and fails loudly on the first discrepancy.

\subsection{Determinism}

Every split is a hash of the request text, so it is reproducible without any
state file. Every fit is a closed-form ridge solution, so there is no training
seed to record. The figures and tables are generated from the JSON artifacts by a
single script each, so a figure cannot disagree with the table it accompanies.

\subsection{Compute}

Two model families, four probe layers each, on a single node with 8 H20 GPUs. The
extractions are the expensive part; the analyses run on CPU and take seconds.
The total extraction budget for the ablated conditions reported here is
approximately 3 GPU-hours per family per pool.

\section{Claim Status}
\label{sec:claims}

\cref{tab:claims} summarizes which claims are established, which are
refuted, and which were not tested.

{\footnotesize
\setlength{\tabcolsep}{4pt}
\renewcommand{\arraystretch}{1.05}
\begin{longtable}{@{}p{0.40\textwidth}p{0.15\textwidth}p{0.32\textwidth}@{}}
\caption{Claim status. ``Established'' means measured on the pools named, with
the protocol of \cref{sec:protocol}, out of sample.}
\label{tab:claims}\\
\toprule
claim & status & evidence \\
\midrule
\endfirsthead
\toprule
claim & status & evidence \\
\midrule
\endhead
\bottomrule
\endfoot
the image-token state is request-blind & established & \cref{sec:blind} \\
\addlinespace
the prompt-end state is request-conditioned by construction &
established & \cref{sec:blind}, \cref{sec:protocolfix} \\
\addlinespace
no available public pool passes both admission axes &
established & \cref{tab:pools} \\
\addlinespace
the pixel control is a valid first admission axis &
established & \cref{sec:pools} \\
\addlinespace
the read-out is pool-bound & established & \cref{tab:transfer} \\
\addlinespace
the read-out is a cheap calibrated ranker & established &
\cref{tab:cost}, \cref{tab:fourarm} \\
\addlinespace
the cascade is counterproductive & \textbf{superseded} & \cref{sec:cascade} \\
\addlinespace
the read-out is more sensitive than an audit & \textbf{refuted} &
\cref{tab:fourarm} \\
\addlinespace
the read-out is pool-general & \textbf{refuted} & \cref{tab:transfer} \\
\addlinespace
the transfer deficit has a pixel-shortcut explanation &
not established & \cref{sec:shortcut} \\
\addlinespace
the prompt-end position is the stronger read-out & \textbf{refuted} &
\cref{tab:columns} \\
\addlinespace
the read-out outperforms published guards & \textbf{not tested} &
\cref{sec:limitations} \\
\addlinespace
same-image/different-intent supervision helps & \textbf{not established} &
\cref{sec:limitations} \\
\end{longtable}
}

\section{Limitations}
\label{sec:limitations}

The main paper summarizes these limitations in its Discussion; they are set out in full here.
None of them is addressed by the protocol. \emph{The auditor is not the strongest available}: our audit arm
is the same frozen backbone under a safety prompt, and we do not compare
against published safety-tuned guards \cite{guardreasoner,vlmguard,singguard},
rubric judges, or ensembles, so the axis-2 failure is a statement about the pools and says nothing about those guard models. \emph{Every target pool is exposed}: all
five candidate source pools were audited, so there is no locally
held-out clean same-request pool, and that absence is itself a finding
as \cref{tab:pools} records. \emph{Two backbones, one label}:
all results are binary situational risk, with two model families
\cite{qwen3vl,smolvlm2,llama3} at four probe layers each. \emph{One implementation,
one node, unresolved transfer}: the cost numbers are latencies from one
implementation on one node; transfer performance remains unexplained and
the mixed-metric gap is superseded; and the axis-2 control
has a known blind spot: a read-out that conditions on the request in a way
that correlates with risk without being the intended reasoning would pass it.
\emph{And the label structure is a property of the pools we could obtain}: a
benchmark shipping a genuine same-image/different-intent split would change
what the axis-2 verdict means, which is why \cref{sec:positivecontrol} builds
one and does not assume the result.

\subsection{Scope of the prompt-ablation result}

Because the direction of the inference matters for how this result may be
used, we state it once more in its correct form. The theorem of
\cref{sec:context} runs in one direction: on an image-determined pool the
axis-2 control cannot be passed. Its contrapositive is that a pool on which
\emph{every} read-out fails the control is \emph{consistent with} being
image-determined, but it does not prove it, because a weak read-out would fail
on any pool. We therefore never infer a pool's label structure from the
axis-2 verdict alone: we measure it independently, on the labels alone
as \cref{tab:labelstructure} reports, and the two pieces of evidence agree. It is the
agreement: two measurements made independently that both point the same
way, makes the negative result immune to the objection that our
read-out was too weak. The same reasoning motivates the agreement checks
reported in the supplement, which follow the classical practice of treating
two independent instruments that concur as stronger evidence than either
alone~\cite{pearson_corr}.